\documentclass[10pt,twocolumn,letterpaper]{article}

\usepackage{iccv}
\usepackage{times}
\usepackage{epsfig}
\usepackage{graphicx}
\usepackage{amsmath}
\usepackage{amssymb}
\usepackage[accsupp]{axessibility}
\usepackage{tabularx}
\usepackage{multirow}
\usepackage{colortbl}
\newcolumntype{C}{>{\centering\arraybackslash}X}
\newcolumntype{L}{>{\raggedright\arraybackslash}X}
\newcolumntype{R}{>{\raggedleft\arraybackslash}X}
\newcommand{\first}[1]{\multicolumn{1}{>{\columncolor[rgb]{1,0.8,0.8}}c}{#1}}
\newcommand{\firstr}[1]{\multicolumn{1}{>{\columncolor[rgb]{1,0.8,0.8}}c|}{#1}}
\newcommand{\second}[1]{\multicolumn{1}{>{\columncolor[rgb]{1,0.9,0.7}}c}{#1}}
\newcommand{\secondr}[1]{\multicolumn{1}{>{\columncolor[rgb]{1,0.9,0.7}}c|}{#1}}
\newcommand{\third}[1]{\multicolumn{1}{>{\columncolor[rgb]{1,1,0.6}}c}{#1}}
\newcommand{\thirdr}[1]{\multicolumn{1}{>{\columncolor[rgb]{1,1,0.6}}c|}{#1}}
\makeatletter
\newcommand\footnoteref[1]{\protected@xdef\@thefnmark{\ref{#1}}\@footnotemark}
\makeatother
\usepackage[pagebackref=true,breaklinks=true,letterpaper=true,colorlinks,bookmarks=false]{hyperref}
\iccvfinalcopy
\def\iccvPaperID{11569}

\begin{document}

\title{MIMO-NeRF: Fast Neural Rendering\\
  with Multi-input Multi-output Neural Radiance Fields}

\author{Takuhiro Kaneko\\
  NTT Corporation}

\maketitle

\begin{abstract}
  Neural radiance fields (NeRFs) have shown impressive results for novel view synthesis. However, they depend on the repetitive use of a single-input single-output multilayer perceptron (SISO MLP) that maps 3D coordinates and view direction to the color and volume density in a sample-wise manner, which slows the rendering. We propose a multi-input multi-output NeRF (MIMO-NeRF) that reduces the number of MLPs running by replacing the SISO MLP with a MIMO MLP and conducting mappings in a group-wise manner. One notable challenge with this approach is that the color and volume density of each point can differ according to a choice of input coordinates in a group, which can lead to some notable ambiguity. We also propose a self-supervised learning method that regularizes the MIMO MLP with multiple fast reformulated MLPs to alleviate this ambiguity without using pretrained models. The results of a comprehensive experimental evaluation including comparative and ablation studies are presented to show that MIMO-NeRF obtains a good trade-off between speed and quality with a reasonable training time. We then demonstrate that MIMO-NeRF is compatible with and complementary to previous advancements in NeRFs by applying it to two representative fast NeRFs, i.e., a NeRF with sample reduction (DONeRF) and a NeRF with alternative representations (TensoRF).\footnote{\label{foot:project_page}The project page is available at \url{https://www.kecl.ntt.co.jp/people/kaneko.takuhiro/projects/mimo-nerf/}.}
\end{abstract}

\section{Introduction}
\label{sec:introduction}

Images are two-dimensional (2D) projections of three-dimensional (3D) scenes.
Solving the inverse problem, that is, learning 3D representations from 2D images and synthesizing novel views, is a fundamental concern in computer vision and graphics and has been extensively studied for various applications such as photo editing, content creation, virtual reality, and environmental understanding.

With the advent of implicit neural representations (e.g.,~\cite{VSitzmannNeurIPS2019,MNiemeyerCVPR2020,BMildenhallECCV2020,CHLinNeurIPS2020,LYarivNeurIPS2020,AYuCVPR2021,VSitzmannNeurIPS2021}), substantial advancements have been made towards addressing this problem.
Neural radiance fields (NeRFs)~\cite{BMildenhallECCV2020} have been noted as a successful approach. A NeRF represents a scene using a continuous function that maps 3D coordinates and view direction to the color and volume density and renders a pixel by integrating the outputs on a ray using volume rendering~\cite{NMaxTVCG1995}.
This formulation enables a NeRF to learn to synthesize geometrically consistent and high-fidelity novel views with only 2D supervision.

\begin{figure}[t]
  \begin{center}
    \includegraphics[width=\columnwidth]{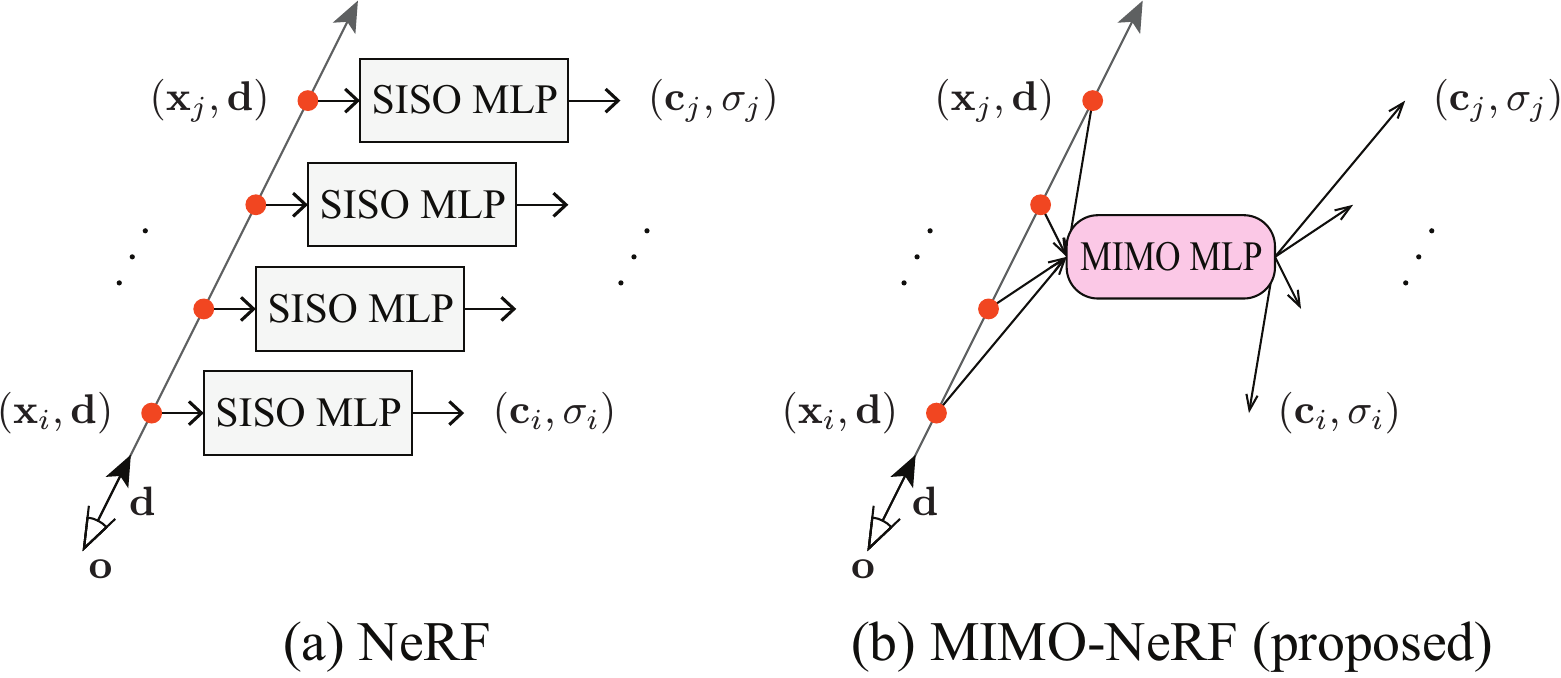}
  \end{center}
  \caption{Comparison between NeRF and MIMO-NeRF (proposed).
    (a) A typical NeRF uses a SISO MLP that maps 3D coordinates and view direction to the color and volume density in a \textit{sample-wise} manner.
    (b) In contrast, the proposed MIMO-NeRF uses a MIMO MLP that performs mappings in a \textit{group-wise} manner.
    This change reduces the number of MLPs running and improves the rendering speed, but also requires addressing ambiguity in the color and volume density caused by the fact that these values are determined in a non-unique manner by a set of input coordinates that vary by viewpoint, grouping, and sampling.
    The main technical contribution of the present work is that of providing methods to mitigate this challenge.
    We demonstrate the impact of the proposed technique in Figure~\ref{fig:challenge}.}
  \label{fig:concept}
\end{figure}

Despite this advantage, a typical NeRF suffers from slow rendering because it uses a \textit{single-input single-output (SISO)} MLP that calculates the RGB color and volume density in a \textit{sample-wise} manner (Figure~\ref{fig:concept}(a)).
Although this architecture ensures the independent representation of each point, which is useful, for example, for learning view-independent volume density, its computational cost increases in proportion to the number of samples for each ray (e.g., on the order of hundreds).
Several methods developed to address this issue can be roughly categorized into two approaches, including (1) \textit{sample reduction} and (2) \textit{alternative representations}.

A typical sample reduction strategy reduces the number of samples on a ray using a sampling network based on the depth~\cite{TNeffCVF2021} or density of a pretrained NeRF~\cite{MPiala3DV2021} or using a sampling network with an adaptive optimization mechanism~\cite{DLindellCVPR2021,JFangArXiv2021,AKurzECCV2022}.
These methods successfully accelerate the rendering process while retaining image quality adequately.
However, most of these techniques still use a SISO MLP to predict the colors and volume densities of selected samples; therefore, they still need to run MLPs many times in proportion to the number of selected samples.
This issue can be alleviated by reducing the number of selected samples, although this deteriorates the quality of the synthesized images accordingly.

As alternative representations, various sophisticated and faster representations such as 3D voxel grids~\cite{SGarbinICCV2021,PHedmanICCV2021,LLiuNeurIPS2020,LWuCVPR2022,CSunCVPR2022}, sparse voxel-based octrees~\cite{AYu2021ICCV,SFridovichCVPR2022}, multiplane images~\cite{SWizadwongsaCVPR2021}, tri-planes~\cite{EChanCVPR2022}, vector-matrix decomposition~\cite{AChenECCV2022}, hashes~\cite{TMullerTOG2022}, NeRF-specific structures~\cite{THuCVPR2022}, and space-wise MLPs~\cite{DRebainCVPR2021,CReiserICCV2021} have been devised.
These representations contribute to achieving fast rendering while retaining image quality moderately well.
However, after powerful features are extracted using alternative representations, SISO MLPs are still commonly used for the final prediction of color or volume density owing to their memory-efficient and continuous nature.
Hence, part of the calculation cost still increases depending on the number of samples.

Consequently, owing to its compact, continuous (i.e., resolution-free), and independent (e.g., view-independent) nature, a SISO MLP is commonly used in various NeRFs.
However, as mentioned above, the calculation cost increases with the number of samples.
This is not preferable when considering the improvement in rendering speed.
Possible simple solutions include, for example, a reduction of the number of samples or a reduction of the size of a model with a corresponding sacrifice of image quality.
However, these solutions are not necessarily the best for handling the trade-off between quality and speed.\footnote{We discuss this trade-off in detail in Section~\ref{subsec:investigation_tradeoff}.}
Alternatively, we propose a \textit{multi-input multi-output NeRF (MIMO-NeRF)}, which is a novel variant of NeRF that represents a scene using a \textit{MIMO} MLP that conducts mappings in a \textit{group-wise} manner (Figure~\ref{fig:concept}(b)).
This modification enables a reduction in the number of MLPs running according to the number of grouped samples and consequently improves rendering speed.

\begin{figure}[t]
  \begin{center}
    \includegraphics[width=\columnwidth]{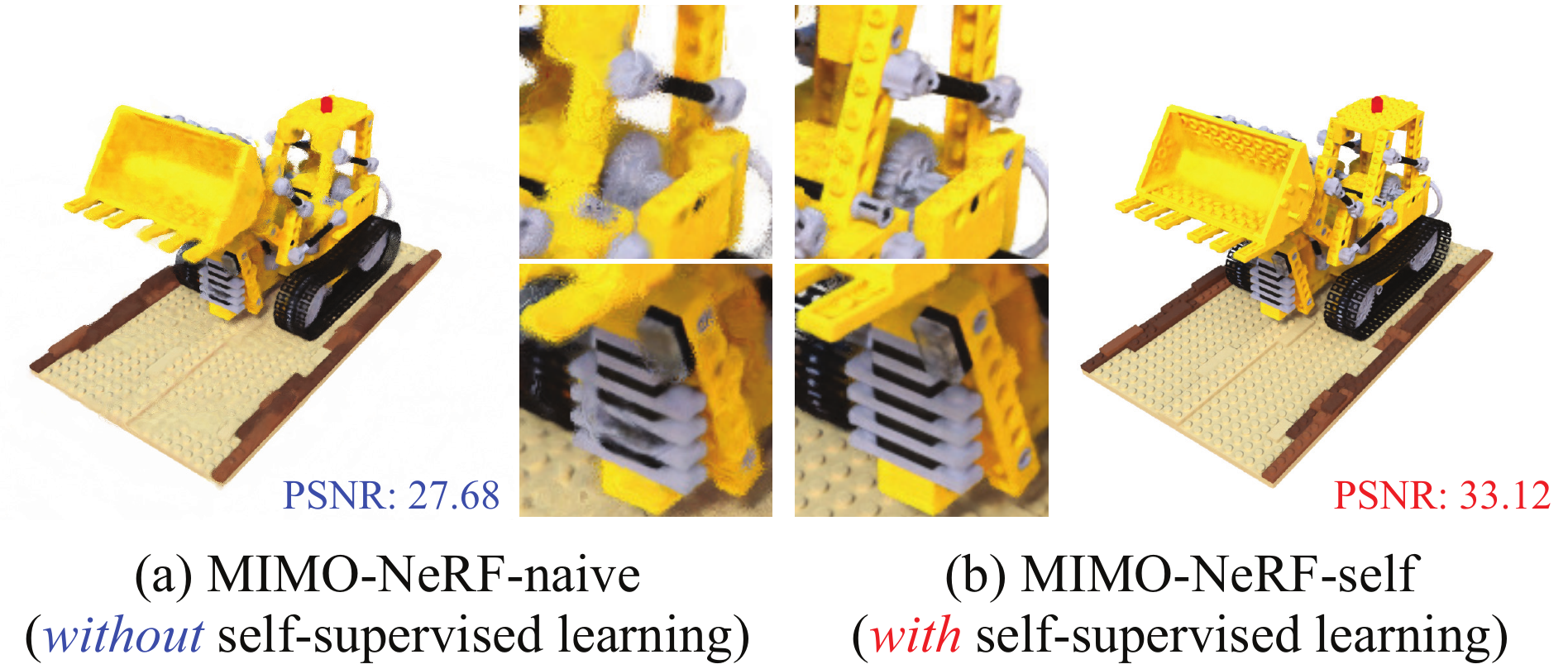}
  \end{center}
  \caption{Challenge of training MIMO-NeRF and the impact of the proposed self-supervised learning.
    (a) MIMO-NeRF-naive suffers from ambiguity in the color and volume density of each point and deteriorates image quality.
    (b) We propose self-supervised learning to address this problem without relying on pretrained models.
    This mitigates fluctuation artifacts and improves image quality.}
  \label{fig:challenge}
\end{figure}

However, in this approach, the uniqueness of the color and volume density of each point is not ensured because they are determined not only by the coordinates of the corresponding point but also by the coordinates of the other points in a group, which vary by viewpoint, grouping, and sampling.
This leads to some \textit{ambiguity} and causes fluctuation artifacts as shown in Figure~\ref{fig:challenge}(a).
In particular, this ambiguity can be problematic when learning a 3D representation using only 2D supervision because obtaining direct supervision that can resolve the ambiguity is difficult.
One possible solution is to train a standard (i.e., SISO) NeRF first and then distill the model onto the corresponding MIMO-NeRF.
However, this increases training time because both student and teacher NeRFs must be trained.
Alternatively, we have also developed a novel \textit{self-supervised learning} approach in which we reformulate a MIMO MLP in several ways (in particular, we use \textit{group shift} (Figure~\ref{fig:group_shift}) and \textit{variation reduction} (Figure~\ref{fig:variation_reduction})) and impose a consistent regularization so that the reformulated MIMO MLPs produce the same outputs.
Because each reformulated MIMO MLP can render a pixel faster than the original SISO MLP, we can prevent a large sacrifice of training time even when using multiple reformulated MIMO MLPs by adequately adjusting the reformulation configuration.
Figure~\ref{fig:challenge}(b) shows an example of the effects of this learning.

We investigated the benchmark performance of MIMO-NeRF by comparing it with possible alternatives (including the distillation of a pretrained NeRF, reduction of the number of samples, and reduction of the model size).
We also performed ablation studies to examine the validity of each component of the proposed self-supervised learning method.
We apply MiMO-NeRF to two representative fast NeRFs, including a NeRF with sample reduction (DONeRF~\cite{TNeffCVF2021}) and a NeRF with alternative representations (TensoRF~\cite{AChenECCV2022}) to demonstrate that it is compatible with and complementary to previous advancements in NeRFs.

The main contributions of this study are summarized as follows.
\begin{itemize}
  \setlength{\parskip}{1.5pt}
  \setlength{\itemsep}{1.5pt}
\item To speed up the rendering of NeRF, we propose \textit{MIMO-NeRF}, which represents a scene using a \textit{MIMO} MLP that maps the coordinates on a ray to the colors and volume densities in a \textit{group-wise} manner.
\item We introduce novel \textit{self-supervised learning} to mitigate the ambiguity in the color and volume density of each point and enable MIMO-NeRF to be trained without relying on pretrained models.
\item We examined the effectiveness of MIMO-NeRF through a comprehensive experimental evaluation, and the results demonstrate the versatility of MIMO-NeRF in applications to two representative fast NeRFs.
  We also provide more detailed analyses and extended results in the Appendix~\ref{sec:further_analyses} and on the \href{https://www.kecl.ntt.co.jp/people/kaneko.takuhiro/projects/mimo-nerf/}{project page}.\footnoteref{foot:project_page}
\end{itemize}

\section{Related work}
\label{sec:related_work}

\noindent\textbf{Implicit neural representations.}
Implicit neural representations have attracted attention in 3D shape~\cite{JParkCVPR2019,LMeschederCVPR2019,ZChenCVPR2019,MMichalkiewiczICCV2019,SSaitoICCV2019,KGenovaICCV2019,MAtzmonCVPR2020,AGroppICML2020} and 3D scene~\cite{CJiangCVPR2020,SPengECCV2020,RChabraECCV2020,JChibaneNeurIPS2020,VSitzmannNeurIPS2020} reconstructions owing to their memory-efficient, continuous (i.e., resolution-free), and 3D-aware characteristics.
Although research began with explicit 3D supervision, learning implicit 3D only from 2D supervision (i.e., inverse graphics) has also been achieved by incorporating differentiable rendering~\cite{VSitzmannNeurIPS2019,SLiuNeurIPS2019,MNiemeyerCVPR2020,SLiuCVPR2020,BMildenhallECCV2020,CHLinNeurIPS2020,LYarivNeurIPS2020,AYuCVPR2021,VSitzmannNeurIPS2021}.
In this study, we focus on NeRFs as a representative example of the latter owing to their remarkable success in synthesizing geometrically consistent and high-quality novel view.
However, applying our ideas to other implicit neural representations, such as those mentioned above, remains as an interesting direction for future research.

\smallskip\noindent\textbf{Advancements in NeRFs.}
Various extensions have been proposed since the emergence of NeRFs.
For example, representative research topics include (1) improving image quality and enhancing applicable scenes (e.g.,~\cite{KZhangArXiv2020,RMartinCVPR2021,APumarolaCVPR2021,GGafniCVPR2021,WXianCVPR2021,JBarronICCV2021,JBarronCVPR2022,BMildenhallCVPR2022,DVerbinCVPR2022,XChenCVPR2022}), (2) incorporation into other models, e.g., deep generative models, such as generative adversarial networks (GANs)~\cite{IGoodfellowNIPS2014} and diffusion probabilistic models~\cite{YSongNeurIPS2019,JHoNeurIPS2020} (e.g.,~\cite{KSchwarzNeurIPS2020,EChanCVPR2021,MNiemeyerCVPR2021,MNiemeyer3DV2021,JGuICLR2022,EChanCVPR2022,YDengCVPR2022,YXueCVPR2022,TKanekoCVPR2022,ISkorokhodovNeurIPS2022,BPooleICLR2023}), and (3) accelerating NeRFs for fast inference or fast training (e.g.,~\cite{TNeffCVF2021,MPiala3DV2021,DLindellCVPR2021,JFangArXiv2021,AKurzECCV2022,SGarbinICCV2021,PHedmanICCV2021,LLiuNeurIPS2020,LWuCVPR2022,CSunCVPR2022,AYu2021ICCV,SFridovichCVPR2022,SWizadwongsaCVPR2021,EChanCVPR2022,AChenECCV2022,TMullerTOG2022,THuCVPR2022,DRebainCVPR2021,CReiserICCV2021}).
The present work falls into the third category.
However, our proposed approach is complementary to previous studies, including most of the abovementioned works in all categories, because SISO MLPs have commonly been used as a partial or main network in previous studies and improving their rendering speed by replacing SISO MLPs with the proposed MIMO MLP is feasible.
The results of the experimental evaluation validated this potential (Sections~\ref{subsec:application_donerf} and \ref{subsec:application_tensorf}).

\smallskip\noindent\textbf{Acceleration of NeRFs.}
As discussed in Section~\ref{sec:introduction}, a typical NeRF is well known for its slow rendering because it uses a SISO MLP.
Several approaches have been developed to address this issue.
These can be roughly categorized into two approaches, including (1) \textit{sample reduction} and (2) \textit{alternative representations}.
(1) As described in Section~\ref{sec:introduction}, methods that reduce the number of samples using a sampling network can improve rendering speed by replacing a SISO MLP with the proposed MIMO MLP.
We validated this statement during an experiment (Section~\ref{subsec:application_donerf}) by applying our ideas to a representative NeRF in this category (DONeRF~\cite{TNeffCVF2021}).
Another common approach in the first category is to render pixels using a light-field network~\cite{VSitzmannNeurIPS2021,BAttalCVPR2022,MSuhailCVPR2022,HWangECCV2022} instead of volume rendering~\cite{NMaxTVCG1995}.
This approach has been shown to achieve fast rendering by running only a single MLP for a given ray.
However, owing to the lack of explicit geometry-aware representations driven by the use of volume densities, these methods suffer from limitations that are not faced by a standard NeRF in terms of restrictions on applicable scenes (e.g., toy datasets~\cite{VSitzmannNeurIPS2021} and forward-facing datasets~\cite{BAttalCVPR2022}), a requirement for high-capacity models (e.g., a deeper MLP~\cite{HWangECCV2022} and a transformer~\cite{MSuhailCVPR2022}), and the need for extra modules (e.g., meta-learned priors~\cite{VSitzmannNeurIPS2021}, pretrained NeRFs~\cite{BAttalCVPR2022,HWangECCV2022}, or additional encoders~\cite{MSuhailCVPR2022}).
(2) As explained in Section~\ref{sec:introduction}, alternative representations have the potential to accelerate the rendering speed by replacing the SISO MLP with the proposed MIMO MLP.
We present an empirical investigation of this potential in Section~\ref{subsec:application_tensorf} by incorporating MIMO-NeRF into TensoRF~\cite{AChenECCV2022}, a representative model in this category.

\smallskip\noindent\textbf{Learning of fast NeRFs.}
Knowledge distillation (or baking) methods are commonly used to train fast NeRFs. In these methods, a standard NeRF is first trained and then baked to faster representations~\cite{SGarbinICCV2021,PHedmanICCV2021,THuCVPR2022,DRebainCVPR2021,CReiserICCV2021}.
However, this approach is disadvantageous in terms of training time because two separate models must be trained, i.e., teacher and student NeRFs.
As an alternative, we consider a self-supervised learning approach in which we can train a model without a large increase in training time.
We examine the performance differences between the proposed self-supervised learning scheme and a knowledge distillation scheme in Section~\ref{subsec:investigation_performance}.

\section{Preliminaries: NeRF}
\label{sec:preliminaries}

We begin by explaining NeRFs as the basis for our model.
As shown in Figure~\ref{fig:concept}(a), a NeRF represents a point in a 3D space using a continuous SISO function $f_{\text{SISO}}$ that maps the 3D position $\mathbf{x} \in \mathbb{R}^3$ and view direction $\mathbf{d} \in \mathbb{S}^2$ to the RGB color $\mathbf{c}(\mathbf{x}, \mathbf{d}) \in \mathbb{R}^3$ and volume density $\sigma(\mathbf{x}) \in \mathbb{R}^+$ in a sample-wise manner.
\begin{flalign}
  \label{eq:siso}
  f_{\text{SISO}}: \mathbb{R}^3 \times \mathbb{S}^2 \rightarrow \mathbb{R}^3 \times \mathbb{R}^+, \:\: (\mathbf{x}, \mathbf{d}) \mapsto (\mathbf{c}, \sigma).
\end{flalign}
Specifically, positional encoding~\cite{BMildenhallECCV2020,MTancikNeurIPS2020} is applied to $\mathbf{x}$ and $\mathbf{d}$ to represent the high-frequency details of an image.
Subsequently, an MLP is applied to the encoded inputs to obtain $\mathbf{c}$ and $\sigma$.
For simplicity, we represent these series of processes in a unified manner as $f_{\text{SISO}}$.

A NeRF is based on ray tracing, in which a camera ray is defined as $\mathbf{r}(t) = \mathbf{o} + t \mathbf{d}$,
where $\mathbf{o}$ and $\mathbf{d}$ respectively denote the origin and direction of the camera and $t$ denotes a distance from the origin.
A NeRF calculates the color of each pixel $\mathbf{\hat{C}}(\mathbf{r})$ by integrating the colors and volume densities on a ray $\mathbf{r}(t)$ within $t \in [t_n, t_f]$ using volume rendering~\cite{NMaxTVCG1995}.
In implementation, the calculation of the integral is intractable; therefore, a ray is discretized into $N$ points; alternatively, the following discretized formulation can be used.
\begin{flalign}
  \label{eq:volume_rendering}
  \mathbf{\hat{C}}(\mathbf{r}) = \sum_{i = 1}^N T_i \alpha_i \mathbf{c}_i,
  \text{where } T_i = \prod_{j = 1}^{i - 1} (1 - \alpha_j),
\end{flalign}
where the subscript $i$ indicates that the variable corresponds to the $i$-th point on a ray, $\alpha_i = 1 - \exp(-\sigma_i \delta_i)$ is an alpha value, and $\delta_i = t_{i+1} - t_i$ is the distance between the $i$-th and $(i+1)$-th points.
$f_{\text{SISO}}$ is optimized by minimizing the following pixel-wise loss.
\begin{flalign}
  \label{eq:pixel_loss}
  \mathcal{L}_{\text{pixel}} = \| \mathbf{\hat{C}}(\mathbf{r}) - \mathbf{C}(\mathbf{r}) \|_2^2,
\end{flalign}
where $\mathbf{C}(\mathbf{r})$ is the ground-truth color for a ray $\mathbf{r}$.
In implementation, this loss is calculated for $\mathbf{r} \in \mathcal{R}$, where $\mathcal{R}$ denotes a set of rays in each batch.

In practice, a NeRF uses coarse and fine networks.
In the coarse network, a ray is discretized into $N_c$ points using stratified sampling, whereas in the fine network, a ray is discretized into $N_c + N_f$ points using hierarchical sampling in which $N_f$ additional points are sampled according to the output of the coarse network.
The two networks are optimized by minimizing $\mathcal{L}_{\text{pixel}}$ for $\mathbf{\hat{C}}(\mathbf{r})$ predicted by each network.
Hereafter, we omit a variable in parentheses (e.g., $(\mathbf{r})$) for simplicity.

\section{MIMO-NeRF}
\label{sec:method}

\subsection{MIMO formulation}
\label{subsec:formulation}

There are several ways to group the input samples when constructing a MIMO MLP.
For example, we can construct a general MLP that can accept any combination of samples in a 3D space, or we can construct a specific MLP that only accepts a group of nearby samples.
In preliminary experiments (Appendix~\ref{subsec:effect_grouping}), we found that the latter significantly outperformed the former because general models are more difficult to train than more specific models.
Therefore, we adopted the latter in this study.
In particular, we group neighboring samples on a ray as shown in Figure~\ref{fig:concept}(b).\footnote{One possible alternative is to group near samples on different rays.
  However, in NeRFs, searching near points across different rays is not trivial because points are sampled unevenly via hierarchical sampling.
  Therefore, we simply group neighboring samples on the same ray in this study.}

More formally, given $N$ samples on a ray, we group $N_p$ samples from the sample nearest to the camera and create $N / N_p$ groups.
Subsequently, we apply a MIMO function $f_{\text{MIMO}}$ to each group as follows.
\begin{flalign}
  \label{eq:mimo}
  f_{\text{MIMO}}: (\mathbb{R}^3)^{N_p} \times \mathbb{S}^2 &\rightarrow (\mathbb{R}^3 \times \mathbb{R}^+)^{N_p},
  \nonumber \\
  (\mathbf{x}_i, \dots, \mathbf{x}_j, \mathbf{d}) &\mapsto (\mathbf{c}_i, \dots, \mathbf{c}_j, \sigma_i, \dots, \sigma_j),
\end{flalign}
where $i \in \{ 1, 1 + N_p, \dots, 1 + N - N_p \}$ and $j = i + N_p - 1$.
Assuming that the grouped samples are lined on a ray, we use a single direction $\mathbf{d}$ in the input of $f_\text{MIMO}$.
In this formulation, the number of MLPs running to render a pixel (\textit{\# Run}) is equal to the number of groups and is calculated as $\text{\# Run} = N / N_p$.
Therefore, we can reduce the calculation cost, particularly that of \#~Run, by increasing $N_p$.

In inference, the only necessary modification to a NeRF is the simple replacement of $f_{\text{SISO}}$ with $f_{\text{MIMO}}$ and the same volume rendering (Equation~\ref{eq:volume_rendering}) and sampling scheme (i.e., stratified and hierarchical sampling) can be used.
With this modification strategy, MIMO-NeRF exhibits high compatibility and complementarity with previous NeRFs.

During training, $\mathcal{L}_{\text{pixel}}$ (Equation~\ref{eq:pixel_loss}) can be used for $\mathbf{\hat{C}}$ obtained using $f_{\text{MIMO}}$.
However, this loss does not necessarily suffice to address the ambiguity in the color and volume density of each point because this ambiguity occurs in a 3D space and the loss cannot regularize the 3D representations explicitly.
Hence, we introduce self-supervised learning as discussed in subsequent sections.

\subsection{MIMO reformulation}
\label{subsec:reformulation}

One possible solution to this ambiguity is to train a standard (i.e., SISO) NeRF first and then distill the model onto a corresponding MIMO-NeRF.
However, this solution involves an increase in training time because it requires training not only a MIMO-NeRF but also a SISO NeRF.
Consequently, this solution cannot entirely take advantage of the fast rendering of MIMO-NeRF in training.

Alternatively, we reformulate a MIMO MLP in multiple ways and impose a consistent regularization to produce the same RGB colors and alpha values.
Because each reformulated MIMO MLP can render a pixel faster than the original SISO MLP, we can prevent a large increase in training time even when using multiple reformulated MIMO MLPs by adequately adjusting the reformulation configuration.
When implementing this idea, the question arises as to how best to reformulate a MIMO MLP.
To this end, we developed two methods, including (1) \textit{group shift} and (2) \textit{variation reduction}.

\begin{figure}[t]
  \begin{center}
    \includegraphics[width=\columnwidth]{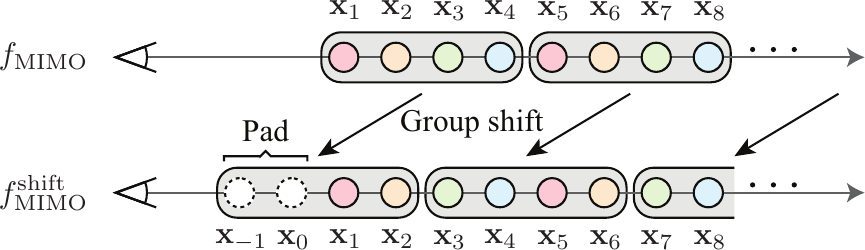}
  \end{center}
  \caption{Example of the group shift when $N_p = 4$ and $s = 2$.
    We shift each group by $s$ toward the camera after padding $s$ samples before the front sample.
    This procedure enables assessing each sample in multiple ways using different groups.}
  \label{fig:group_shift}
\end{figure}

\smallskip\noindent\textbf{Group shift.}
We consider restricting this ambiguity by assessing each point in multiple ways using different groups and imposing consistency on the assessed results.
More formally, we implement this by shifting groups and rewriting Equation~\ref{eq:mimo} as follows.
\begin{flalign}
  \label{eq:mimo_shift}
  f_{\text{MIMO}}^{\text{shift}}: (\mathbb{R}^3)^{N_p} \times \mathbb{S}^2 &\rightarrow (\mathbb{R}^3 \times \mathbb{R}^+)^{N_p},
  \nonumber \\
  (\mathbf{x}_{i'}, \dots, \mathbf{x}_{j'}, \mathbf{d}) &\mapsto (\mathbf{c}_{i'}, \dots, \mathbf{c}_{j'}, \sigma_{i'}, \dots, \sigma_{j'}),
\end{flalign}
where $i' \in \{ k, k + N_p, \dots, k + N \}$ and $j' = i' + N_p - 1$.
Here, $k$ is the head index of the first group, which is shifted by $s \in \{ 1, \dots, N_p - 1 \}$ toward the camera.
Hence, $k \in \{ 2 - N_p, \dots, 0 \}$.
In practice, it is randomly sampled during training.
More strictly, we add padding before this process to represent a sample with an index exceeding the original index, i.e., $i' < 1$ or $j' > N$.
For clarity, we present an example in which $N_p = 4$ and $s = 2$ in Figure~\ref{fig:group_shift}.

\begin{figure}[t]
  \begin{center}
    \includegraphics[width=\columnwidth]{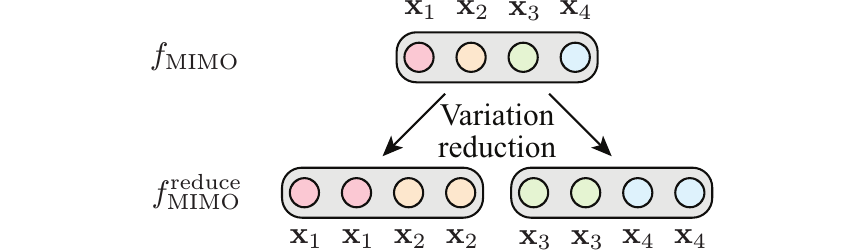}
  \end{center}
  \caption{Example of variation reduction when $N_p = 4$ and $R = 2$.
    Samples with the same color correspond to the same coordinate.
    In $f_{\text{MIMO}}^{\text{reduce}}$, we reduce the number of unique samples in a group from $N_p$ $(= 4)$ to $\frac{N_p}{R}$ $(= 2)$ by repeating each sample $R$ $(= 2)$ times to reduce the variation in a group.}
  \label{fig:variation_reduction}
\end{figure}

\smallskip\noindent\textbf{Variation reduction.}
In the original MIMO formulation, the abovementioned ambiguity is caused by $N_p$ different input samples.
To mitigate this, we consider reducing the variation in the input by replacing Equation~\ref{eq:mimo} with the following.
\begin{flalign}
  \label{eq:mimo_reduce}
  f_{\text{MIMO}}^{\text{reduce}}: (\mathbb{R}^3)^{N_p} \times \mathbb{S}^2 &\rightarrow (\mathbb{R}^3 \times \mathbb{R}^+)^{N_p},
  \nonumber \\
  ([\mathbf{x}_{i''}]^{R}, \dots, [\mathbf{x}_{j''}]^{R}, \mathbf{d}) &\mapsto
  (\mathbf{c}_{i''_1}, \dots, \mathbf{c}_{i''_R}, \dots, \mathbf{c}_{j''_1}, \dots, \mathbf{c}_{j''_R},
  \nonumber \\ &
  \sigma_{i''_1}, \dots, \sigma_{i''_R}, \dots, \sigma_{j''_1}, \dots, \sigma_{j''_R}),
\end{flalign}
where $[\cdot]^R$ denotes the operation of repeating the given variable $R$ times, $i'' \in \{ 1, 1 + \frac{N_p}{R}, \dots, 1 + N - \frac{N_p}{R} \}$, and $j'' = i'' + \frac{N_p}{R} - 1$.
After applying $f_{\text{MIMO}}^{\text{reduce}}$, we average $(\mathbf{c}_{k''_1}, \dots, \mathbf{c}_{k''_R})$ and $(\sigma_{k''_1}, \dots, \sigma_{k''_R})$ to obtain $\mathbf{c}_{k''}$ and $\sigma_{k''}$, where $k'' \in \{ i'', \dots, j'' \}$.
In this formulation, the mentioned ambiguity is reduced by decreasing input variation from $N_p$ to $\frac{N_p}{R}$.
Conversely, the number of MLPs running increases by factor of $R$, i.e., $\text{\# Run} = N / N_p \times R$; therefore, we need to select $R$ carefully in practice to avoid a large increase in training time.
For clarity, we present an example case in which $N_p = 4$ and $R = 2$ in Figure~\ref{fig:variation_reduction}.

\subsection{MIMO objective}
\label{subsec:objective}

Through the above processes, we obtain $M$ reformulated MIMO MLPs in which we use different $s$ for each MIMO MLP and set $R$ such that the total number of \#~Run is not larger than that of the original SISO MLP.
Hereafter, we use the superscript $m \in \{ 1, \dots, M \}$ to denote the variable corresponding to the $m$-th reformulated MIMO MLP, e.g., $\mathbf{\hat{C}}^m$ and $R^m$.
We train these MLPs using two loss functions, including pixel-wise and 3D consistency losses.

\smallskip\noindent\textbf{Pixel-wise loss.}
We apply the pixel-wise loss (Equation~\ref{eq:pixel_loss}) to each $\mathbf{\hat{C}}^m$ as follows.
\begin{flalign}
  \label{eq:mimo_pixel_loss}
  \mathcal{L}_{\text{pixel}}^{\text{MIMO}} = \sum_{m = 1}^M \| \mathbf{\hat{C}}^m - \mathbf{C} \|_2^2.
\end{flalign}

\smallskip\noindent\textbf{3D consistency loss.}
The pixel-wise loss provides supervision in a 2D space; however, it cannot impose an explicit regularization in a 3D space.
Hence, we introduce a 3D consistency loss that encourages the reformulated MIMO MLPs to produce the same colors and alpha values in the 3D space.
The 3D consistency loss consists of a color 3D consistency loss $\mathcal{L}_{\text{3D}}^{\text{color}}$ and an alpha value 3D consistency loss $\mathcal{L}_{\text{3D}}^{\text{alpha}}$ as follows.
\begin{flalign}
  \label{eq:color_loss}
  \mathcal{L}_{\text{3D}}^{\text{color}} = \sum_{m_1 = 1}^{M - 1} \sum_{m_2 = m_1 + 1}^M \frac{1}{N} \sum_{i=1}^N [ \mu_{m_1}^{m_2} \| \mathbf{c}_i^{m_1} - \text{sg}(\mathbf{c}_i^{m_2}) \|_2^2
  \nonumber \\
  + \mu_{m_2}^{m_1} \| \text{sg}(\mathbf{c}_i^{m_1}) - \mathbf{c}_i^{m_2} \|_2^2 ],
  \\
  \label{eq:alpha_loss}
  \mathcal{L}_{\text{3D}}^{\text{alpha}} = \sum_{m_1 = 1}^{M - 1} \sum_{m_2 = m_1 + 1}^M \frac{1}{N} \sum_{i=1}^N [ \mu_{m_1}^{m_2} \| \alpha_i^{m_1} - \text{sg}(\alpha_i^{m_2}) \|_2^2
  \nonumber \\
  + \mu_{m_2}^{m_1} \| \text{sg}(\alpha_i^{m_1}) - \alpha_i^{m_2} \|_2^2 ].
\end{flalign}
where ``$\text{sg}$'' indicate a stop-gradient operation.
The 3D consistency loss $\mathcal{L}_{\text{3D}}$ is calculated by $\mathcal{L}_{\text{3D}} = \mathcal{L}_{\text{3D}}^{\text{color}} + \mathcal{L}_{\text{3D}}^{\text{alpha}}$.
We define $\mu_{m_i}^{m_j}$ as $\mu_{m_i}^{m_j} = \frac{\sqrt{R^{m_j}}}{\sqrt{R^{\text{max}}} \sqrt{R^{m_i}}}$, where $R^{\text{max}}$ is the maximum of $R^m$ in $m \in \{ 1, \dots, M \}$.
We use this asymmetric weight on the assumption that $\mathbf{c}_i^{m}$ and $\alpha_i^{m}$ with greater $R^m$ have lower ambiguity and are more reliable.
Hence, the effect of $\mathcal{L}_{\text{3D}}$ is reduced.
We empirically investigated the importance of this effect through an ablation study as described in Section~\ref{subsec:ablation_studies}.

\smallskip\noindent\textbf{Full objective.}
The full objective is defined as follows.
\begin{flalign}
  \label{eq:mimo_objective}
  \mathcal{L}_{\text{MIMO}} = \mathcal{L}_{\text{pixel}}^{\text{MIMO}} + \lambda \mathcal{L}_{\text{3D}},
\end{flalign}
where $\lambda$ is a hyperparameter that balances the pixel-wise loss and 3D consistency loss.

\section{Experiments}
\label{sec:experiments}

We conducted five experiments to investigate the effectiveness of MIMO-NeRF.
In the first three experiments, we conducted a comprehensive study, including an investigation of benchmark performance (Section~\ref{subsec:investigation_performance}), an investigation of the trade-off between speed and quality (Section~\ref{subsec:investigation_tradeoff}), and ablation studies (Section~\ref{subsec:ablation_studies}).
In the remaining two experiments, we examined the versatility of MIMO-NeRF by applying it to two representative fast NeRFs, including a NeRF with sample reduction, i.e., DONeRF~\cite{TNeffCVF2021} (Section~\ref{subsec:application_donerf}), and a NeRF with alternative representations, i.e., TensoRF~\cite{AChenECCV2022} (Section~\ref{subsec:application_tensorf}).
The main results of these experiments are provided here, and detailed analyses are presented with extended results in Appendix~\ref{sec:further_analyses}.
The implementation details are presented in Appendix~\ref{sec:implementation_details}.

\subsection{Investigation of benchmark performance}
\label{subsec:investigation_performance}

We investigated the benchmark performance of MIMO-NeRF by applying our ideas to the original \textit{NeRF}~\cite{BMildenhallECCV2020}.
In particular, we examined three variants of MIMO-NeRF, including
\textit{MIMO-NeRF-naive}, which simply replaced $f_{\text{SISO}}$ (Equation~\ref{eq:siso}) with $f_{\text{MIMO}}$ (Equation~\ref{eq:mimo}) and was trained with a standard pixel-wise loss (Equation~\ref{eq:pixel_loss}).\footnote{For simplicity and a fair comparison, we only increased the input and output of the original SISO MLP and retained the other parameters (e.g., depth and width).
  Hence, the increase in model size was relatively small.}
\textit{MIMO-NeRF-distill}, which is a student model distilled from a pretrained standard (i.e., SISO) NeRF.
During training, we used a 3D consistency loss (Equations~\ref{eq:color_loss} and \ref{eq:alpha_loss}) that was adjusted for knowledge distillation, in addition to a standard pixel-wise loss (Equation~\ref{eq:pixel_loss}).
\textit{MIMO-NeRF-self}, which is MIMO-NeRF that adopted the proposed self-supervised learning.
We examined the performance of these models when $N_p$ was varied within $\{ 2, 4, 8 \}$.

\smallskip\noindent\textbf{Datasets.}
We investigated the benchmark performance on two commonly-used datasets.
(1) \textit{Blender dataset}~\cite{BMildenhallECCV2020} includes eight scenes, each of which consists of $360^\circ$ views of complex objects at a resolution of $800 \times 800$ pixels.
We used $100$ and $200$ views for training and testing, respectively.
(2) \textit{Local Light Field Fusion (LLFF) dataset}~\cite{BMildenhallTOG2019,BMildenhallECCV2020}, which consists of eight complex real-world scenes, each of which includes $20$--$62$ forward-facing views at $1008 \times 756$ pixels.
One-eighth of the images were used for testing, and the remainder were used for training.
Where not otherwise specified, we used half-sized images following the default settings of a widely-used source code for NeRF\footnote{\label{foot:nerf-pytorch}\url{https://github.com/yenchenlin/nerf-pytorch}} to better investigate the various configurations.

\smallskip\noindent\textbf{Implementation.}
For a fair comparison, we implemented all the models with a commonly-used source code for NeRF\footnoteref{foot:nerf-pytorch} and trained the models using the default settings provided in the code.
The number of samples was set as $N_c = 64$ and $N_f = 128$ for the Blender dataset and $N_c = 64$ and $N_f = 64$ for the LLFF dataset.
For MIMO-NeRF-self, we used two formulations with $R^1 = R^2 = 1$ when $N_p = 2$,
two formulations with $R^1 = 1$ and $R^2 = 2$ when $N_p = 4$,
and three formulations with $R^1 = 1$, $R^2 = 2$, and $R^3 = 4$ when $N_p = 8$.
MIMO-NeRF-self was trained individually depending on $N_p$.
An investigation of different reformulation methods is presented in Appendix~\ref{subsec:effect_reformulation}.
Group shifts were applied to all cases.
We set $\lambda$ to $1$ and $0.4$ for the Blender and LLFF datasets, respectively.
The effect of $\lambda$ is analyzed in Appendix~\ref{subsec:effect_hyperparameter}.
The implementation details are presented in Appendix~\ref{subsec:implementation_details_nerf}.

\smallskip\noindent\textbf{Evaluation metrics.}
Following the original NeRF study~\cite{BMildenhallECCV2020}, we used the peak signal-to-noise ratio (\textit{PSNR}), structural similarity index (\textit{SSIM})~\cite{ZWangTIP2004}, and learned perceptual image patch similarity (\textit{LPIPS})~\cite{RZhangCVPR2018} to quantitatively evaluate the image quality.
To assess the calculation cost of inference and training, we report inference time (\textit{I-time}) measured with an NVIDIA GeForce RTX 3080 Ti GPU and training time (\textit{T-time}) measured with an NVIDIA A100-SXM4-80GB GPU.\footnote{For simplicity and a fair comparison, we measured the calculation time using a standard PyTorch implementation\footnoteref{foot:nerf-pytorch} for all the models.
  Optimizing the implementation for faster rendering (e.g., using custom CUDA kernels) would be interesting for future research.}
We also provide \textit{\#~Run} $\left( = \frac{N_c + (N_c + N_f)}{N_p} \right)$ and the number of parameters (\textit{\#~Params}) as supplementary information.
\#~Params increases in MIMO-NeRF mainly because the total dimension of the positional embeddings is increased by $N_p$ times according to the increase in the inputs.

\begin{table*}
  \setlength{\tabcolsep}{0.5pt}
  \begin{center}
    \scriptsize
    \begin{tabularx}{\textwidth}{lc|CCCCCCC|CCCCCCC}
      & & \multicolumn{7}{c|}{Blender} & \multicolumn{7}{c}{LLFF}
      \\
      \multicolumn{1}{c}{Model} & $N_p$ & PSNR$\uparrow$ & SSIM$\uparrow$ & LPIPS$\downarrow$ & \# Run$\downarrow$ & I-time$\downarrow$ & T-time$\downarrow$ & \# Params & PSNR$\uparrow$ & SSIM$\uparrow$ & LPIPS$\downarrow$ & \# Run$\downarrow$ & I-time$\downarrow$ & T-time$\downarrow$ & \# Params \\
      & & & & & & (s) & (h) & (M) & & & & & (s) & (h) & (M) 
      \\ \hline
      NeRF & 1
      & 31.04 & 0.951 & 0.055 & 256 & 9.60 & 4.70 & 1.19
      & 27.72 & 0.871 & 0.150 & 192 & 8.38 & 3.39 & 1.19
      \\ \hline
      MIMO-NeRF-naive & \multirow{3}{*}{2}
      & 30.18 & 0.944 & 0.065 & 128 & 5.15 & \first{3.09} & 1.26
      & 27.31 & 0.860 & \third{0.167} & 96 & 4.55 & \first{2.12} & 1.26
      \\
      MIMO-NeRF-distill &
      & \third{30.76} & \third{0.949} & \third{0.058} & 128 & 5.15 & 9.46 & 1.26
      & \third{27.50} & \third{0.863} & 0.169 & 96 & 4.55 & 6.81 & 1.26
      \\
      MIMO-NeRF-self &
      & \first{31.26} & \first{0.953} & \first{0.054} & 128 & 5.15 & \third{5.36} & 1.26
      & \first{27.70} & \first{0.870} & \first{0.155} & 96 & 4.55 & \third{3.97} & 1.26
      \\ \hline
      MIMO-NeRF-naive & \multirow{3}{*}{4}
      & 28.62 & 0.927 & 0.091 & 64 & 2.79 & \first{2.02} & 1.39
      & 26.29 & 0.824 & 0.218 & 48 & 2.46 & \first{1.57} & 1.39
      \\
      MIMO-NeRF-distill &
      & \third{30.22} & \third{0.946} & \third{0.065} & 64 & 2.79 & 8.42 & 1.39
      & \third{27.37} & \third{0.861} & \third{0.172} & 48 & 2.46 & 6.25 & 1.39
      \\
      MIMO-NeRF-self &
      & \first{30.94} & \first{0.950} & \first{0.058} & 64 & 2.79 & \third{4.68} & 1.39
      & \first{27.51} & \first{0.865} & \first{0.162} & 48 & 2.46 & \third{3.44} & 1.39
      \\ \hline
      MIMO-NeRF-naive & \multirow{3}{*}{8}
      & 26.34 & 0.895 & 0.133 & 32 & 1.66 & \first{1.66} & 1.65
      & 25.10 & 0.774 & 0.284 & 24 & 1.45 & \first{1.24} & 1.65
      \\
      MIMO-NeRF-distill &
      & \third{29.39} & \third{0.937} & \third{0.075} & 32 & 1.66 & 8.07 & 1.65
      & \first{27.01} & \first{0.851} & \third{0.184} & 24 & 1.45 & 5.91 & 1.65
      \\
      MIMO-NeRF-self &
      & \first{30.40} & \first{0.945} & \first{0.065} & 32 & 1.66 & \third{5.86} & 1.65
      & \third{26.97} & \first{0.851} & \first{0.180} & 24 & 1.45 & \third{4.43} & 1.65
      \\
    \end{tabularx}
  \end{center}
  \caption{Benchmark performance of MIMO-NeRFs.
    MIMO-NeRF-self outperformed MIMO-NeRF-naive and MIMO-NeRF-distill in terms of PSNR, SSIM, and LPIPS in most cases with shorter training time than MIMO-NeRF-distill.
    All MIMO-NeRFs outperformed the original NeRF in terms of inference time owing to the reduction of \#~Run.}
  \label{tab:benchmark}
\end{table*}

\smallskip\noindent\textbf{Results.}
From Table~\ref{tab:benchmark}, the following is observed.

\smallskip\noindent\textit{Image quality.}
MIMO-NeRF-self outperformed not only MIMO-NeRF-naive but also MIMO-NeRF-distill in most cases in terms of PSNR, SSIM, and LPIPS.
We conjecture that this occurred because the joint optimization of the teacher and student networks in MIMO-NeRF-self was more effective for training than the student-only optimization in MIMO-NeRF-distill.
Even MIMO-NeRF-self suffered from a trade-off between speed and quality with increasing $N_p$; however, MIMO-NeRF-self performed better than or comparably to the original NeRF when $N_p = 2$.
This may have occurred because the advantage of accumulating neighboring information and the disadvantage of handling the ambiguity were antagonistic in this case.

\begin{figure}[t]
  \begin{center}
    \includegraphics[width=\columnwidth]{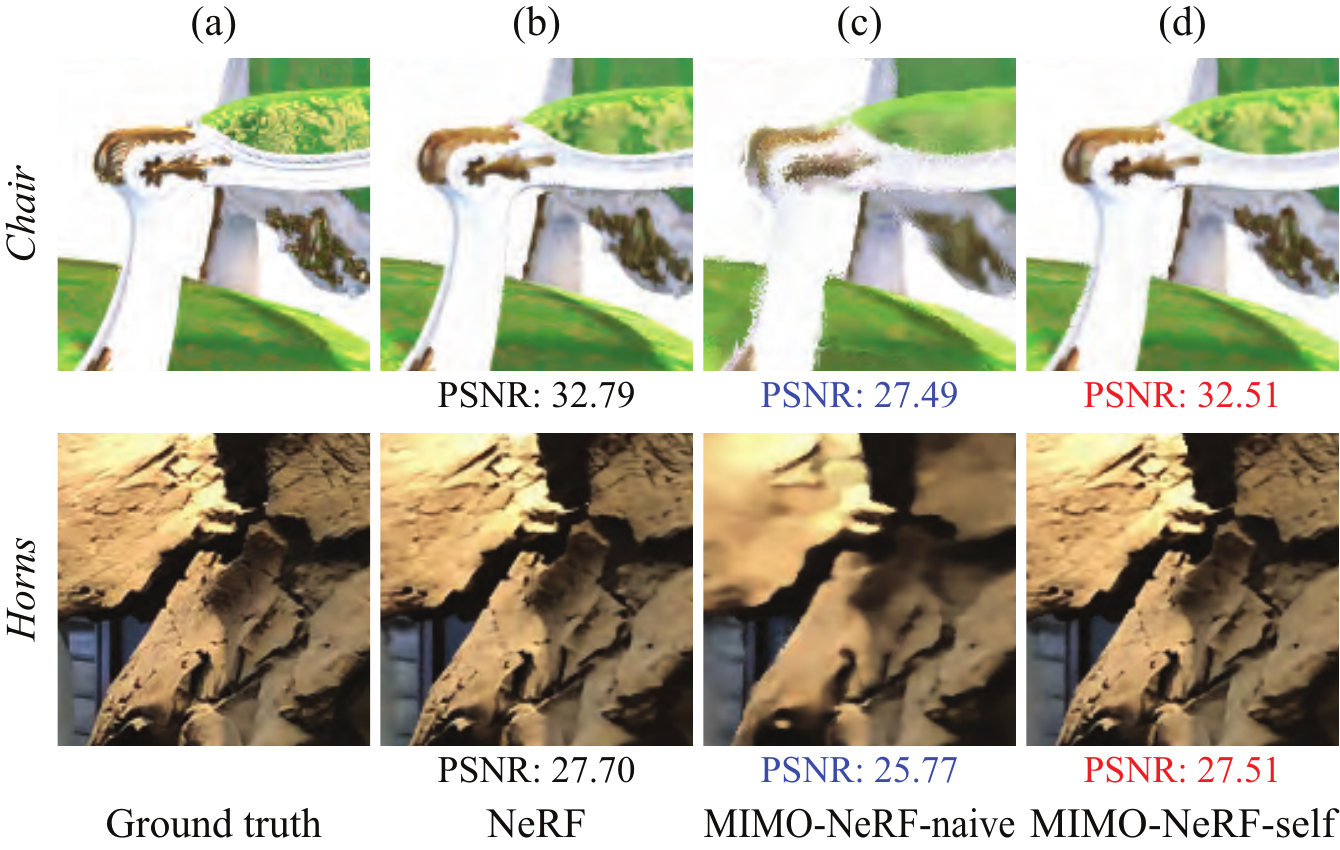}
  \end{center}
  \caption{Qualitative comparison between NeRF and MIMO-NeRFs with $N_p = 8$.
    The models were trained using full-sized images.
    MIMO-NeRF-naive (c) produced some artifacts owing to the ambiguity in the color and volume density of each point.
    MIMO-NeRF-self (d) was useful for addressing this issue, and its results were close to those of NeRF (b), while inference time were improved by a factor of $5.8$.}
  \label{fig:results_nerf}
\end{figure}

\smallskip\noindent\textit{Inference speed.}
All MIMO-NeRFs with the same inference procedure showed inference times improved by a factor of $1.84$--$5.78$ with increasing $N_p$.

\smallskip\noindent\textit{Training speed.}
MIMO-NeRF-naive achieved the fastest training because it used only a single MIMO formulation during training.
MIMO-NeRF-self required more training time because it uses multiple reformulated MIMO MLPs; however, each calculation cost is low. Therefore, it did not suffer from a large increase in training time compared with MIMO-NeRF-distill, which requires training not only a MIMO-NeRF but also a SISO NeRF.

\smallskip\noindent\textit{Summary.}
From these results, we found that when $N_p = 2$, MIMO-NeRF-self improved the inference speed of NeRF without compromising image quality, and when $N_p$ was larger, there was a trade-off between speed and quality.
We examine the validity of this trade-off in Section~\ref{subsec:investigation_tradeoff}.

\smallskip\noindent\textbf{Qualitative results.}
Figure~\ref{fig:results_nerf} shows a qualitative comparison between NeRF, MIMO-NeRF-naive, and MIMO-NeRF-self.
In this experiment, we used full-sized images to train the models and set $N_p$ to $8$ for MIMO-NeRFs.
MIMO-NeRF-naive produced some artifacts, whereas MIMO-NeRF-self adequately addressed this issue.
The additional results are provided in Appendix~\ref{subsec:full_size}.

\subsection{Investigation of speed-quality trade-off}
\label{subsec:investigation_tradeoff}

We compared MIMO-NeRF with possible alternatives to investigate whether it achieved a good trade-off between speed and quality.
In particular, we focused on methods that are general and applicable to various NeRFs, similar to MIMO-NeRF, and examined two variants, including
\textit{NeRF-few}, which reduced the number of samples on a ray, and
\textit{NeRF-small}, which reduced the number of features in the hidden layers.
We adjusted the parameters such that their FLOPs were comparable to those of MIMO-NeRF.\footnote{We tuned the models based on FLOPs because the performance of different methods for improving speed in terms of inference time may vary depending on the calculation tools used, such as GPU processor hardware.
  As a reference, we provide the relationship between the inference time and image quality in Appendix~\ref{subsec:analysis_tradeoff}.}

\begin{figure}[t]
  \begin{center}
    \includegraphics[width=\columnwidth]{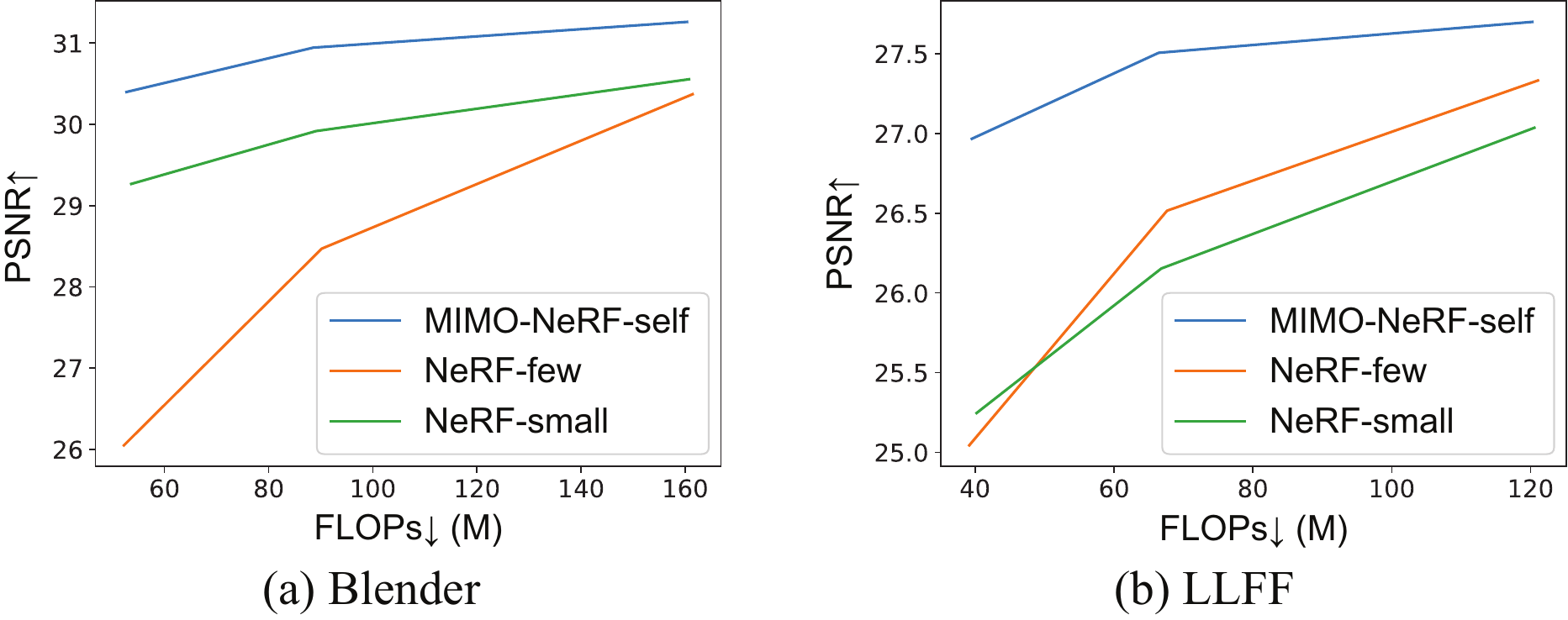}
  \end{center}
  \caption{Relationship between FLOPs and PSNR.
    Higher values indicate better the image quality.
    Faster speeds are shown to the left.
    MIMO-NeRF-self achieves the best trade-off between speed and quality.}
  \label{fig:tradeoff_flops_psnr}
\end{figure}

\smallskip\noindent\textbf{Results.}
The relationship between the FLOPs and PSNR is plotted in Figure~\ref{fig:tradeoff_flops_psnr}.
We found that MIMO-NeRF-self obtained a better trade-off between speed and quality than NeRF-few and NeRF-small.
We provide other relationships (e.g., relationships between FLOPs/inference time and PSNR/SSIM/LPIPS) in Appendix~\ref{subsec:analysis_tradeoff}.\footnote{\label{foot:calculation_cost}During training, the calculation cost of MIMO-NeRF-self was larger than those of NeRF-small and NeRF-few because multiple reformulated MIMO MLPs were used.
  To confirm this effect, we examined the performance of NeRF-few and NeRF-small with increasing batch sizes such that the calculation costs became almost the same as that of MIMO-NeRF-self.
  We found that MIMO-NeRF-self achieved a better trade-off between speed and quality than the other variants.
  Detailed results are provided in Appendix~\ref{subsec:analysis_tradeoff}.}

\subsection{Ablation studies}
\label{subsec:ablation_studies}

We also conducted ablation studies to better understand the performance of each element of the proposed self-supervised learning method.
Specifically, we investigated the importance of group shift, 3D consistency loss, and asymmetric weights.
When asymmetric weights were ablated, $\mu_{m_i}^{m_j}$ was set to $1$.

\smallskip\noindent\textbf{Results.}
The results are listed in Table~\ref{tab:ablation_study}.
We found that the full model achieved the best performance in most cases.
The results validate the importance of each technique.

\begin{table}
  \setlength{\tabcolsep}{3.5pt}
  \begin{center}
    \scriptsize
    \begin{tabularx}{\columnwidth}{c|ccc|ccc|ccc}
      & & & & \multicolumn{3}{c|}{Blender} & \multicolumn{3}{c}{LLFF}
      \\
      $N_p $ & GS & CL & AW & PSNR$\uparrow$ & SSIM$\uparrow$ & LPIPS$\downarrow$ & PSNR$\uparrow$ & SSIM$\uparrow$ & LPIPS$\downarrow$
      \\ \hline
      \multirow{3}{*}{2} & & \checkmark &
      & \third{30.17} & \third{0.943} & \thirdr{0.067}
      & \third{27.21} & \third{0.856} & \third{0.170}
      \\
      & \checkmark & &
      & \second{30.54} & \second{0.945} & \secondr{0.065}
      & \second{27.48} & \second{0.865} & \second{0.161}
      \\ 
      & \checkmark & \checkmark &
      & \first{31.26} & \first{0.953} & \firstr{0.054}
      & \first{27.70} & \first{0.870} & \first{0.155}
      \\ \hline
      \multirow{4}{*}{4} & & \checkmark & \checkmark
      & \third{30.84} & \second{0.949} & \firstr{0.058}
      & \third{27.39} & \third{0.862} & \third{0.166}
      \\
      & \checkmark & &
      & 29.48 & 0.936 & 0.077
      & 26.46 & 0.832 & 0.206
      \\
      & \checkmark & \checkmark &
      & \second{30.87} & \second{0.949} & \thirdr{0.060}
      & \second{27.44} & \second{0.864} & \second{0.164}
      \\ 
      & \checkmark & \checkmark & \checkmark
      & \first{30.94} & \first{0.950} & \firstr{0.058}
      & \first{27.51} & \first{0.865} & \first{0.162}
      \\ \hline
      \multirow{4}{*}{8} & & \checkmark & \checkmark
      & \third{29.81} & \third{0.941} & \secondr{0.069}
      & \first{26.97} & \first{0.851} & \first{0.179}
      \\
      & \checkmark & &
      & 27.39 & 0.907 & 0.116
      & 24.29 & 0.734 & 0.332
      \\
      & \checkmark & \checkmark &
      & \second{30.16} & \second{0.942} & \thirdr{0.071}
      & \third{26.69} & \third{0.843} & \third{0.192}
      \\ 
      & \checkmark & \checkmark & \checkmark
      & \first{30.40} & \first{0.945} & \firstr{0.065}
      & \first{26.97} & \first{0.851} & \second{0.180}
      \\
    \end{tabularx}
  \end{center}
  \caption{Results of ablation studies.
    Check marks in GS, CL, and AW indicate the use of group shift, 3D consistency loss, and asymmetric weights.
    In $N_p = 2$, asymmetric weights were not used in the full model because $R^1 = R^2 = 1$.
    Hence, it was not ablated.}
  \label{tab:ablation_study}
\end{table}

\subsection{Application to DONeRF}
\label{subsec:application_donerf}

We incorporated MIMO-NeRF into DONeRF~\cite{TNeffCVF2021}, a representative NeRF with sample reduction, to demonstrate that MIMO-NeRF can complement existing fast NeRFs.
DONeRF uses a sampling network called a depth oracle network to select samples and calculates the colors and volume densities of the selected samples using a shading network.
It handles the trade-off between speed and image quality by adjusting the number of selected samples ($N_s$).
We examined whether MIMO-NeRF could be used as an alternative to handle this trade-off.

\smallskip\noindent\textbf{Dataset.}
We evaluated the performance using the \textit{DONeRF dataset}~\cite{TNeffCVF2021} comprising six synthetic indoor and outdoor scenes.
Each scene included $300$ forward-facing views at a resolution of $800 \times 800$ pixels.
$70\%$, $10\%$, and $20\%$ of the images were used for training, validation, and testing, respectively.

\smallskip\noindent\textbf{Implementation.}
We implemented the models according to the source code of DONeRF\footnote{\label{foot:donerf}\url{https://github.com/facebookresearch/DONERF}} and trained them using the default settings.
We applied MIMO-NeRF with $N_p = 4$ to \textit{DONeRF-16} (i.e., DONeRF with $N_s = 16$).
In the self-supervised learning process, we used two formulations with $R^1 = R^2 = 1$ and group shifts.
This model is referred to as \textit{MIMO-DONeRF-16/4}.
We set $\lambda$ to $0.001$.
As a baseline, we examined DONeRF-4, which had the same \#~Run as MIMO-DONeRF-16/4.
The implementation details are presented in Appendix~\ref{subsec:implementation_details_donerf}.

\smallskip\noindent\textbf{Evaluation metrics.}
Following the study on DONeRF~\cite{TNeffCVF2021}, we assessed the image quality using the \textit{PSNR} and \textit{FLIP}~\cite{PAnderssonCGIT2020}.
In addition, we used the \textit{\#~Run}, \textit{I-time}, \textit{T-time}, and \textit{\#~Params} described in Section~\ref{subsec:investigation_performance}.
In DONeRF, \#~Run was calculated as $1 + \frac{N_s}{N_p}$.

\begin{table}
  \setlength{\tabcolsep}{1.75pt}
  \begin{center}
    \scriptsize
    \begin{tabularx}{\columnwidth}{l|cccccc}
      \multicolumn{1}{c|}{Model} & PSNR$\uparrow$ & FLIP$\downarrow$ & \# Run$\downarrow$ & I-time$\downarrow$ & T-time$\downarrow$ & \# Params
      \\
      & & & & (s) & (h) & (M)
      \\ \hline
      DONeRF-16
      & \first{33.06} & \first{0.061} & 17 & 0.429 & 3.79 & 0.94
      \\
      DONeRF-4
      & 31.21 & 0.070 & 5 & \first{0.140} & \first{3.23} & 0.94
      \\ \hline
      MIMO-DONeRF-16/4-naive
      & \third{32.30} & \third{0.063} & 5 & \second{0.155} & \second{3.26} & 0.99
      \\
      MIMO-DONeRF-16/4-self
      & \second{32.72} & \first{0.061} & 5 & \second{0.155} & \third{3.56} & 0.99
      \\
    \end{tabularx}
  \end{center}
  \caption{Comparison of quantitative scores between DONeRFs and MIMO-DONeRFs.
    MIMO-DONeRF-16/4-naive outperformed DONeRF-4 in terms of PSNR and FLIP with a small increase in I-time and T-time.
    The performance of MIMO-DONeRF-16/4-self was close to DONeRF-16 in terms of PSNR and FLIP with shorter I-time and T-time.}
  \label{tab:scores_donerf}
\end{table}

\smallskip\noindent\textbf{Results.}
The results are summarized in Table~\ref{tab:scores_donerf}.
It is observed that MIMO-DONeRF-16/4-naive outperformed DONeRF-4 in terms of PSNR and FLIP with a small increase in I-time and T-time.
MIMO-DONeRF-16/4-self enhanced the image quality with an increase in T-time, and its image quality approached that of DONeRF-16 in terms of PSNR and FLIP with faster inference and training.
These results suggest that the increase in $N_p$ (i.e., the replacement of the SISO MLP by the MIMO MLP) can be used as an alternative to the reduction in $N_s$ (the number of selected samples) to obtain a better trade-off between speed and quality.
A detailed analysis is presented in Appendix~\ref{subsec:detailed_analysis_donerf}.

\subsection{Application to TensoRF}
\label{subsec:application_tensorf}

A NeRF with alternative representations is another representative fast approach.
To demonstrate that MIMO-NeRF is also compatible with this model, we applied it to TensoRF~\cite{AChenECCV2022}, a representative model in this category.
TensoRF uses a vector-matrix decomposition to calculate the volume densities and color features and applies a SISO MLP to the color features to decode the RGB colors.
The corresponding ambiguity was relatively limited because the volume densities were extracted using an explicit representation.
Hence, we simply replaced the SISO MLP with a MIMO MLP without modifying the training process while prioritizing training speed.
These models are denoted as \textit{MIMO-TensoRF-$N_p$}, where $N_p$ was varied among $\{ 2, 4, 8 \}$.

\smallskip\noindent\textbf{Datasets.}
We examined the performance of our approach on the \textit{Blender}~\cite{BMildenhallECCV2020} and \textit{LLFF}~\cite{BMildenhallTOG2019} datasets described in Section~\ref{subsec:investigation_performance}.
Full-size images were used in this experiment.

\smallskip\noindent\textbf{Implementation.}
We implemented the models based on the official source code of TensoRF\footnote{\label{foot:tensorf}\url{https://github.com/apchenstu/TensoRF}} and trained all the models using the same default settings for a fair comparison.
The implementation details are presented in Appendix~\ref{subsec:implementation_details_tensorf}.

\smallskip\noindent\textbf{Evaluation metrics.}
Following the study on TensoRF~\cite{AChenECCV2022}, we measured the image quality using PSNR and SSIM~\cite{ZWangTIP2004}.
In addition, we used the \textit{\#~Run}, \textit{I-time}, \textit{T-time}, and \textit{\#~Params} described in Section~\ref{subsec:investigation_performance}.
In TensoRF, \#~Run is determined adaptively for each pixel.
Therefore, we report the average.

\begin{table}
  \setlength{\tabcolsep}{3.5pt}
  \begin{center}
    \scriptsize
    \begin{tabularx}{\columnwidth}{l|cccccc}
      \multicolumn{1}{c|}{Model} & PSNR$\uparrow$ & SSIM$\uparrow$ & \# Run$\downarrow$ & I-time$\downarrow$ & T-time$\downarrow$ & \# Params
      \\
      & & & & (s) & (m) & (M)
      \\ \hline
      TensoRF
      & \second{33.23} & \first{0.963} & 9.95 & 1.25 & 11.50 & 18.8
      \\ \hline
      MIMO-TensoRF-2
      & \first{33.26} & \first{0.963} & 4.76 & \third{1.18} & \third{10.89} & 18.8
      \\
      MIMO-TensoRF-4
      & \third{32.98} & \third{0.961} & 2.40 & \second{1.15} & \second{10.67} & 18.8
      \\
      MIMO-TensoRF-8
      & 32.37 & 0.956 & 1.27 & \first{1.14} & \first{10.57} & 18.9
      \\
      \multicolumn{7}{c}{(a) Blender}
      \vspace{1mm}
    \end{tabularx}
    \begin{tabularx}{\columnwidth}{l|cccccc}
      \multicolumn{1}{c|}{Model} & PSNR$\uparrow$ & SSIM$\uparrow$ & \# Run$\downarrow$ & I-time$\downarrow$ & T-time$\downarrow$ & \# Params
      \\
      & & & & (s) & (m) & (M)
      \\ \hline
      TensoRF
      & \first{26.73} & \first{0.837} & 126.73 & 6.64 & 23.41 & 46.8
      \\ \hline
      MIMO-TensoRF-2
      & \second{26.72} & \first{0.837} & 62.14 & \third{6.18} & \third{21.63} & 46.8
      \\
      MIMO-TensoRF-4
      & \second{26.72} & \third{0.836} & 30.16 & \second{5.76} & \second{21.15} & 46.8
      \\
      MIMO-TensoRF-8
      & 26.64 & 0.835 & 14.52 & \first{5.52} & \first{20.68} & 46.9
      \\
      \multicolumn{7}{c}{(b) LLFF}
    \end{tabularx}
  \end{center}
  \caption{Comparison of quantitative scores between TensoRF and MIMO-TensoRF.
    MIMO-TensoRF improved I-time and T-time while retaining PSNR and SSIM when $N_p \leq 2$ and $N_p \leq 4$ on the Blender and LLFF datasets, respectively.}
  \label{tab:scores_tensorf}
\end{table}

\smallskip\noindent\textbf{Results.}
The results are presented in Table~\ref{tab:scores_tensorf}.
It is observed that MIMO-TensoRF improved I-time and T-time with similar image quality when $N_p$ was set within an adequate range (in particular, $N_p \leq 2$ on the Blender dataset and $N_p \leq 4$ on the LLFF dataset).
These results suggest that MIMO-TensoRF can strengthen the inference and training speed of TensoRF without negative effects by adequately selecting $N_p$.
A detailed analysis is presented in Appendix~\ref{subsec:detailed_analysis_tensorf}.

\section{Discussion}
\label{sec:discussion}

The results of these experiments in various situations demonstrate that MIMO-NeRF achieved a good trade-off between speed and quality.
However, we also found that the quality degradation became significant with increasing $N_p$.
One possible reason for this is that we did not modify the baseline network except for its input and output and did not increase the capacity of the models.
It might be natural to implement a model of larger capacity to handle larger combinations of inputs and outputs.
We did not adopt this strategy to ensure a fair comparison.
However, searching for the best configurations considering the number of samples, the number of groups (the proposed new searching area), and the size of the model remain as a practically imperative and promising direction for further research.

\section{Conclusion}
\label{sec:conclusion}

In this study, we have proposed MIMO-NeRF to improve the rendering speed of NeRF.
Our core idea is that of replacing the SISO MLP used in standard NeRFs with a MIMO-MLP.
We have developed a novel self-supervised learning method to address the ambiguity in the color and volume density of each point without relying on pretrained models.
The results of an experimental evaluation have shown that MIMO-NeRF achieves a good trade-off between speed and quality with a reasonable training time.
Although we have demonstrated the versatility of MIMO-NeRF by applying it to various NeRFs, many implicit neural representations aside from NeRFs also partially or primarily use SISO MLPs.
We expect our ideas to be utilized with a few modifications to speed up the execution of such models.

{\small
\bibliographystyle{ieee_fullname}
\bibliography{egbib}
}

\clearpage
\appendix

\section{Further analyses}
\label{sec:further_analyses}

In this appendix, the following analyses are presented:
\begin{itemize}
\item Appendix~\ref{subsec:effect_grouping}:
  Effect of grouping methods
\item Appendix~\ref{subsec:effect_reformulation}:
  Effect of reformulation methods
\item Appendix~\ref{subsec:effect_hyperparameter}:
  Effect of hyperparameter
\item Appendix~\ref{subsec:analysis_tradeoff}:
  Detailed analysis of speed-quality trade-off
\item Appendix~\ref{subsec:effectiveness_increasing_n}:
  Effectiveness when increasing $N$
\item Appendix~\ref{subsec:full_size}:
  Effectiveness for full-sized images
\item Appendix~\ref{subsec:comparison_with_autoint}:
  Comparison with AutoInt
\item Appendix~\ref{subsec:detailed_analysis_donerf}:
  Detailed analysis of application to DONeRF
\item Appendix~\ref{subsec:detailed_analysis_tensorf}:
  Detailed analysis of application to TensoRF
\end{itemize}

\subsection{Effect of grouping methods}
\label{subsec:effect_grouping}

As discussed in Section~\ref{subsec:formulation}, several methods exist for grouping the input samples when constructing a MIMO MLP.
For example, when focusing on a method for grouping samples on a ray,\footnote{We focused on grouping methods that can be conducted per ray for two reasons:
  (1) In typical NeRF training, rendering is performed for randomly sampled rays.
  Therefore, a batch does not necessarily include near rays.
  (2) In NeRF, searching for near points across different rays is not trivial because points are sampled unevenly via hierarchical sampling.} two opposite methods could be considered:
(1) Construction of a \textit{general} MIMO MLP that can accept any combination of samples in a ray.
(2) Construction of a \textit{specific} MIMO MLP that accepts only a group of nearby samples.
This study adopts the latter method, assuming that learning a general model is more difficult than learning a specific one.
This appendix examined their difference in performance to verify this statement.
More precisely, we compared \textit{MIMO-NeRF-naive}, which grouped neighboring samples on a ray, with \textit{MIMO-NeRF-random}, which randomly grouped samples on a ray.
To focus on the comparison of the grouping methods, we did not use an advanced training scheme such as self-supervised learning.

\smallskip\noindent\textbf{Results.}
Table~\ref{tab:grouping_methods} summarizes the results.
We only present the image quality scores, that is, PSNR, SSIM, and LPIPS, because the difference in the grouping methods did not affect the other scores, that is, \#~Run, I-time, T-time, and \#~Params.
As can be observed, MIMO-NeRF-naive outperforms MIMO-NeRF-random in all cases.
These results indicated that the construction of a specific MIMO MLP was better in our experimental settings.
We note that there is a possibility that a general MIMO-MLP can achieve comparable performance when using a larger-capacity model.
However, in this case, the rendering speed slows down.
Therefore, such a model is beyond the scope of this study.

\begin{table}[t]
  \setlength{\tabcolsep}{2pt}
  \begin{center}
    \scriptsize
    \begin{tabularx}{\columnwidth}{lc|CCC|CCC}
      & & \multicolumn{3}{c|}{Blender} & \multicolumn{3}{c}{LLFF}
      \\
      \multicolumn{1}{c}{Model} & $N_p $ & PSNR$\uparrow$ & SSIM$\uparrow$ & LPIPS$\downarrow$ & PSNR$\uparrow$ & SSIM$\uparrow$ & LPIPS$\downarrow$
      \\ \hline
      MIMO-NeRF-naive & \multirow{2}{*}{2}
      & \first{30.18} & \first{0.944} & \firstr{0.065}
      & \first{27.31} & \first{0.860} & \first{0.167}
      \\
      MIMO-NeRF-random &
      & 28.48 & 0.920 & 0.102
      & 24.90 & 0.766 & 0.294
      \\ \hline
      MIMO-NeRF-naive & \multirow{2}{*}{4}
      & \first{28.62} & \first{0.927} & \firstr{0.091}
      & \first{26.29} & \first{0.824} & \first{0.218}
      \\
      MIMO-NeRF-random &
      & 25.40 & 0.871 & 0.167
      & 22.73 & 0.634 & 0.424
      \\ \hline
      MIMO-NeRF-naive & \multirow{2}{*}{8}
      & \first{26.34} & \first{0.895} & \firstr{0.133}
      & \first{25.10} & \first{0.774} & \first{0.284}
      \\
      MIMO-NeRF-random &
      & 23.17 & 0.836 & 0.207
      & 21.46 & 0.563 & 0.476
      \\
    \end{tabularx}
  \end{center}
  \caption{Effect of grouping methods.
    MIMO-NeRF-naive, which groups neighboring samples on a ray, outperforms MIMO-NeRF-random, which groups samples on a ray randomly, in all the cases.}
  \label{tab:grouping_methods}
\end{table}

\subsection{Effect of reformulation methods}
\label{subsec:effect_reformulation}

In Section~\ref{subsec:investigation_performance}, for $N_p = 2^L$ ($L > 1$), we used $L$ reformulated MIMO MLPs with
\begin{flalign}
  \label{eq:r1}
  R^1 = 1, \dots, R^{L} = 2^{L-1}.
\end{flalign}
In this case, the total number of MLPs running is calculated as
\begin{flalign}
  \label{eq:r1_total}
  \sum_{m=1}^L \frac{N}{N_p} {R^m}
  = N \left( \frac{1}{2^L} + \cdots + \frac{1}{2} \right)
  = N \left( 1 - \frac{1}{2^L} \right)
  < N.
\end{flalign}
Therefore, we can prevent a large increase in the training time compared with the original (i.e., SISO) MLP, in which the number of MLPs running is $N$.\footnote{More strictly, when a group shift is conducted, padding is performed.
  In this case, the number of group shifts is added to the number of MLPs running in Equation~\ref{eq:r1_total}.
  Note that the number of group shifts is equal to or smaller than the number of reformulated MIMO MLPs.
  Therefore, the effect was small.}
We denote MIMO-NeRF with this reformulation method as \textit{MIMO-NeRF-self-R1}.
For further analysis, this appendix investigates other reformulation methods.
In particular, when $N_p = 2$, the use of two reformulated MIMO MLPs with $R^1 = 1$ and $R^2 = 1$ is the only effective option in which the total number of MLPs running does not exceed $N$.
Therefore, we investigated different reformulation methods for $N_p > 2$.
Specifically, five reformulation methods were examined.

\begin{table*}[t]
  \setlength{\tabcolsep}{3pt}
  \begin{center}
    \scriptsize
    \begin{tabularx}{0.85\textwidth}{lcl|CCCC|CCCC}
      & & & \multicolumn{4}{c|}{Blender} & \multicolumn{4}{c}{LLFF}
      \\
      \multicolumn{1}{c}{Model} & $N_p $ & \multicolumn{1}{c|}{$R$} & PSNR$\uparrow$ & SSIM$\uparrow$ & LPIPS$\downarrow$ & T-time$\downarrow$ & PSNR$\uparrow$ & SSIM$\uparrow$ & LPIPS$\downarrow$ & T-time$\downarrow$
      \\
      & & & & & & (h) & & & & (h)
      \\ \hline
      NeRF & 1 & \multicolumn{1}{c|}{--}
      & 31.04 & 0.951 & 0.055 & 4.70
      & 27.72 & 0.871 & 0.150 & 3.39
      \\ \hline
      MIMO-NeRF-naive & \multirow{3}{*}{2} & \multicolumn{1}{c|}{--}
      & \third{30.18} & \third{0.944} & \third{0.065} & \first{3.09}
      & \third{27.31} & \third{0.860} & \second{0.167} & \first{2.12}
      \\
      MIMO-NeRF-distill & & \multicolumn{1}{c|}{--}
      & \second{30.76} & \second{0.949} & \second{0.058} & \thirdr{9.46}
      & \second{27.50} & \second{0.863} & \third{0.169} & \third{6.81}
      \\
      MIMO-NeRF-self & & $R^1 = 1, R^2 = 1$
      & \first{31.26} & \first{0.953} & \first{0.054} & \secondr{5.36}
      & \first{27.70} & \first{0.870} & \first{0.155} & \second{3.97}
      \\ \hline
      MIMO-NeRF-naive & \multirow{5}{*}{4} & \multicolumn{1}{c|}{--}
      & 28.62 & 0.927 & 0.091 & \firstr{2.02}
      & 26.29 & 0.824 & 0.218 & \first{1.57}
      \\
      MIMO-NeRF-distill & & \multicolumn{1}{c|}{--}
      & 30.22 & 0.946 & 0.065 & 8.42
      & \second{27.37} & \second{0.861} & 0.172 & 6.25
      \\
      MIMO-NeRF-self-R1 & & $R^1 = 1, R^2 = 2$
      & \second{30.94} & \first{0.950} & \first{0.058} & \thirdr{4.68}
      & \first{27.51} & \first{0.865} & \first{0.162} & \third{3.44}
      \\
      MIMO-NeRF-self-R2 & & $R^1 = 1, R^2 = 1$
      & \first{30.95} & \first{0.950} & \second{0.060} & \secondr{3.65}
      & \third{27.35} & \second{0.861} & \second{0.169} & \second{2.70}
      \\
      MIMO-NeRF-self-R3 & & $R^1 = 1, R^2 = 1, R^3 = 1$
      & \third{30.89} & \third{0.949} & \second{0.060} & 5.05
      & 27.27 & 0.860 & \third{0.171} & 3.78
      \\ \hline
      MIMO-NeRF-naive & \multirow{8}{*}{8} & \multicolumn{1}{c|}{--}
      & 26.34 & 0.895 & 0.133 & \firstr{1.66}
      & 25.10 & 0.774 & 0.284 & \first{1.24}
      \\
      MIMO-NeRF-distill & & \multicolumn{1}{c|}{--}
      & 29.39 & 0.937 & \third{0.075} & 8.07
      & \first{27.01} & \first{0.851} & \second{0.184} & 5.91
      \\
      MIMO-NeRF-self-R1 & & $R^1 = 1, R^2 = 2, R^3 = 4$
      & \first{30.40} & \first{0.945} & \first{0.065} & 5.86
      & \second{26.97} & \first{0.851} & \first{0.180} & 4.43
      \\
      MIMO-NeRF-self-R2 & & $R^1 = 1, R^2 = 1$
      & \third{30.02} & \third{0.940} & 0.076 & \secondr{2.61}
      & 26.52 & 0.833 & 0.207 & \second{2.13}
      \\
      MIMO-NeRF-self-R3 & & $R^1 = 1, \dots, R^7 = 1$
      & 29.88 & 0.937 & 0.080 & 7.75
      & 25.66 & 0.797 & 0.243 & 5.97
      \\
      MIMO-NeRF-self-R4 & & $R^1 = 1, R^2 = 2$
      & 29.86 & 0.939 & 0.077 & \thirdr{3.33}
      & \third{26.61} & 0.836 & 0.205 & \third{2.41}
      \\
      MIMO-NeRF-self-R5 & & $R^1 = 1, R^2 = 4$
      & 29.81 & 0.939 & 0.076 & 4.38
      & \third{26.61} & \third{0.838} & \third{0.202} & 3.37
      \\
      MIMO-NeRF-self-R6 & & $R^1 = 1, R^2 = 1, R^3 = 1$
      & \second{30.11} & \second{0.941} & \second{0.074} & 3.76
      & 26.41 & 0.830 & 0.208 & 2.73
      \\
    \end{tabularx}
  \end{center}
  \caption{Effect of reformulation methods.
    We examined the PSNR, SSIM, LPIPS, and T-time scores when changing the reformulation methods.
    MIMO-NeRF-self-R1, which is used in the main experiments, achieves the best or comparable performance in terms of PSNR, SSIM, and LPIPS.
    Other variants are outperformed by it in most cases; however, some of them, e.g., MIMO-NeRF-self-R2 with $N_p = 4$ on the Blender and LLFF datasets and MIMO-NeRF-self-R2/R6 with $N_p = 8$ on the Blender dataset, achieve a performance comparable with that of MIMO-NeRF-distill while achieving faster training than the original NeRF.}
  \label{tab:reformulation_methods}
\end{table*}

\smallskip\noindent\textit{MIMO-NeRF-self-R2:}
This variant uses two reformulated MIMO MLPs with
\begin{flalign}
  \label{eq:r2}
  R^1 = 1, R^2 = 1.
\end{flalign}
In this case, the total number of MLPs running is calculated as
\begin{flalign}
  \label{eq:r2_total}
  \sum_{m=1}^2 \frac{N}{N_p} R^m = N \frac{2}{N_p} < N \text{ when } N_p > 2.
\end{flalign}

\smallskip\noindent\textit{MIMO-NeRF-self-R3:}
This variant employs $N_p - 1$ reformulated MIMO MLPs with
\begin{flalign}
  \label{eq:r3}
  R^1 = 1, \dots, R^{N_p - 1} = 1.
\end{flalign}
In this case, the total number of MLPs running is calculated as
\begin{flalign}
  \label{eq:r3_total}
  \sum_{m=1}^{N_p - 1} \frac{N}{N_p} R^m = N \frac{N_p - 1}{N_p} < N.
\end{flalign}

\smallskip\noindent\textit{MIMO-NeRF-self-R4:}
This variant adopts two reformulated MIMO MLPs with
\begin{flalign}
  \label{eq:r4}
  R^1 = 1, R^2 = 2.
\end{flalign}
In this case, the total number of MLPs running is calculated as
\begin{flalign}
  \label{eq:r4_total}
  \sum_{m=1}^2 \frac{N}{N_p} R^m = N \frac{3}{N_p} < N \text{ when } N_p > 3.
\end{flalign}
This method is the same as MIMO-NeRF-self-R1 when $N_p = 4$.

\smallskip\noindent\textit{MIMO-NeRF-self-R5:}
This variant uses two reformulated MIMO MLPs with
\begin{flalign}
  \label{eq:r5}
  R^1 = 1, R^2 = \frac{N_p}{2}.
\end{flalign}
In this case, the total number of MLPs running is calculated as
\begin{flalign}
  \label{eq:r5_total}
  \sum_{m=1}^2 \frac{N}{N_p} R^m = N \left( \frac{1}{2} + \frac{1}{N_p} \right) < N \text{ when } N_p > 2.
\end{flalign}
This method is the same as MIMO-NeRF-self-R1 when $N_p = 4$.

\smallskip\noindent\textit{MIMO-NeRF-self-R6:}
This variant uses three reformulated MIMO MLPs with
\begin{flalign}
  \label{eq:r6}
  R^1 = 1, R^2 = 1, R^3 = 1.
\end{flalign}
In this case, the total number of MLPs running is calculated as
\begin{flalign}
  \label{eq:r6_total}
  \sum_{m=1}^3 \frac{N}{N_p} R^m = N \frac{3}{N_p} < N \text{ when } N_p > 3.
\end{flalign}
This method is identical to MIMO-NeRF-self-R3 when $N_p = 4$.

\smallskip\noindent\textbf{Results}.
Table~\ref{tab:reformulation_methods} summarizes the results.
Our findings were as follows:

\smallskip\noindent\textit{MIMO-NeRF-self-R2 vs. MIMO-NeRF-self-R3 vs. MIMO-NeRF-self-R6.}
For these variants, the same variation reduction methods (i.e., $R^m = 1$) are used, whereas the numbers of reformulated MIMO MLPs (i.e., $M$) are different.
We found that too many reformulated MIMO MLPs (i.e., MIMO-NeRF-self-R3) did not necessarily achieve the best performance.
A possible reason for this is that an excessive number of constraints causes statistical averaging and deteriorates the image quality.
As $M$ increased, the training time increased.
Therefore, the results suggest that the use of the MIMO-NeRF with a moderate value of $M$ is preferable.

\smallskip\noindent\textit{MIMO-NeRF-self-R2 vs. MIMO-NeRF-self-R4 vs. MIMO-NeRF-self-R5.}
In these variants, the number of reformulated MIMO-MLPs is the same (i.e., $M = 2$), whereas different reduction methods are used.
We observed different tendencies in the results of the Blender dataset and those for the LLFF dataset.
In the Blender dataset, PSNR, SSIM, and LPIPS improved as $R^2$ decreased, whereas, in the LLFF dataset, they improved as $R^2$ increased.
Although not the same, similar tendencies exist between MIMO-NeRF-self-R1 and MIMO-NeRF-self-R2 when $N_p = 4$.
The results indicate that the variation reduction is more effective for the LLFF dataset, which includes forward-facing views, than for the Blender dataset, which contains $360^\circ$ views when $M = 2$.
However, it is noteworthy that MIMO-NeRF-self-R1 outperformed MIMO-NeRF-self-R6 on both datasets.
These results indicate that variation reduction is effective for both datasets when $M$ is sufficiently large.
Delving deeper into these differences will be an interesting topic for future research.

\smallskip\noindent\textit{MIMO-NeRF-self-R1 vs. the others.}
MIMO-NeRF-self-R1 achieved the best or comparable performance in terms of the image quality metrics, that is, PSNR, SSIM, and LPIPS, in all cases.
We note that some other variants, such as MIMO-NeRF-self-R2 with $N_p = 4$ on the Blender and LLFF datasets and MIMO-NeRF-self-R2/R6 with $N_p = 8$ on the Blender dataset, are worse than MIMO-NeRF-self-R1 in terms of all or some of the image quality metrics, but are comparable with MIMO-NeRF-distill while having a shorter training time than NeRF.
The results suggest the possibility of obtaining a reasonable quality and fast-inference model with a shorter training time by tuning the reformulation configurations.

\subsection{Effect of hyperparameter}
\label{subsec:effect_hyperparameter}

In the experiments presented in Sections~\ref{subsec:investigation_performance}--\ref{subsec:ablation_studies}, we set hyperparameter $\lambda$ to $1$ and $0.4$ for the Blender and LLFF datasets, respectively.
To analyze the effect of this hyperparameter, we examined the quantitative scores when varying $\lambda$ within $\{ 0.4, 1 \}$.

\smallskip\noindent\textbf{Results.}
Table~\ref{tab:hyperparameter} presents the results.
We present only the image quality scores because the modification of $\lambda$ does not affect the other scores, i.e., \#~Run, I-time, T-time, and \#~Params.
As can be observed, MIMO-NeRF is sensitive to $\lambda$, and in all cases, it achieved the best performance when using the values utilized in the experiments presented in Sections~\ref{subsec:investigation_performance}--\ref{subsec:ablation_studies} (i.e., $\lambda = 1$ for the Blender dataset and $\lambda = 0.4$ for the LLFF dataset).
However, the difference is relatively small, and the scores in the worst case are still comparable to those of MIMO-NeRF-distill (Table~\ref{tab:reformulation_methods}).
Therefore, we consider that this sensitivity is within an allowable range if $\lambda \in [ 0.4, 1 ]$.

\begin{table}[t]
  \setlength{\tabcolsep}{2pt}
  \begin{center}
    \scriptsize
    \begin{tabularx}{\columnwidth}{cc|CCC|CCC}
      & & \multicolumn{3}{c|}{Blender} & \multicolumn{3}{c}{LLFF}
      \\
      $N_p$ & $\lambda$ & PSNR$\uparrow$ & SSIM$\uparrow$ & LPIPS$\downarrow$ & PSNR$\uparrow$ & SSIM$\uparrow$ & LPIPS$\downarrow$
      \\ \hline
      \multirow{2}{*}{2} & 0.4
      & 31.20 & 0.952 & \firstr{0.054}
      & \first{27.70} & \first{0.870} & \first{0.155}
      \\
      & 1.0
      & \first{31.26} & \first{0.953} & \firstr{0.054}
      & 27.56 & 0.866 & 0.160
      \\ \hline
      \multirow{2}{*}{4} & 0.4
      & 30.93 & \first{0.950} & \firstr{0.058}
      & \first{27.51} & \first{0.865} & \first{0.162}
      \\
      & 1.0
      & \first{30.94} & \first{0.950} & \firstr{0.058}
      & 27.40 & 0.863 & 0.165
      \\ \hline
      \multirow{2}{*}{8} & 0.4
      & 30.22 & 0.944 & 0.067
      & \first{26.97} & \first{0.851} & \first{0.180}
      \\
      & 1.0
      & \first{30.40} & \first{0.945} & \firstr{0.065}
      & 26.83 & 0.848 & 0.184
      \\
    \end{tabularx}
  \end{center}
  \caption{Effect of hyperparameter $\lambda$.
    MIMO-NeRF is sensitive to $\lambda$; however, the difference is relatively small, and the scores in the worst case are still comparable to those of MIMO-NeRF-distill (Table~\ref{tab:reformulation_methods}).}
  \label{tab:hyperparameter}
\end{table}

\begin{figure*}[t]
  \begin{center}
    \includegraphics[width=\textwidth]{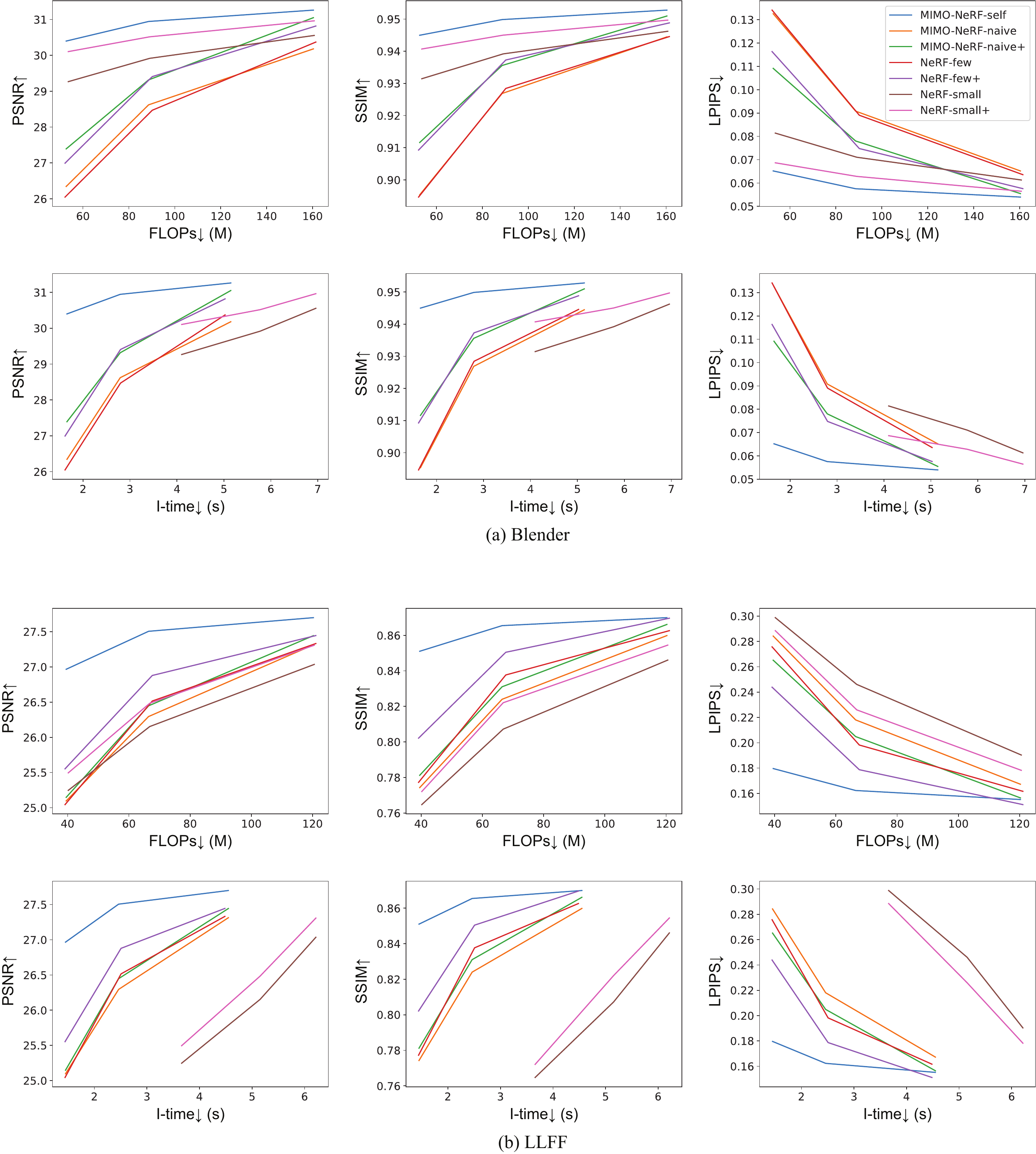}
  \end{center}
  \caption{Relationships between FLOPs/inference time and PSNR/SSIM/LPIPS.
    The legend is provided in the upper right figure.
    In the ``FLOPs'' axis, the more to the left, the lower the calculation cost.
    In the ``I-time'' axis, the more to the left, the faster the inference.
    In the ``PSNR'' and ``SSIM'' axis, the more to the upper side, the better the image quality.
    In the ``LPIPS'' axis, the more to the lower side, the better the image quality.
    MIMO-NeRF-self (blue line) achieved the best trade-off between speed and quality in almost all cases.}
  \label{fig:tradeoff}
\end{figure*}

\subsection{Detailed analysis of speed-quality trade-off}
\label{subsec:analysis_tradeoff}

In Section~\ref{subsec:investigation_tradeoff}, we present the relationship between FLOPs and PSNR as a method to demonstrate the trade-off between speed and quality.
For a detailed analysis, this appendix provides other relationships, including those between FLOPs/inference time and PSNR/SSIM/LPIPS.

\smallskip\noindent\textbf{Comparison models.}
In Section~\ref{subsec:investigation_tradeoff}, we compared \textit{MIMO-NeRF-self} with two possible alternatives:
\textit{NeRF-few}, which reduced the number of samples on a ray, and
\textit{NeRF-small}, which reduced the number of features in the hidden layers.
In particular, we adjusted the parameters so that their FLOPs in inference were comparable to those of MIMO-NeRF-self.
We describe the details of these models in Appendix~\ref{subsubsec:model_nerf}.
As discussed in the footnote,\footnoteref{foot:calculation_cost} an unignorable difference between MIMO-NeRF-self, MIMO-NeRF-few, and MIMO-NeRF-small is the difference in the calculation cost during training.
Because MIMO-NeRF-self uses multiple reformulated MIMO MLPs during training, the calculation cost is higher than that of NeRF-small and NeRF-few.
To confirm this effect, we examined the performance of NeRF-few and NeRF-small when increasing the batch size such that the calculation cost became almost the same as that of MIMO-NeRF-self.
These variants are referred to as \textit{NeRF-small+} and \textit{NeRF-few+}.
Furthermore, to confirm whether the proposed self-supervised learning was more effective than a simple increase in the batch size, we examined \textit{MIMO-NeRF-naive+}, where we increased the batch size, similar to NeRF-small+ and NeRF-few+.
More precisely, when $N_p = 2$, we used two reformulated MIMO MLPs with $R^1 = 1$ and $R^2 = 1$ for MIMO-NeRF-self.
In this case, \#~Run was twice that of MIMO-NeRF-naive.\footnote{More strictly, \#~Run increases more when a group shift is conducted because padding is performed.
  In this case, the number of group shifts, which is equal to or smaller than the number of reformulated MIMO MLPs, is added to \#~Run.
  However, this was relatively small compared to the number of samples.
  Therefore, we ignore its effect here.}
Therefore, we increased the batch size twice for NeRF-few+ and NeRF-small+.
Similarly, when compared to MIMO-NeRF-self with $N_p = 4$, we increased the batch size three times, and when compared to MIMO-NeRF-self with $N_p = 8$, we increased the batch size seven times.

\smallskip\noindent\textbf{Results.}
Figure~\ref{fig:tradeoff} presents the relationship between FLOPs/inference time and PSNR/SSIM/LPIPS.
We can observe that in most cases, MIMO-NeRF-self achieves a better trade-off between speed and quality in terms of every relationship than not only MIMO-NeRF-naive, NeRF-few, and NeRF-small, which are presented in Sections~\ref{subsec:investigation_performance} and \ref{subsec:investigation_tradeoff}, but also MIMO-NeRF-naive+, NeRF-few+, and NeRF-small+, which are trained under better conditions.
These results strengthen our statement in the main text, that is, MIMO-NeRF-self achieves a better trade-off between speed and quality than the possible alternatives.

\subsection{Effectiveness when increasing $N$}
\label{subsec:effectiveness_increasing_n}

In the main experiments, we investigated the performance of MIMO-NeRF when the number of samples (i.e., $N$) is fixed.
An interesting question is how MIMO-NeRF works well when increasing $N$ within the range in which its FLOPs are comparable to those of the original NeRF.
We conducted an additional experiment to answer this question.

\smallskip\noindent\textbf{Results.}
Table~\ref{tab:increasing_n} presents the results.
The models were evaluated using the Blender dataset.
It can be seen that MIMO-NeRF-self outperforms NeRF in terms of all metrics, and all scores improve as $N$ and $N_p$ increase.
The results indicate that tuning not only $N$ but also $N_p$ is important for obtaining the best performance under the same computational budget.

\begin{table}[t]
  \setlength{\tabcolsep}{5pt}
  \begin{center}
    \scriptsize
    \begin{tabularx}{\columnwidth}{lcc|CCCC}
      \multicolumn{1}{c}{Model} & N & $N_p $ & PSNR$\uparrow$ & SSIM$\uparrow$ & LPIPS$\downarrow$ & FLOPs
      \\
      & & & & & & (M)
      \\ \hline
      NeRF & 256 & 1 & 31.04 & 0.951 & 0.055 & 303.82
      \\
      MIMO-NeRF-self ($N_p = 2$) & 360 & 2 & \third{31.59} & \third{0.955} & \third{0.050} & 300.63
      \\
      MIMO-NeRF-self ($N_p = 4$) & 648 & 4 & \first{31.65} & \first{0.956} & \first{0.049} & 298.99
      \\
    \end{tabularx}
  \end{center}
  \caption{Effectiveness when increasing $N$.
    We compared NeRF and MIMO-NeRF-self when the FLOPs are almost the same.
    We evaluated the models on the Blender dataset.
    All the scores become better as $N$ and $N_p$ increase.}
  \label{tab:increasing_n}
\end{table}

\subsection{Effectiveness for full-sized images}
\label{subsec:full_size}

In Sections~\ref{subsec:investigation_performance}--\ref{subsec:ablation_studies}, half-sized images are used to better investigate the various configurations.
This appendix examines the effectiveness of MIMO-NeRF for full-sized images to verify whether the same conclusion holds independently of the image size.
In particular, we investigate the benchmark performance for full-sized images using a protocol similar to that described in Section~\ref{subsec:investigation_performance}.

\smallskip\noindent\textbf{Quantitative results.}
Table~\ref{tab:scores_nerf_full} summarizes the results for all the metrics (i.e., PSNR, SSIM, LPIPS, \# Run, I-time, T-time, and \#~Params).
Table~\ref{tab:scores_nerf_full_each} lists PSNR, SSIM, and LPIPS for each scene.
Similar to the analysis conducted in Section~\ref{subsec:investigation_performance}, we analyze the results from three perspectives:

\smallskip\noindent\textit{Image quality.}
Similar to the results for half-sized images, MIMO-NeRF-self outperformed MIMO-NeRF-self-naive but also MIMO-NeRF-self-distill in most cases in terms of PSNR, SSIM, and LPIPS.
Even MIMO-NeRF-self suffers from a trade-off between speed and quality as $N_p$ increases; however, MIMO-NeRF-self is comparable to the original NeRF when $N_p = 2$.

\smallskip\noindent\textit{Inference speed.}
Similar to the results for half-sized images, all MIMO-NeRFs improved the inference time by $1.83$--$5.77$ times as $N_p$ increased.

\smallskip\noindent\textit{Training speed.}
Similar to the results for half-sized images, MIMO-NeRF-naive achieved the fastest training because it used only a single MIMO formulation during training.
MIMO-NeRF-self increases the training time owing to the introduction of multiple reformulated MIMO MLPs; however, each calculation cost is lower than that of a SISO MLP in the original NeRF. Therefore, it does not suffer from a large increase in training time compared with MIMO-NeRF-distill, which requires the training of two networks, that is, a SISO-NeRF and a MIMO-NeRF.

\smallskip\noindent\textit{Summary.}
From these results, we found that when $N_p = 2$, MIMO-NeRF-self improves the inference speed of NeRF while retaining the image quality, and when $N_p$ is larger, MIMO-NeRF-self suffers from a trade-off between speed and quality; however, it achieves better image quality with a shorter training time than MIMO-NeRF-distill.
These tendencies are the same as those for the half-sized images.

\begin{figure*}[t]
  \centering
  \includegraphics[width=\textwidth]{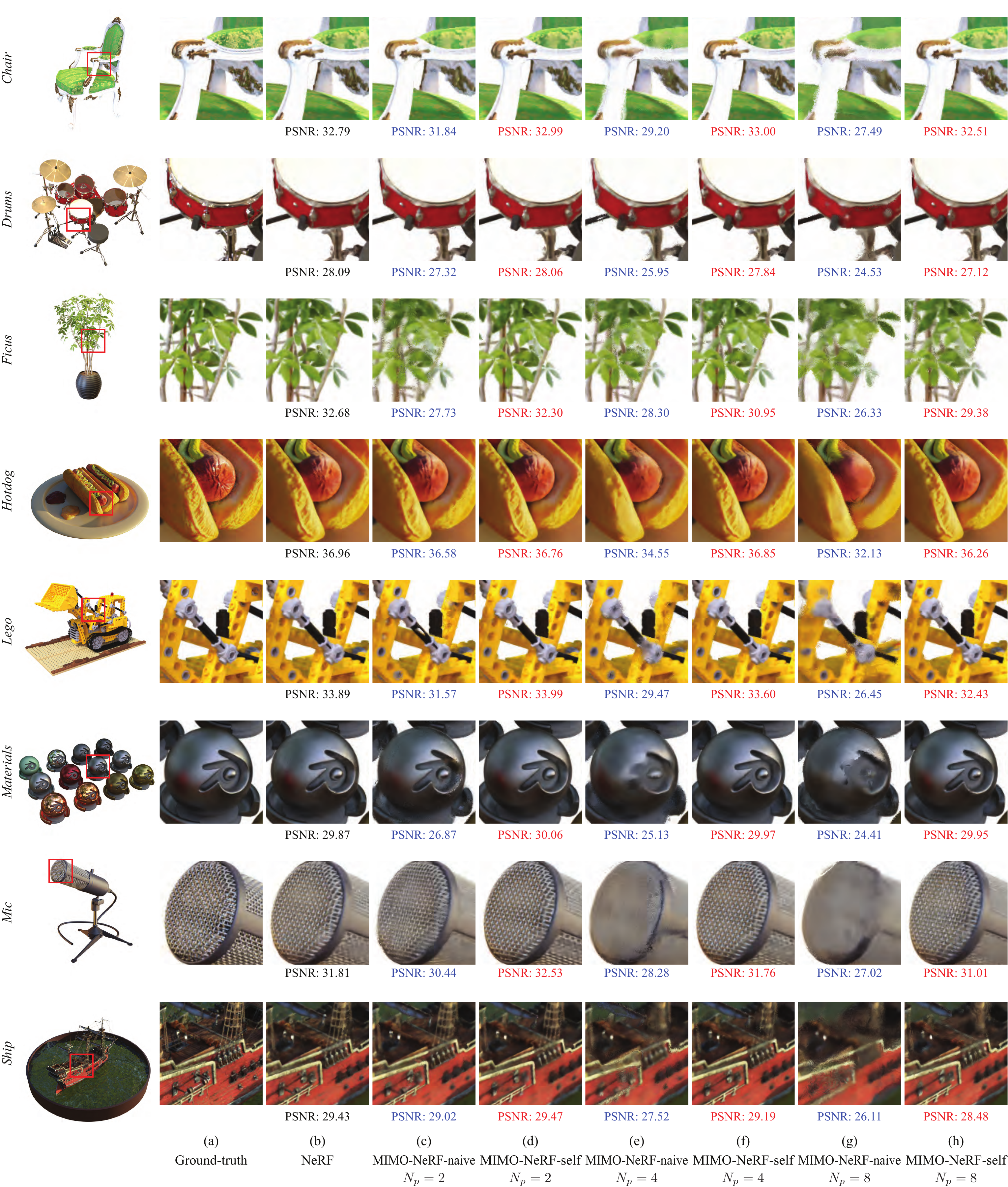}
  \caption{Qualitative comparison between NeRF, MIMO-NeRF-naive, and MIMO-NeRF-self on the Blender dataset.
    This figure is an extension of Figure~\ref{fig:results_nerf}.
    We report PSNR for the displayed view.
    The average scores for all views are presented in Table~\ref{tab:scores_nerf_full_each}.
    As shown in (c), (e), and (g), the deterioration of image quality becomes obvious in MIMO-NeRF-naive as $N_p$ increases.
    In contrast, MIMO-NeRF-self is resistant to this deterioration, as shown in (d), (f), and (h).}
  \label{fig:results_nerf_blender}
\end{figure*}

\begin{figure*}[t]
  \centering
  \includegraphics[width=\textwidth]{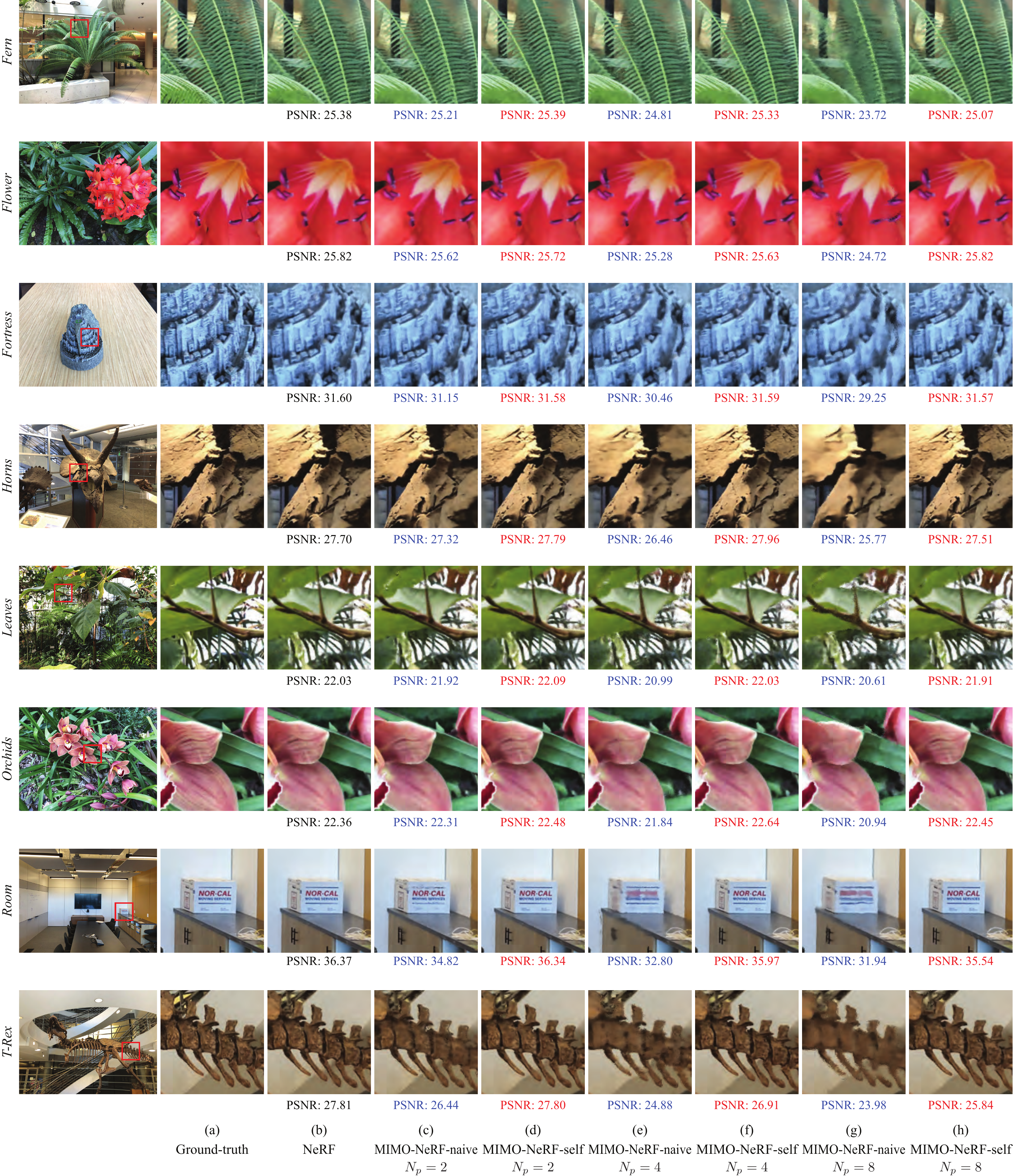}
  \caption{Qualitative comparison between NeRF, MIMO-NeRF-naive, and MIMO-NeRF-self on the LLFF dataset.
    This figure is an extension of Figure~\ref{fig:results_nerf}.
    We report PSNR for the displayed view.
    The average scores for all views are listed in Table~\ref{tab:scores_nerf_full_each}.
    As shown in (c), (e), and (g), the deterioration of image quality becomes obvious in MIMO-NeRF-naive as $N_p$ increases.
    In contrast, MIMO-NeRF-self is robust against this deterioration, as shown in (d), (f), and (h).}
  \label{fig:results_nerf_llff}
\end{figure*}

\smallskip\noindent\textbf{Qualitative results.}
Figures~\ref{fig:results_nerf_blender} and \ref{fig:results_nerf_llff} present the qualitative results for the Blender and LLFF datasets, respectively.
Examples of the synthesized videos are provided on the \href{https://www.kecl.ntt.co.jp/people/kaneko.takuhiro/projects/mimo-nerf/}{project page}.\footnoteref{foot:project_page}

\begin{table*}
  \setlength{\tabcolsep}{0.5pt}
  \begin{center}
    \scriptsize
    \begin{tabularx}{\textwidth}{lc|CCCCCCC|CCCCCCC}
      & & \multicolumn{7}{c|}{Blender} & \multicolumn{7}{c}{LLFF}
      \\
      \multicolumn{1}{c}{Model} & $N_p$ & PSNR$\uparrow$ & SSIM$\uparrow$ & LPIPS$\downarrow$ & \# Run$\downarrow$ & I-time$\downarrow$ & T-time$\downarrow$ & \# Params & PSNR$\uparrow$ & SSIM$\uparrow$ & LPIPS$\downarrow$ & \# Run$\downarrow$ & I-time$\downarrow$ & T-time$\downarrow$ & \# Params \\
      & & & & & & (s) & (h) & (M) & & & & & (s) & (h) & (M) 
      \\ \hline
      NeRF~\cite{BMildenhallECCV2020} & 1
      & 30.94 & 0.946 & 0.070 & 256 & 38.22 & 12.54 & 1.19
      & 26.45 & 0.811 & 0.249 & 256 & 45.46 & 16.22 & 1.19
      \\ \hline
      MIMO-NeRF-naive & \multirow{3}{*}{2}
      & 29.36 & 0.932 & 0.091 & 128 & 20.67 & \first{8.61} & 1.26
      & 26.00 & 0.796 & \third{0.269} & 128 & 24.82 & \first{8.67} & 1.26
      \\
      MIMO-NeRF-distill &
      & \third{30.55} & \third{0.943} & \third{0.077} & 128 & 20.67 & 25.27 & 1.26
      & \third{26.21} & \third{0.799} & 0.278 & 128 & 24.82 & 30.79 & 1.26
      \\
      MIMO-NeRF-self &
      & \first{31.01} & \first{0.947} & \first{0.071} & 128 & 20.67 & \third{14.13} & 1.26
      & \first{26.46} & \first{0.812} & \first{0.253} & 128 & 24.82 & \third{17.51} & 1.26
      \\ \hline
      MIMO-NeRF-naive & \multirow{3}{*}{4}
      & 27.72 & 0.914 & 0.114 & 64 & 11.17 & \first{5.95} & 1.39
      & 25.09 & 0.758 & 0.320 & 64 & 13.47 & \first{4.87} & 1.39
      \\
      MIMO-NeRF-distill &
      & \third{30.01} & \third{0.939} & \third{0.083} & 64 & 11.17 & 22.58 & 1.39
      & \third{26.14} & \third{0.798} & \third{0.279} & 64 & 13.47 & 26.97 & 1.39
      \\
      MIMO-NeRF-self &
      & \first{30.66} & \first{0.944} & \first{0.075} & 64 & 11.17 & \third{12.37} & 1.39
      & \first{26.35} & \first{0.809} & \first{0.258} & 64 & 13.47 & \third{14.52} & 1.39
      \\ \hline
      MIMO-NeRF-naive & \multirow{3}{*}{8}
      & 25.78 & 0.889 & 0.145 & 32 & 6.62 & \first{5.08} & 1.65
      & 24.15 & 0.716 & 0.376 & 32 & 8.01 & \first{3.25} & 1.65
      \\
      MIMO-NeRF-distill &
      & \third{28.85} & \third{0.929} & \third{0.095} & 32 & 6.62 & 21.74 & 1.65
      & \third{25.91} & \third{0.793} & \third{0.285} & 32 & 8.01 & 25.34 & 1.65
      \\
      MIMO-NeRF-self &
      & \first{29.92} & \first{0.938} & \first{0.084} & 32 & 6.62 & \third{14.95} & 1.65
      & \first{25.99} & \first{0.800} & \first{0.270} & 32 & 8.01 & \third{19.16} & 1.65
      \\ \hline \hline
      NeRF~\cite{BMildenhallECCV2020} & 1
      & 31.01 & 0.947 & 0.081 & -- & -- & -- & --
      & 26.50 & 0.811 & 0.250 & -- & -- & -- & --
      \\
    \end{tabularx}
  \end{center}
  \caption{Benchmark performance of MIMO-NeRFs for full-sized images.
    The scores for the model with citation~\cite{BMildenhallECCV2020} are taken from another report~\cite{BMildenhallECCV2020}.
    We provide them as references.
    The other scores were calculated in our environment.
    We implemented all the models based on the commonly-used source code of NeRF.
    See Appendix~\ref{subsec:implementation_details_nerf} for the implementation details.
    The PSNR, SSIM, and LPIPS for each scene are provided in Table~\ref{tab:scores_nerf_full_each}.}
  \label{tab:scores_nerf_full}
\end{table*}

\begin{table*}
  \setlength{\tabcolsep}{0.5pt}
  \begin{center}
    \scriptsize
    \begin{tabularx}{\textwidth}{lc|CCCCCCCC|C|CCCCCCCC|C}
      & & \multicolumn{18}{c}{PSNR$\uparrow$}
      \\
      & & \multicolumn{9}{c|}{Blender} & \multicolumn{9}{c}{LLFF}
      \\
      \multicolumn{1}{c}{Model} & $N_p$ & Chair & Drums & Ficus & Hotdog & Lego & Materials & Mic & Ship & Avg. & Fern & Flower & Fortress & Horns & Laves & Orchids & Room & T-Rex & Avg.
      \\ \hline
      NeRF & 1
      & 32.82 & 25.04 & 30.10 & 36.28 & 32.60 & 29.63 & 32.77 & 28.32 & 30.94
      & 24.99 & 27.57 & 31.16 & 27.33 & 20.96 & 20.35 & 32.57 & 26.64 & 26.45
      \\ \hline
      MIMO-NeRF-naive & \multirow{3}{*}{2}
      & 31.82 & 24.42 & 25.59 & \third{35.72} & 30.86 & 27.13 & 31.50 & \thirdr{27.88} & 29.36
      & 24.76 & \first{27.50} & \third{30.75} & 26.68 & 20.88 & 20.27 & 31.62 & 25.49 & 26.00
      \\
      MIMO-NeRF-distill &
      & \third{32.35} & \third{25.10} & \first{29.78} & 35.37 & \third{31.74} & \third{29.43} & \third{32.74} & 27.86 & \thirdr{30.55}
      & \third{24.97} & 27.39 & 30.59 & \third{26.87} & \third{20.89} & \first{20.52} & \third{32.17} & \thirdr{26.29} & \third{26.21}
      \\
      MIMO-NeRF-self &
      & \first{32.92} & \first{25.17} & \third{29.49} & \first{36.10} & \first{32.60} & \first{30.06} & \first{33.18} & \firstr{28.53} & \firstr{31.01}
      & \first{25.05} & \third{27.42} & \first{31.24} & \first{27.24} & \first{21.00} & \third{20.49} & \first{32.52} & \firstr{26.72} & \first{26.46}
      \\ \hline
      MIMO-NeRF-naive & \multirow{3}{*}{4}
      & 29.21 & 23.06 & 25.39 & 34.20 & 28.33 & 26.18 & 28.93 & 26.48 & 27.72
      & 24.31 & 27.01 & 29.98 & 25.41 & 20.11 & 19.81 & 29.96 & 24.11 & 25.09
      \\
      MIMO-NeRF-distill &
      & \third{32.07} & \third{24.75} & \third{27.85} & \third{35.20} & \third{31.20} & \third{29.26} & \third{32.20} & \thirdr{27.53} & \thirdr{30.01}
      & \third{24.87} & \third{27.40} & \third{30.60} & \third{26.79} & \third{20.86} & \third{20.44} & \third{32.06} & \first{26.07} & \third{26.14}
      \\
      MIMO-NeRF-self &
      & \first{32.84} & \first{24.82} & \first{28.44} & \first{36.18} & \first{32.29} & \first{29.87} & \first{32.66} & \firstr{28.19} & \firstr{30.66}
      & \first{24.95} & \first{27.52} & \first{31.24} & \first{27.25} & \first{20.98} & \first{20.47} & \first{32.37} & \thirdr{26.03} & \first{26.35}
      \\ \hline
      MIMO-NeRF-naive & \multirow{3}{*}{8}
      & 27.17 & 21.33 & 23.38 & 32.01 & 25.32 & 25.16 & 27.25 & 24.60 & 25.78
      & 23.06 & 26.06 & 28.76 & 24.51 & 19.85 & 18.79 & 29.11 & 23.10 & 24.15
      \\
      MIMO-NeRF-distill &
      & \third{31.40} & \third{23.61} & \third{25.38} & \third{34.78} & \third{29.48} & \third{28.90} & \third{30.58} & \thirdr{26.70} & \thirdr{28.85}
      & \third{24.54} & \third{27.32} & \third{30.56} & \third{26.54} & \third{20.80} & \first{20.23} & \third{31.71} & \firstr{25.55} & \third{25.91}
      \\
      MIMO-NeRF-self &
      & \first{32.37} & \first{24.18} & \first{26.91} & \first{35.64} & \first{31.17} & \first{29.96} & \first{31.59} & \firstr{27.58} & \firstr{29.92}
      & \first{24.58} & \first{27.57} & \first{31.13} & \first{26.66} & \first{20.82} & \first{20.23} & \first{31.84} & \thirdr{25.12} & \first{25.99}
      \\ \hline
      NeRF~\cite{BMildenhallECCV2020} & 1
      & 33.00 & 25.01 & 30.13 & 36.18 & 32.54 & 29.62 & 32.91 & 28.65 & 31.01
      & 25.17 & 27.40 & 31.16 & 27.45 & 20.92 & 20.36 & 32.70 & 26.80 & 26.50
      \\
      \\
      & & \multicolumn{18}{c}{SSIM$\uparrow$}
      \\
      & & \multicolumn{9}{c|}{Blender} & \multicolumn{9}{c}{LLFF}
      \\
      \multicolumn{1}{c}{Model} & $N_p$ & Chair & Drums & Ficus & Hotdog & Lego & Materials & Mic & Ship & Avg. & Fern & Flower & Fortress & Horns & Laves & Orchids & Room & T-Rex & Avg.
      \\ \hline
      NeRF~\cite{BMildenhallECCV2020} & 1
      & 0.966 & 0.924 & 0.962 & 0.975 & 0.962 & 0.949 & 0.980 & 0.852 & 0.946
      & 0.790 & 0.832 & 0.881 & 0.826 & 0.690 & 0.644 & 0.951 & 0.878 & 0.811
      \\ \hline
      MIMO-NeRF-naive & \multirow{3}{*}{2}
      & 0.957 & 0.914 & 0.918 & 0.973 & 0.949 & 0.921 & 0.973 & 0.847 & 0.932
      & 0.781 & \third{0.825} & \third{0.863} & 0.799 & \third{0.684} & 0.625 & 0.944 & 0.847 & 0.796
      \\
      MIMO-NeRF-distill &
      & \third{0.962} & \first{0.926} & \first{0.960} & \third{0.969} & \third{0.954} & \third{0.949} & \third{0.980} & \thirdr{0.844} & \thirdr{0.943}
      & \third{0.783} & 0.818 & 0.855 & \third{0.802} & 0.680 & \third{0.640} & \third{0.947} & \thirdr{0.869} & \third{0.799}
      \\
      MIMO-NeRF-self &
      & \first{0.967} & \third{0.925} & \third{0.958} & \first{0.975} & \first{0.962} & \first{0.954} & \first{0.982} & \firstr{0.853} & \firstr{0.947}
      & \first{0.791} & \first{0.827} & \first{0.882} & \first{0.822} & \first{0.695} & \first{0.646} & \first{0.950} & \firstr{0.881} & \first{0.812}
      \\ \hline
      MIMO-NeRF-naive & \multirow{3}{*}{4}
      & 0.927 & 0.891 & 0.919 & 0.964 & 0.920 & 0.913 & 0.956 & 0.822 & 0.914
      & 0.758 & 0.801 & 0.824 & 0.742 & 0.630 & 0.589 & 0.923 & 0.801 & 0.758
      \\
      MIMO-NeRF-distill &
      & \third{0.959} & \third{0.920} & \third{0.947} & \third{0.968} & \third{0.950} & \third{0.947} & \third{0.977} & \thirdr{0.840} & \thirdr{0.939}
      & \third{0.780} & \third{0.819} & \third{0.857} & \third{0.803} & \third{0.678} & \third{0.636} & \third{0.946} & \thirdr{0.866} & \third{0.798}
      \\
      MIMO-NeRF-self &
      & \first{0.967} & \first{0.921} & \first{0.949} & \first{0.975} & \first{0.960} & \first{0.953} & \first{0.979} & \firstr{0.850} & \firstr{0.944}
      & \first{0.787} & \first{0.830} & \first{0.882} & \first{0.823} & \first{0.693} & \first{0.644} & \first{0.948} & \firstr{0.869} & \first{0.809}
      \\ \hline
      MIMO-NeRF-naive & \multirow{3}{*}{8}
      & 0.900 & 0.858 & 0.891 & 0.951 & 0.876 & 0.898 & 0.946 & 0.794 & 0.889
      & 0.701 & 0.760 & 0.771 & 0.700 & 0.610 & 0.528 & 0.907 & 0.754 & 0.716
      \\
      MIMO-NeRF-distill &
      & \third{0.953} & \third{0.905} & \third{0.923} & \third{0.966} & \third{0.940} & \third{0.944} & \third{0.972} & \thirdr{0.830} & \thirdr{0.929}
      & \third{0.772} & \third{0.818} & \third{0.856} & \third{0.797} & \third{0.677} & \third{0.625} & \third{0.943} & \firstr{0.853} & \third{0.793}
      \\
      MIMO-NeRF-self &
      & \first{0.963} & \first{0.911} & \first{0.934} & \first{0.973} & \first{0.952} & \first{0.953} & \first{0.975} & \firstr{0.842} & \firstr{0.938}
      & \first{0.773} & \first{0.828} & \first{0.879} & \first{0.809} & \first{0.686} & \first{0.629} & \first{0.944} & \thirdr{0.848} & \first{0.800}
      \\ \hline \hline
      NeRF~\cite{BMildenhallECCV2020} & 1
      & 0.967 & 0.925 & 0.964 & 0.974 & 0.961 & 0.949 & 0.980 & 0.856 & 0.947
      & 0.792 & 0.827 & 0.881 & 0.828 & 0.690 & 0.641 & 0.948 & 0.880 & 0.811
      \\
      \\
      & & \multicolumn{18}{c}{LPIPS$\downarrow$}
      \\
      & & \multicolumn{9}{c|}{Blender} & \multicolumn{9}{c}{LLFF}
      \\
      \multicolumn{1}{c}{Model} & $N_p$ & Chair & Drums & Ficus & Hotdog & Lego & Materials & Mic & Ship & Avg. & Fern & Flower & Fortress & Horns & Laves & Orchids & Room & T-Rex & Avg.
      \\ \hline
      NeRF~\cite{BMildenhallECCV2020} & 1
      & 0.046 & 0.091 & 0.045 & 0.045 & 0.048 & 0.063 & 0.026 & 0.198 & 0.070
      & 0.281 & 0.214 & 0.173 & 0.273 & 0.312 & 0.314 & 0.173 & 0.254 & 0.249
      \\ \hline
      MIMO-NeRF-naive & \multirow{3}{*}{2}
      & 0.057 & 0.107 & 0.108 & \third{0.047} & 0.067 & 0.103 & 0.035 & \thirdr{0.205} & 0.091
      & \third{0.294} & \first{0.222} & \third{0.201} & \third{0.303} & \third{0.319} & 0.338 & \third{0.188} & 0.284 & \third{0.269}
      \\
      MIMO-NeRF-distill &
      & \third{0.052} & \third{0.091} & \first{0.049} & 0.057 & \third{0.060} & \third{0.062} & \third{0.024} & 0.220 & \thirdr{0.077}
      & 0.311 & 0.247 & 0.225 & 0.318 & 0.327 & \third{0.333} & 0.189 & \thirdr{0.274} & 0.278
      \\
      MIMO-NeRF-self &
      & \first{0.045} & \first{0.091} & \third{0.055} & \first{0.046} & \first{0.050} & \first{0.057} & \first{0.023} & \firstr{0.203} & \firstr{0.071}
      & \first{0.283} & \third{0.225} & \first{0.175} & \first{0.284} & \first{0.311} & \first{0.319} & \first{0.176} & \firstr{0.252} & \first{0.253}
      \\ \hline
      MIMO-NeRF-naive & \multirow{3}{*}{4}
      & 0.088 & 0.141 & 0.104 & 0.062 & 0.110 & 0.107 & 0.063 & 0.239 & 0.114
      & 0.321 & 0.255 & 0.272 & 0.377 & 0.374 & 0.386 & 0.236 & 0.336 & 0.320
      \\
      MIMO-NeRF-distill &
      & \third{0.054} & \third{0.100} & \third{0.069} & \third{0.058} & \third{0.066} & \third{0.064} & \third{0.027} & \thirdr{0.223} & \thirdr{0.083}
      & \third{0.311} & \third{0.245} & \third{0.222} & \third{0.317} & \third{0.329} & \third{0.339} & \third{0.192} & \thirdr{0.276} & \third{0.279}
      \\
      MIMO-NeRF-self &
      & \first{0.045} & \first{0.098} & \first{0.069} & \first{0.046} & \first{0.052} & \first{0.059} & \first{0.026} & \firstr{0.205} & \firstr{0.075}
      & \first{0.288} & \first{0.221} & \first{0.176} & \first{0.284} & \first{0.314} & \first{0.325} & \first{0.182} & \firstr{0.272} & \first{0.258}
      \\ \hline
      MIMO-NeRF-naive & \multirow{3}{*}{8}
      & 0.112 & 0.181 & 0.134 & 0.096 & 0.163 & 0.123 & 0.079 & 0.273 & 0.145
      & 0.386 & 0.323 & 0.348 & 0.427 & 0.402 & 0.445 & 0.281 & 0.396 & 0.376
      \\
      MIMO-NeRF-distill &
      & \third{0.060} & \third{0.123} & \third{0.095} & \third{0.061} & \third{0.080} & \third{0.069} & \third{0.037} & \thirdr{0.236} & \thirdr{0.095}
      & \third{0.318} & \third{0.245} & \third{0.222} & \third{0.322} & \third{0.334} & \third{0.352} & \third{0.201} & \firstr{0.289} & \third{0.285}
      \\
      MIMO-NeRF-self &
      & \first{0.048} & \first{0.114} & \first{0.089} & \first{0.050} & \first{0.066} & \first{0.061} & \first{0.033} & \firstr{0.215} & \firstr{0.084}
      & \first{0.303} & \first{0.227} & \first{0.179} & \first{0.301} & \first{0.323} & \first{0.345} & \first{0.190} & \thirdr{0.292} & \first{0.270}
      \\ \hline \hline
      NeRF~\cite{BMildenhallECCV2020} & 1
      & 0.046 & 0.091 & 0.044 & 0.121 & 0.050 & 0.063 & 0.028 & 0.206 & 0.081
      & 0.280 & 0.219 & 0.171 & 0.268 & 0.316 & 0.321 & 0.178 & 0.249 & 0.250
      \\
    \end{tabularx}
  \end{center}
  \caption{Comparison of PSNR, SSIM, and LPIPS for each scene on the Blender and LLFF datasets with full-sized images.
    The scores for the model with citation~\cite{BMildenhallECCV2020} are taken from another report~\cite{BMildenhallECCV2020}.
    We provide them as references.
    The other scores were calculated in our environment.
    We implemented all the models based on the commonly-used source code of NeRF.
    See Appendix~\ref{subsec:implementation_details_nerf} for the implementation details.
    The scores for the other metrics are provided in Table~\ref{tab:scores_nerf_full}.}
  \label{tab:scores_nerf_full_each}
\end{table*}

\clearpage
\clearpage
\subsection{Comparison with AutoInt}
\label{subsec:comparison_with_autoint}

To further clarify the utility of MIMO-NeRF, we compared it with AutoInt~\cite{DLindellCVPR2021}, which reduces the number of MLPs running (\#~Run) using an integral network that calculates the colors and volume densities \textit{per segment} instead of \textit{per point}.
In particular, we investigated the difference in performance between MIMO-NeRF-self and AutoInt when \#~Run was the same.

\smallskip\noindent\textbf{Results.}
Table~\ref{tab:scores_autoint} summarizes these results.
The model was evaluated using the Blender dataset (full-size images).
It can be observed that MIMO-NeRF-self outperformed AutoInt in most cases.
Another important difference is that AutoInt requires the use of a specific and complex grad network during training, whereas MIMO-NeRF can be trained using a standard network such as that implemented using PyTorch.

\begin{table}[h]
  \setlength{\tabcolsep}{6pt}
  \begin{center}
    \scriptsize
    \begin{tabularx}{\columnwidth}{l|cccc}
      \multicolumn{1}{c|}{Model} & \# Run$\downarrow$ & PSNR$\uparrow$ & SSIM$\uparrow$ & LPIPS$\downarrow$
      \\ \hline
      AutoInt ($N = 32$)~\cite{DLindellCVPR2021} & 32
      & 26.83 & 0.926 & 0.151
      \\
      MIMO-NeRF-self ($N_p = 8$) & 32
      & \first{29.92} & \first{0.938} & \first{0.084}
      \\ \hline
      AutoInt ($N = 16$)~\cite{DLindellCVPR2021} & 16
      & 26.04 & 0.916 & 0.167
      \\
      MIMO-NeRF-self ($N_p = 16$) & 16
      & \first{28.69} & \first{0.925} & \first{0.099}
      \\ \hline
      AutoInt ($N = 8$)~\cite{DLindellCVPR2021} & 8
      & 25.55 & \first{0.911} & 0.170
      \\
      MIMO-NeRF-self ($N_p = 32$) & 8
      & \first{27.19} & 0.908 & \first{0.118}
      \\
    \end{tabularx}
  \end{center}
  \caption{Comparison of AutoInt and MIMO-NeRF-self.
    We compared AutoInt and MIMO-NeRF-self when \#~Run was the same.
    We evaluated the models on the Blender dataset (full-sized images).
    The scores for AutoInt are taken from the AutoInt paper~\cite{DLindellCVPR2021}.
    In most cases, MIMO-NeRF-self outperforms AutoInt.}
  \label{tab:scores_autoint}
\end{table}

\clearpage
\clearpage
\subsection{Detailed analysis of application to DONeRF}
\label{subsec:detailed_analysis_donerf}

In Section~\ref{subsec:application_donerf}, we compared MIMO-DONeRF-16/4-naive and MIMO-DONeRF-16/4-self with \textit{DONeRF-16}, in which the number of selected samples ($N_s$) is the same as that of MIMO-DONeRF-16/4-naive and MIMO-DONeRF-16/4-self (i.e., $N_s = 16$), and \textit{DONeRF-4}, in which the number of MLPs running (\#~Run) is the same as that of MIMO-DONeRF-16/4-naive and MIMO-DONeRF-16/4-self (i.e., $\text{\#~Run}=5$).
For further analysis, this appendix provides a comparison with \textit{DONeRF-11}, in which the training time (T-time) is almost the same as that of MIMO-DONeRF-16/4-self, and \textit{DONeRF-5}, in which the inference time (I-time) is close to (more strictly, slightly longer than) that of MIMO-DONeRF-16/4-naive and MIMO-DONeRF-16/4-self.
We evaluated the models using the same metrics as those described in Section~\ref{subsec:application_donerf}.

\smallskip\noindent\textbf{Quantitative results.}
Table~\ref{tab:scores_donerf_ex} summarizes the results for all metrics.
Table~\ref{tab:scores_donerf_each} lists the PSNR and FLIP for each scene.
Our findings are as follows:

\smallskip\noindent\textit{MIMO-DONeRF-16/4-naive vs. DONeRF-5 (close I-time).}
We found that MIMO-DONeRF-16/4-naive outperformed or was comparable to DONeRF-5 in terms of PSNR and FLIP for all scenes.
MIMO-DONeRF-16/4-naive also slightly outperformed DONeRF-5 in terms of the I-time and T-time.
Therefore, MIMO-DONeRF-16/4-naive does not have any disadvantages compared to MIMO-DONeRF-5.

\smallskip\noindent\textit{MIMO-DONeRF-16/4-self vs. DONeRF-11 (close T-time).}
MIMO-DONeRF-16/4-self and DONeRF-11 were comparable in terms of average PSNR and FLIP, and whether they were better or worse depended on the view and metrics.
Although T-time was almost the same between these two models, MIMO-DONeRF-16/4 outperforms DONeRF-11 significantly in terms of I-time (approximately half of it).
Overall, MIMO-DONeRF-16/4-self is better than DONeRF-11 in terms of significantly better I-time.

\smallskip\noindent\textit{Summary.}
Even when considering the models in which I-time is close to that of MIMO-DONeRFs and T-time is close to that of MIMO-DONeRF-self, the results indicate that MIMO-DONeRFs have advantages.
As discussed in Section~\ref{subsec:application_donerf}, the results suggest that an increase in $N_p$ (i.e., the replacement of the SISO MLP by the MIMO MLP) can be used as a better alternative to a reduction in $N_s$ (the number of selected samples) when seeking a better trade-off between speed and quality.

\begin{figure*}[t]
  \centering
  \includegraphics[width=0.86\textwidth]{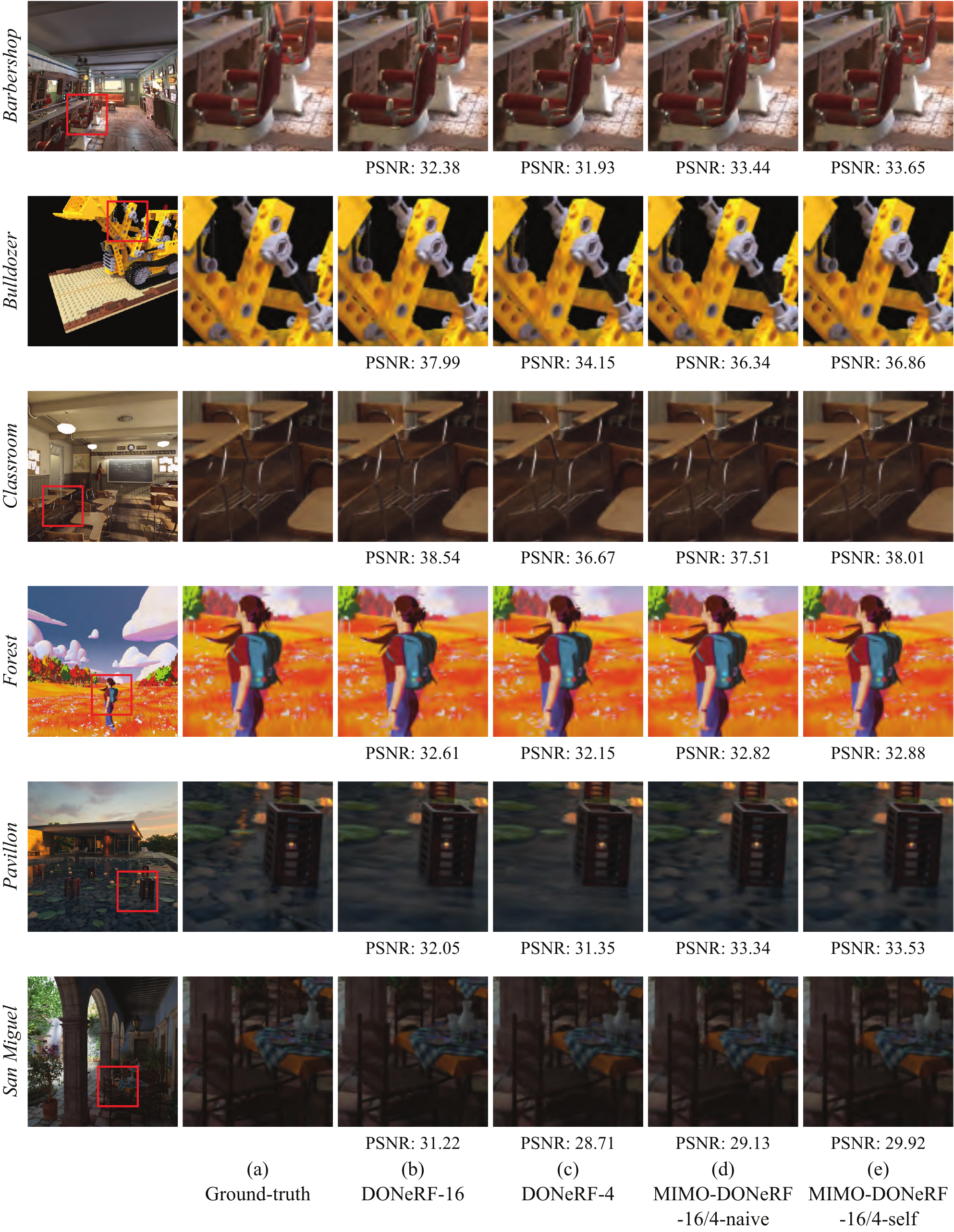}
  \caption{Qualitative comparison between DONeRF-16, DONeRF-4, MIMO-DONeRF-16/4-naive, and MIMO-DONeRF-16/4-self.
    Best viewed zoomed in.
    We report PSNR for the displayed view.
    The average scores for all views are given in Table~\ref{tab:scores_donerf_each}.
    DONeRF-4 (c) sometimes yields artifacts, e.g., for the belt in the ``Bulldozer'' scene or for the hair in the ``Forest'' scene.
    DONeRF-16 (b), MIMO-DONeRF-16/4-naive (d), and MIMO-DONeRF-16/4-self (e) mitigate this defect by increasing the number of samples.
    It should be noted that DONeRF-16 increases the inference time approximately three times, while MIMO-DONeRF-16/4-naive and MIMO-DONeRF-16/4--self only increase the inference time $1.1$ times.
    Another interesting finding is that MIMO-DONeRF-16/4-naive (d) and MIMO-DONeRF-16/4-self (e) succeed in representing lotus leaves in the ``Pavillon'' scene, whereas DONeRF-16 (b) and DONeRF-4 (c) fail to do so.
    The possible reason is that MIMO-NeRFs can accumulate neighbor information using grouped samples, and this provides a positive effect.}
  \label{fig:results_donerf}
\end{figure*}

\smallskip\noindent\textbf{Qualitative results.}
Figure~\ref{fig:results_donerf} shows the qualitative results.
Examples of the synthesized videos are provided on the \href{https://www.kecl.ntt.co.jp/people/kaneko.takuhiro/projects/mimo-nerf/}{project page}.\footnoteref{foot:project_page}

\begin{table*}
  \setlength{\tabcolsep}{3pt}
  \begin{center}
    \scriptsize
    \begin{tabularx}{0.7\textwidth}{lcc|CCCCCC}
      \multicolumn{1}{c}{Model} & $N_s$ & $N_p$ & PSNR$\uparrow$ & FLIP$\downarrow$ & \# Run$\downarrow$ & I-time$\downarrow$ & T-time$\downarrow$ & \# Params
      \\
      & & & & & & (s) & (h) & (M)
      \\ \hline
      DONeRF-4
      & 4 & 1 & 31.21 & 0.070 & 5 & \first{0.140} & \first{3.23} & 0.94
      \\
      DONeRF-5
      & 5 & 1 & 31.65 & 0.067 & 6 & 0.164 & \third{3.29} & 0.94
      \\
      DONeRF-11
      & 11 & 1 & \second{32.76} & \third{0.063} & 12 & 0.304 & 3.57 & 0.94
      \\
      DONeRF-16
      & 16 & 1 & \first{33.06} & \first{0.061} & 17 & 0.429 & 3.79 & 0.94
      \\ \hline
      MIMO-DONeRF-16/4-naive
      & 16 & 4 & 32.30 & \third{0.063} & 5 & \second{0.155} & \second{3.26} & 0.99
      \\
      MIMO-DONeRF-16/4-self
      & 16 & 4 & \third{32.72} & \first{0.061} & 5 & \second{0.155} & 3.56 & 0.99
      \\ \hline \hline
      DONeRF-4~\cite{TNeffCVF2021}
      & 4 & 1 & 31.14 & 0.071 & -- & -- & -- & --
      \\
      DONeRF-16~\cite{TNeffCVF2021}
      & 16 & 1 & 33.03 & 0.062 & -- & -- & -- & --
      \\
    \end{tabularx}
  \end{center}
  \caption{Comparison of quantitative scores between DONeRFs and MIMO-DONeRFs.
    The scores for the model with citation~\cite{TNeffCVF2021} are taken from another report~\cite{TNeffCVF2021}.
    We provide them as references.
    The other scores were calculated in our environment.
    We implemented all the models based on the official DONeRF source code.
    See Appendix~\ref{subsec:implementation_details_donerf} for the implementation details.
    This table is an extended version of Table~\ref{tab:scores_donerf}.
    In addition to the scores provided in Table~\ref{tab:scores_donerf}, this table provides the scores for DONeRF-5, in which I-time is close to those of MIMO-DONeRF-16/4-naive and MIMO-DONeRF-16/4-self, and DONeRF-11, in which T-time is close to that of MIMO-DONeRF-16/4-self.
    The PSNR and FLIP for each scene are presented in Table~\ref{tab:scores_donerf_each}.}
  \label{tab:scores_donerf_ex}
\end{table*}

\begin{table*}
  \setlength{\tabcolsep}{2pt}
  \begin{center}
    \scriptsize
    \begin{tabularx}{0.75\textwidth}{lcc|CCCCCC|C}
      & & & \multicolumn{7}{c}{PSNR$\uparrow$}
      \\
      \multicolumn{1}{c}{Model} & $N_s$ & $N_p$ & Barbershop & Bulldozer & Classroom & Forest & Pavillon & San Miguel & Avg.
      \\ \hline
      DONeRF-4
      & 4 & 1
      & 30.76 & 33.29 & 34.03 & 30.90 & 31.02 & 27.24 & 31.21
      \\
      DONeRF-5
      & 5 & 1
      & 31.14 & 34.33 & 34.52 & 31.12 & 31.22 & 27.58 & 31.65
      \\
      DONeRF-11
      & 11 & 1
      & \third{31.90} & \second{36.37} & \second{35.90} & \first{31.80} & 31.60 & \secondr{28.98} & \second{32.76}
      \\
      DONeRF-16
      & 16 & 1
      & \first{32.13} & \first{36.87} & \first{36.15} & \second{31.79} & \third{31.71} & \firstr{29.71} & \first{33.06}
      \\ \hline
      MIMO-DONeRF-16/4-naive
      & 16 & 4
      & 31.60 & 35.14 & 35.19 & 31.61 & \second{32.50} & 27.77 & 32.30
      \\
      MIMO-DONeRF-16/4-self
      & 16 & 4
      & \second{32.11} & \third{35.57} & \third{35.65} & \third{31.76} & \first{32.80} & \thirdr{28.41} & \third{32.72}
      \\ \hline \hline
      DONeRF-4~\cite{TNeffCVF2021}
      & 4 & 1
      & 30.84 & 33.46 & 33.43 & 30.63 & 31.07 & 27.41 & 31.14
      \\
      DONeRF-16~\cite{TNeffCVF2021}
      & 16 & 1
      & 32.15 & 36.98 & 36.27 & 31.32 & 31.79 & 29.67 & 33.03
      \\
      \\
      & & & \multicolumn{7}{c}{FLIP$\downarrow$}
      \\
      \multicolumn{1}{c}{Model} & $N_s$ & $N_p$ & Barbershop & Bulldozer & Classroom & Forest & Pavillon & San Miguel & Avg.
      \\ \hline
      DONeRF-4
      & 4 & 1
      & 0.064 & 0.044 & 0.053 & 0.075 & 0.099 & 0.083 & 0.070
      \\
      DONeRF-5
      & 5 & 1
      & 0.064 & 0.040 & 0.051 & 0.074 & 0.097 & 0.078 & 0.067
      \\
      DONeRF-11
      & 11 & 1
      & \third{0.059} & \second{0.033} & \second{0.047} & \first{0.071} & 0.096 & \secondr{0.070} & \third{0.063}
      \\
      DONeRF-16
      & 16 & 1
      & \second{0.058} & \first{0.032} & \second{0.047} & \third{0.072} & \third{0.095} & \firstr{0.065} & \first{0.061}
      \\ \hline
      MIMO-DONeRF-16/4-naive
      & 16 & 4
      & \third{0.059} & 0.036 & 0.049 & \first{0.071} & \second{0.087} & 0.078 & \third{0.063}
      \\
      MIMO-DONeRF-16/4-self
      & 16 & 4
      & \first{0.056} & \third{0.035} & \first{0.045} & \third{0.072} & \first{0.086} & \thirdr{0.072} & \first{0.061}
      \\ \hline \hline
      DONeRF-4~\cite{TNeffCVF2021}
      & 4 & 1
      & 0.065 & 0.048 & 0.058 & 0.077 & 0.098 & 0.080 & 0.071
      \\
      DONeRF-16~\cite{TNeffCVF2021}
      & 16 & 1
      & 0.059 & 0.036 & 0.045 & 0.074 & 0.094 & 0.065 & 0.062
      \\
    \end{tabularx}
  \end{center}
  \caption{Comparison of PSNR and FLIP for each scene between DONeRFs and MIMO-DONeRFs.
    The scores for the model with citation~\cite{TNeffCVF2021} are taken from another report~\cite{TNeffCVF2021}.
    We provide them as references.
    The other scores were calculated in our environment.
    We implemented all the models based on the official DONeRF source code.
    See Appendix~\ref{subsec:implementation_details_donerf} for the implementation details.
    The scores for the other metrics are listed in Table~\ref{tab:scores_donerf_ex}.}
  \label{tab:scores_donerf_each}
\end{table*}

\clearpage
\clearpage
\subsection{Detailed analysis of application to TensoRF}
\label{subsec:detailed_analysis_tensorf}

In Section~\ref{subsec:application_tensorf}, we used the variants of TensoRF that achieved the best image quality as baselines.
Specifically, we used \textit{TensoRF-VM-192-30k} ($R_{\sigma} = 16$, $R_c = 48$, and the iteration of $30k$) for the Blender dataset and used \textit{TensoRF-VM-96} ($R_{\sigma, 1} = R_{\sigma, 2} = 4$, $R_{\sigma, 3} = 16$, $R_{c, 1} = R_{c, 2} = 12$, and $R_{c, 3} = 48$) for the LLFF dataset.
As models with faster training but lower quality, a previous study~\cite{AChenECCV2022} also presented \textit{TensoRF-VM-48} ($R_{\sigma} = R_c = 8$, and the iteration of $30k$) and \textit{TensoRF-VM-192-15k} ($R_{\sigma} = 16$, $R_c = 48$, and the iteration of $15k$) for the Blender dataset, and \textit{TensoRF-VM-48} ($R_{\sigma, 1} = R_{\sigma, 2} = 4$, $R_{\sigma, 3} = 16$, $R_{c, 1} = R_{c, 2} = 4$, and $R_{c, 3} = 16$) for the LLFF dataset.
This appendix examines whether MIMO-NeRF is also effective for these models.
Based on the observation that MIMO-TensoRF-VM-192-30k retains image quality when $N_p \leq 2$ on the Blender dataset (Section~\ref{subsec:application_tensorf}), we examined MIMO-TensoRF-VM-48 and MIMO-TensoRF-VM-192-15k with $N_p \in \{ 2, 4 \}$ on the Blender dataset.
Similarly, based on the observation that MIMO-TensoRF-VM-96 can achieve comparable image quality when $N_p \leq 4$ on the LLFF dataset (Section~\ref{subsec:application_tensorf}), we examined MIMO-TensoRF-VM-96 with $N_p \in \{ 2, 4, 8 \}$ on the LLFF dataset.
We evaluated the models using \textit{VGG$_{\text{VGG}}$} and \textit{VGG$_{\text{Alex}}$}, in addition to the metrics described in Section~\ref{subsec:application_tensorf}.

\smallskip\noindent\textbf{Quantitative results.}
Table~\ref{tab:scores_tensorf_ex} lists the results for all the metrics.
Tables~\ref{tab:scores_tensorf_each_blender} and \ref{tab:scores_tensorf_each_llff} summarize the PSNR, SSIM, LPIPS$_{\text{VGG}}$, and LPIPS$_{\text{Alex}}$ scores for each scene in the Blender and LLFF datasets, respectively.
We observed the same tendencies as those described in Section~\ref{subsec:application_tensorf}.
On the Blender dataset, MIMO-TensoRFs can improve the I-time and T-time of the original TensoRFs with similar image quality when $N_p \leq 2$.
On the LLFF dataset, MIMO-TensoRFs can improve the I-time and T-time of the original TensoRFs with similar image quality when $N_p \leq 4$.
These results suggest that MIMO-TensoRF can strengthen the inference/training speed of TensoRF without deteriorating the image quality by adequately selecting $N_p$.

\begin{figure*}[t]
  \centering
  \includegraphics[width=0.84\textwidth]{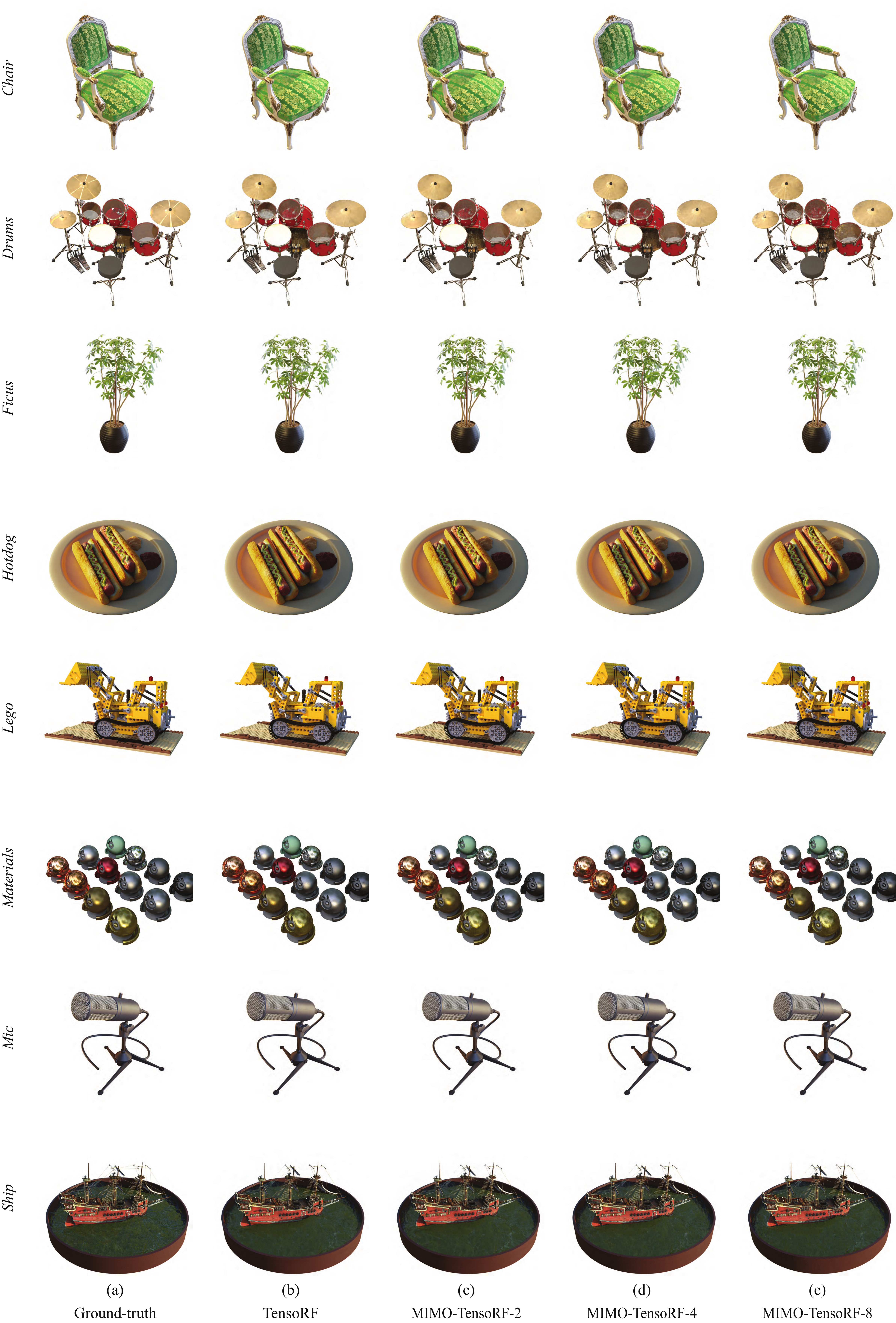}
  \caption{Qualitative comparison between TensoRF, MIMO-TensoRF-2, MIMO-TensoRF-4, and MIMO-TensoRF-8 on the Blender dataset.
    Best viewed zoomed in.
    TensoRF-VM-192-30k was used as a baseline, and MIMO-NeRF was incorporated into it.}
  \label{fig:results_tensorf_blender}
\end{figure*}

\begin{figure*}[t]
  \centering
  \includegraphics[width=\textwidth]{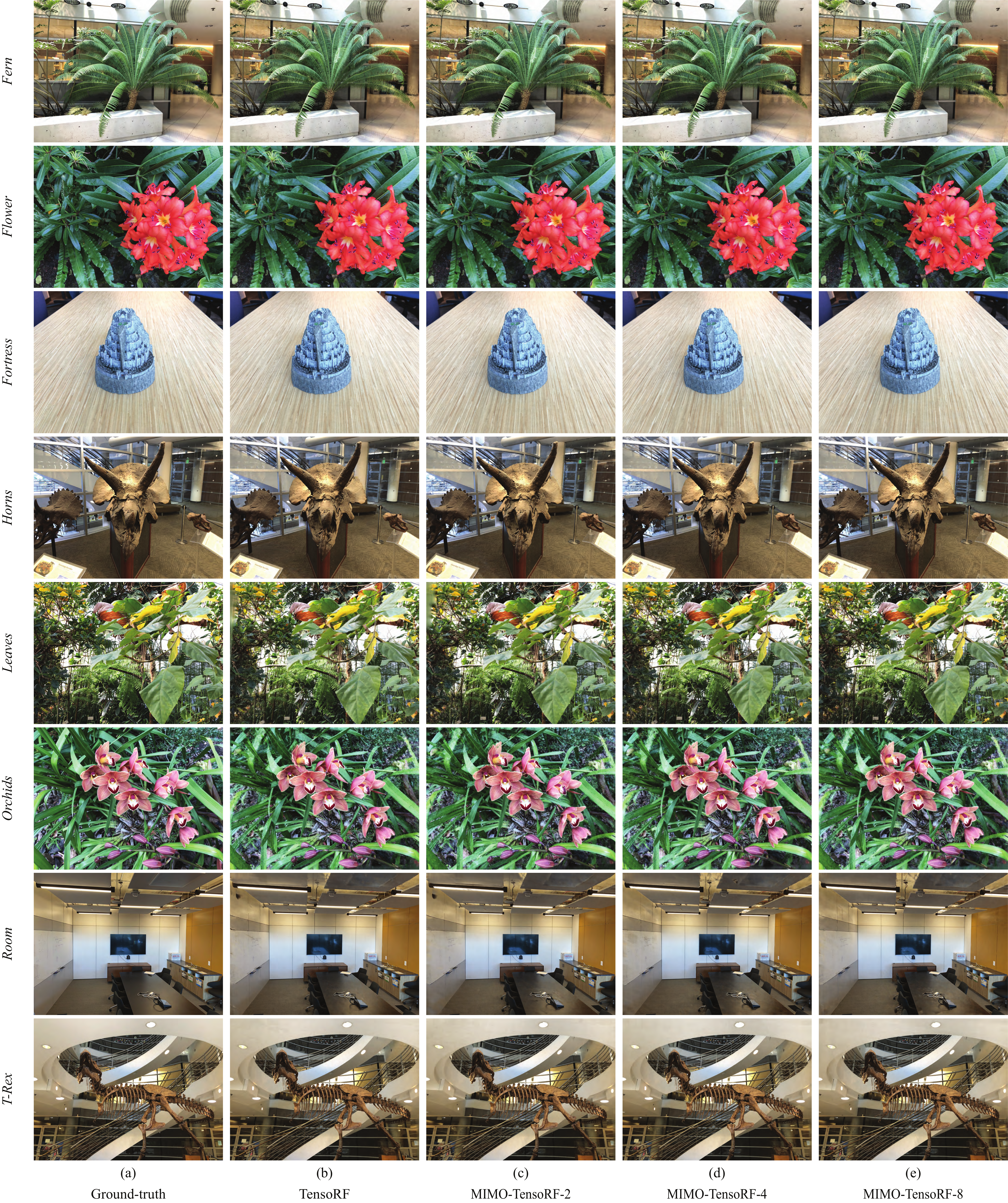}
  \caption{Qualitative comparison between TensoRF, MIMO-TensoRF-2, MIMO-TensoRF-4, and MIMO-TensoRF-8 on the LLFF dataset.
    Best viewed zoomed in.
    TensoRF-VM-96 was used as a baseline, and MIMO-NeRF was incorporated into it.}
  \label{fig:results_tensorf_llff}
\end{figure*}

\smallskip\noindent\textbf{Qualitative results.}
Figures~\ref{fig:results_tensorf_blender} and \ref{fig:results_tensorf_llff} present the qualitative results for the Blender and LLFF datasets, respectively.
Examples of the synthesized videos are provided on the \href{https://www.kecl.ntt.co.jp/people/kaneko.takuhiro/projects/mimo-nerf/}{project page}.\footnoteref{foot:project_page}

\begin{table*}
  \setlength{\tabcolsep}{4pt}
  \begin{center}
    \scriptsize
    \begin{tabularx}{0.8\textwidth}{lc|CCCCCCCC}
      & & \multicolumn{8}{c}{Blender}
      \\
      \multicolumn{1}{c}{Model} & $N_p$ & PSNR$\uparrow$ & SSIM$\uparrow$ & LPIPS$_{\text{VGG}}$ & LPIPS$_{\text{Alex}}$ & \# Run$\downarrow$ & I-time$\downarrow$ & T-time$\downarrow$ & \# Params
      \\
      & & & & & & & (s) & (m) & (M)
      \\ \hline
      TensoRF-VM-48
      & 1 & \second{32.45} & \first{0.957} & \first{0.056} & \first{0.032} & 10.24 & \third{1.16} & \third{9.45} & 4.7
      \\
      MIMO-TensoRF-VM-48-2
      & 2 & \first{32.49} & \first{0.957} & \first{0.056} & \first{0.032} & 4.95 & \second{1.09} & \second{8.95} & 4.7
      \\
      MIMO-TensoRF-VM-48-4
      & 4 & \third{32.25} & \third{0.955} & \third{0.060} & \third{0.034} & 2.49 & \first{1.06} & \first{8.85} & 4.8
      \\ \hline
      TensoRF-VM-192-15k
      & 1 & \second{32.74} & \first{0.961} & \first{0.051} & \first{0.030} & 10.11 & \third{1.27} & \third{5.64} & 18.9
      \\
      MIMO-TensoRF-VM-192-15k-2
      & 2 & \first{32.79} & \first{0.961} & \first{0.051} & \first{0.030} & 4.78 & \second{1.19} & \second{5.36} & 18.9
      \\
      MIMO-TensoRF-VM-192-15k-4
      & 4 & \third{32.55} &  \third{0.958} &  \third{0.055} &  \third{0.032} & 2.40 & \first{1.16} & \first{5.22} & 18.9
      \\ \hline
      TensoRF-VM-192-30k
      & 1 & \second{33.23} & \first{0.963} & \first{0.047} & \first{0.026} & 9.95 & 1.25 & 11.50 & 18.8
      \\
      MIMO-TensoRF-VM-192-30k-2
      & 2 & \first{33.26} & \first{0.963} & \first{0.047} & \first{0.026} & 4.76 & \third{1.18} & \third{10.89} & 18.8      
      \\
      MIMO-TensoRF-VM-192-30k-4
      & 4 & \third{32.98} & \third{0.961} & \third{0.051} & \third{0.028} & 2.40 & \second{1.15} & \second{10.67} & 18.8
      \\
      MIMO-TensoRF-VM-192-30k-8
      & 8 & 32.37 & 0.956 & 0.058 & 0.033 & 1.27 & \first{1.14} & \first{10.57} & 18.9
      \\ \hline \hline
      TensoRF-VM-48~\cite{AChenECCV2022}
      & 1 & 32.39 & 0.957 & 0.057 & 0.032 & -- & -- & -- & --
      \\
      TensoRF-VM-192-15k~\cite{AChenECCV2022}
      & 1 & 32.52 & 0.959 & 0.053 & 0.032 & -- & -- & -- & --
      \\
      TensoRF-VM-192-30k~\cite{AChenECCV2022}
      & 1 & 33.14 & 0.963 & 0.047 & 0.027 & -- & -- & -- & --
      \\
      \\
      & & \multicolumn{8}{c}{LLFF}
      \\
      \multicolumn{1}{c}{Model} & $N_p$ & PSNR$\uparrow$ & SSIM$\uparrow$ & LPIPS$_{\text{VGG}}$ & LPIPS$_{\text{Alex}}$ & \# Run$\downarrow$ & I-time$\downarrow$ & T-time$\downarrow$ & \# Params
      \\
      & & & & & & & (s) & (m) & (M)
      \\ \hline
      TensoRF-VM-48
      & 1 & \third{26.48} & \second{0.832} & \third{0.213} & \third{0.125} & 120.78 & 6.14 & 19.83 & 23.4
      \\
      MIMO-TensoRF-VM-48-2
      & 2 & \first{26.51} & \first{0.833} & \first{0.211} & \first{0.124} & 58.42 & \third{5.70} & \third{18.21} & 23.4
      \\
      MIMO-TensoRF-VM-48-4
      & 4 & \second{26.50} & \second{0.832} & \first{0.211} & \first{0.124} & 28.11 & \second{5.24} & \second{17.20} & 23.4
      \\
      MIMO-TensoRF-VM-48-8
      & 8 & 26.41 & 0.830 & 0.215 & 0.126 & 13.58 & \first{5.03} & \first{16.86} & 23.5
      \\ \hline
      TensoRF-VM-96
      & 1 & \first{26.73} & \first{0.837} & \first{0.201} & \first{0.115} & 126.73 & 6.64 & 23.41 & 46.8
      \\
      MIMO-TensoRF-VM-96-2
      & 2 & \second{26.72} & \first{0.837} & \first{0.201} & \first{0.115} & 62.14 & \third{6.18} & \third{21.63} & 46.8
      \\
      MIMO-TensoRF-VM-96-4
      & 4 & \second{26.72} & \third{0.836} & \third{0.202} & \first{0.115} & 30.16 & \second{5.76} & \second{21.15} & 46.8
      \\
      MIMO-TensoRF-VM-96-8
      & 8 & 26.64 & 0.835 & 0.204 & 0.116 & 14.52 & \first{5.52} & \first{20.68} & 46.9
      \\ \hline \hline
      TensoRF-VM-48~\cite{AChenECCV2022}
      & 1 & 26.51 & 0.832 & 0.217 & 0.135 & -- & -- & -- & --
      \\
      TensoRF-VM-96~\cite{AChenECCV2022}
      & 1 & 26.73 & 0.839 & 0.204 & 0.124 & -- & -- & -- & --
      \\
    \end{tabularx}
  \end{center}
  \caption{Comparison of quantitative scores between TensoRFs and MIMO-TensoRFs.
    The scores for the model with citation~\cite{AChenECCV2022} are taken from another report~\cite{AChenECCV2022}.
    We provide them as references.
    The other scores were calculated in our environment.
    We implemented all the models based on the official TensoRF source code.
    See Appendix~\ref{subsec:implementation_details_tensorf} for the implementation details.
    This table is an extended version of Table~\ref{tab:scores_tensorf}.
    The PSNR, SSIM, LPIPS$_{\text{VGG}}$, and LPIPS$_{\text{Alex}}$ scores for each scene are presented in Tables~\ref{tab:scores_tensorf_each_blender} and \ref{tab:scores_tensorf_each_llff}.}
  \label{tab:scores_tensorf_ex}
\end{table*}

\begin{table*}
  \setlength{\tabcolsep}{0.5pt}
  \begin{center}
    \scriptsize
    \begin{tabularx}{0.8\textwidth}{lc|CCCCCCCC|C}
      \multicolumn{11}{c}{Blender}
      \\
      & & \multicolumn{9}{c}{PSNR$\uparrow$}
      \\
      \multicolumn{1}{c}{Model} & $N_p$
      & Chair & Drums & Ficus & Hotdog & Lego & Materials & Mic & Ship & Avg.
      \\ \hline
      TensoRF-VM-48 & 1
      & \second{34.71} & \first{25.57} & \first{33.44} & \third{36.88} & \second{35.69} & \second{29.38} & \second{33.83} & \firstr{30.07} & \second{32.45}
      \\
      MIMO-TensoRF-VM-48-2 & 2
      & \first{34.85} & \second{25.56} & \second{33.34} & \first{37.00} & \first{35.84} & \first{29.43} & \first{33.91} & \secondr{30.00} & \first{32.49}
      \\
      MIMO-TensoRF-VM-48-4 & 4
      & \third{34.61} & \third{25.24} & \third{33.05} & \second{36.95} & \third{35.61} & \third{29.26} & \third{33.52} & \thirdr{29.75} & \third{32.25}
      \\ \hline
      TensoRF-VM-192-15k & 1
      & \second{35.11} & \first{25.80} & \second{33.73} & \third{37.04} & \second{36.00} & \first{29.80} & \second{34.31} & \firstr{30.10} & \second{32.74}
      \\
      MIMO-TensoRF-VM-192-15k-2 & 2
      & \first{35.32} & \second{25.65} & \first{33.87} & \first{37.22} & \first{36.06} & \second{29.77} & \first{34.35} & \secondr{30.07} & \first{32.79}
      \\
      MIMO-TensoRF-VM-192-15k-4 & 4
      & \third{35.06} & \third{25.36} & \third{33.67} & \second{37.10} & \third{35.77} & \third{29.54} & \third{34.11} & \thirdr{29.80} & \third{32.55}
      \\ \hline
      TensoRF-VM-192-30k & 1
      & \second{35.79} & \first{25.96} & \second{34.14} & \third{37.50} & \first{36.62} & \second{30.10} & \first{34.98} & \firstr{30.72} & \second{33.23}
      \\
      MIMO-TensoRF-VM-192-30k-2 & 2
      & \first{35.91} & \first{25.96} & \first{34.28} & \first{37.64} & \second{36.61} & \first{30.11} & \second{34.97} & \secondr{30.62} & \first{33.26}
      \\
      MIMO-TensoRF-VM-192-30k-4 & 4
      & \third{35.60} & \third{25.56} & \third{34.04} & \second{37.51} & \third{36.42} & \third{29.80} & \third{34.60} & \thirdr{30.28} & \third{32.98}
      \\
      MIMO-TensoRF-VM-192-30k-8 & 8
      & 35.04 & 24.85 & 33.24 & 37.04 & 35.92 & 29.13 & 33.87 & 29.87 & 32.37
      \\ \hline \hline
      TensoRF-VM-48~\cite{AChenECCV2022} & 1
      & 34.68 & 25.58 & 33.37 & 36.81 & 35.51 & 29.45 & 33.59 & 30.12 & 32.39
      \\
      TensoRF-VM-192-15k~\cite{AChenECCV2022} & 1
      & 34.95 & 25.63 & 33.46 & 36.85 & 35.78 & 29.78 & 33.69 & 30.04 & 32.52
      \\
      TensoRF-VM-192-30k~\cite{AChenECCV2022} & 1
      & 35.76 & 26.01 & 33.99 & 37.41 & 36.46 & 30.12 & 34.61 & 30.77 & 33.14
      \\
      \\
      & & \multicolumn{9}{c}{SSIM$\uparrow$}
      \\
      \multicolumn{1}{c}{Model} & $N_p$
      & Chair & Drums & Ficus & Hotdog & Lego & Materials & Mic & Ship & Avg.
      \\ \hline
      TensoRF-VM-48 & 1
      & \second{0.980} & \first{0.930} & \first{0.979} & \third{0.979} & \second{0.979} & \first{0.942} & \first{0.985} & \firstr{0.883} & \first{0.957}
      \\
      MIMO-TensoRF-VM-48-2 & 2
      & \first{0.981} & \first{0.930} & \first{0.979} & \first{0.980} & \first{0.980} & \first{0.942} & \first{0.985} & \secondr{0.881} & \first{0.957}
      \\
      MIMO-TensoRF-VM-48-4 & 4
      & \second{0.980} & \third{0.925} & \third{0.977} & \first{0.980} & \second{0.979} & \third{0.940} & \third{0.983} & \thirdr{0.876} & \third{0.955}
      \\ \hline
      TensoRF-VM-192-15k & 1
      & \second{0.982} & \first{0.935} & \second{0.981} & \second{0.981} & \first{0.982} & \first{0.950} & \first{0.987} & \firstr{0.887} & \first{0.961}
      \\
      MIMO-TensoRF-VM-192-15k-2 & 2
      & \first{0.983} & \second{0.934} & \first{0.982} & \first{0.982} & \first{0.982} & \second{0.949} & \first{0.987} & \secondr{0.886} & \first{0.961}
      \\
      MIMO-TensoRF-VM-192-15k-4 & 4
      & \second{0.982} & \third{0.929} & \second{0.981} & \second{0.981} & \third{0.981} & \third{0.946} & \third{0.986} & \thirdr{0.880} & \third{0.958}
      \\ \hline
      TensoRF-VM-192-30k & 1
      & \first{0.985} & \first{0.937} & \first{0.983} & \first{0.983} & \second{0.983} & \first{0.952} & \first{0.989} & \firstr{0.894} & \first{0.963}
      \\
      MIMO-TensoRF-VM-192-30k-2 & 2
      & \first{0.985} & \first{0.937} & \first{0.983} & \first{0.983} & \first{0.984} & \first{0.952} & \second{0.988} & \secondr{0.893} & \first{0.963}
      \\
      MIMO-TensoRF-VM-192-30k-4 & 4
      & \third{0.984} & \third{0.931} & \third{0.982} & \first{0.983} & \second{0.983} & \third{0.949} & \third{0.987} & \thirdr{0.886} & \third{0.961}
      \\
      MIMO-TensoRF-VM-192-30k-8 & 8
      & 0.982 & 0.922 & 0.978 & 0.980 & 0.981 & 0.942 & 0.984 & 0.877 & 0.956
      \\ \hline \hline
      TensoRF-VM-48~\cite{AChenECCV2022} & 1
      & 0.980 & 0.929 & 0.979 & 0.979 & 0.979 & 0.942 & 0.984 & 0.883 & 0.957
      \\
      TensoRF-VM-192-15k~\cite{AChenECCV2022} & 1
      & 0.982 & 0.933 & 0.981 & 0.980 & 0.981 & 0.949 & 0.985 & 0.886 & 0.959
      \\
      TensoRF-VM-192-30k~\cite{AChenECCV2022} & 1
      & 0.985 & 0.937 & 0.982 & 0.982 & 0.983 & 0.952 & 0.988 & 0.895 & 0.963
      \\
      \\
      & & \multicolumn{9}{c}{LPIPS$_{\text{VGG}}$$\downarrow$}
      \\
      \multicolumn{1}{c}{Model} & $N_p$
      & Chair & Drums & Ficus & Hotdog & Lego & Materials & Mic & Ship & Avg.
      \\ \hline
      TensoRF-VM-48 & 1
      & \second{0.029} & \first{0.085} & \first{0.029} & \second{0.038} & \second{0.023} & \first{0.073} & \first{0.020} & \firstr{0.153} & \first{0.056}
      \\
      MIMO-TensoRF-VM-48-2 & 2
      & \first{0.028} & \second{0.086} & \second{0.030} & \first{0.036} & \first{0.022} & \first{0.073} & \second{0.021} & \secondr{0.154} & \first{0.056}
      \\
      MIMO-TensoRF-VM-48-4 & 4
      & \second{0.029} & \third{0.092} & \third{0.035} & \second{0.038} & \third{0.024} & \third{0.077} & \third{0.025} & \thirdr{0.159} & \third{0.060}
      \\ \hline
      TensoRF-VM-192-15k & 1
      & \second{0.024} & \first{0.076} & \first{0.024} & \second{0.035} & \first{0.020} & \first{0.062} & \first{0.017} & \firstr{0.150} & \first{0.051}
      \\
      MIMO-TensoRF-VM-192-15k-2 & 2
      & \first{0.023} & \second{0.077} & \first{0.024} & \first{0.034} & \first{0.020} & \second{0.063} & \first{0.017} & \firstr{0.150} & \first{0.051}
      \\
      MIMO-TensoRF-VM-192-15k-4 & 4
      & \third{0.025} & \third{0.085} & \third{0.028} & \third{0.036} & \third{0.021} & \third{0.069} & \third{0.021} & \thirdr{0.154} & \third{0.055}
      \\ \hline
      TensoRF-VM-192-30k & 1
      & \second{0.021} & \first{0.071} & \first{0.022} & \second{0.031} & \second{0.018} & \first{0.058} & \first{0.014} & \firstr{0.139} & \first{0.047}
      \\
      MIMO-TensoRF-VM-192-30k-2 & 2
      & \first{0.020} & \second{0.072} & \first{0.022} & \first{0.030} & \first{0.017} & \first{0.058} & \second{0.015} & \secondr{0.140} & \first{0.047}
      \\
      MIMO-TensoRF-VM-192-30k-4 & 4
      & \third{0.022} & \third{0.080} & \third{0.026} & \third{0.032} & \second{0.018} & \third{0.065} & \third{0.019} & \thirdr{0.144} & \third{0.051}
      \\
      MIMO-TensoRF-VM-192-30k-8 & 8
      & 0.026 & 0.089 & 0.032 & 0.041 & 0.021 & 0.076 & 0.025 & 0.153 & 0.058
      \\ \hline \hline
      TensoRF-VM-48~\cite{AChenECCV2022} & 1
      & 0.030 & 0.087 & 0.028 & 0.039 & 0.024 & 0.072 & 0.021 & 0.155 & 0.057
      \\
      TensoRF-VM-192-15k~\cite{AChenECCV2022} & 1
      & 0.026 & 0.078 & 0.025 & 0.038 & 0.021 & 0.063 & 0.020 & 0.153 & 0.053
      \\
      TensoRF-VM-192-30k~\cite{AChenECCV2022} & 1
      & 0.022 & 0.073 & 0.022 & 0.032 & 0.018 & 0.058 & 0.015 & 0.138 & 0.047
      \\
      \\
      & & \multicolumn{9}{c}{LPIPS$_{\text{Alex}}$$\downarrow$}
      \\
      \multicolumn{1}{c}{Model} & $N_p$
      & Chair & Drums & Ficus & Hotdog & Lego & Materials & Mic & Ship & Avg.
      \\ \hline
      TensoRF-VM-48 & 1
      & \first{0.013} & \first{0.057} & \first{0.015} & \second{0.017} & \first{0.009} & \first{0.036} & \first{0.011} & \firstr{0.095} & \first{0.032}
      \\
      MIMO-TensoRF-VM-48-2 & 2
      & \first{0.013} & \first{0.057} & \first{0.015} & \first{0.016} & \first{0.009} & \second{0.037} & \first{0.011} & \secondr{0.096} & \first{0.032}
      \\
      MIMO-TensoRF-VM-48-4 & 4
      & \first{0.013} & \third{0.062} & \third{0.017} & \second{0.017} & \first{0.009} & \third{0.041} & \third{0.013} & \thirdr{0.101} & \third{0.034}
      \\ \hline
      TensoRF-VM-192-15k & 1
      & \first{0.011} & \first{0.054} & \first{0.013} & \second{0.016} & \first{0.008} & \first{0.029} & \first{0.010} & \firstr{0.096} & \first{0.030}
      \\
      MIMO-TensoRF-VM-192-15k-2 & 2
      & \first{0.011} & \second{0.055} & \first{0.013} & \first{0.015} & \first{0.008} & \second{0.030} & \first{0.010} & \secondr{0.097} & \first{0.030}
      \\
      MIMO-TensoRF-VM-192-15k-4 & 4
      & \third{0.012} & \third{0.060} & \third{0.015} & \second{0.016} & \first{0.008} & \third{0.034} & \third{0.011} & \thirdr{0.098} & \third{0.032}
      \\ \hline
      TensoRF-VM-192-30k & 1
      & \first{0.009} & \first{0.049} & \first{0.012} & \second{0.013} & \first{0.007} & \first{0.026} & \first{0.008} & \firstr{0.084} & \first{0.026}
      \\
      MIMO-TensoRF-VM-192-30k-2 & 2
      & \first{0.009} & \second{0.050} & \first{0.012} & \first{0.012} & \first{0.007} & \first{0.026} & \first{0.008} & \secondr{0.086} & \first{0.026}
      \\
      MIMO-TensoRF-VM-192-30k-4 & 4
      & \third{0.010} & \third{0.056} & \third{0.014} & \second{0.013} & \first{0.007} & \third{0.032} & \third{0.009} & \thirdr{0.087} & \third{0.028}
      \\
      MIMO-TensoRF-VM-192-30k-8 & 8
      & 0.011 & 0.064 & 0.018 & 0.017 & 0.008 & 0.041 & 0.013 & 0.094 & 0.033
      \\ \hline \hline
      TensoRF-VM-48~\cite{AChenECCV2022} & 1
      & 0.014 & 0.059 & 0.015 & 0.017 & 0.009 & 0.036 & 0.012 & 0.098 & 0.032
      \\
      TensoRF-VM-192-15k~\cite{AChenECCV2022} & 1
      & 0.013 & 0.056 & 0.014 & 0.017 & 0.009 & 0.029 & 0.013 & 0.101 & 0.032
      \\
      TensoRF-VM-192-30k~\cite{AChenECCV2022} & 1
      & 0.010 & 0.051 & 0.012 & 0.013 & 0.007 & 0.026 & 0.009 & 0.085 & 0.027
      \\
    \end{tabularx}
  \end{center}
  \caption{Comparison of PSNR, SSIM, LPIPS$_{\text{VGG}}$, and LPIPS$_{\text{Alex}}$ for each scene on the Blender dataset between TensoRFs and MIMO-TensoRFs.
    The scores for the model with citation~\cite{AChenECCV2022} are taken from another report~\cite{AChenECCV2022}.
    We provide them as references.
    The other scores were calculated in our environment.
    We implemented all the models based on the official TensoRF source code.
    See Appendix~\ref{subsec:implementation_details_tensorf} for the implementation details.
    The scores for the other metrics are summarized in Table~\ref{tab:scores_tensorf_ex}.}
  \label{tab:scores_tensorf_each_blender}
\end{table*}

\begin{table*}
  \setlength{\tabcolsep}{0.5pt}
  \begin{center}
    \scriptsize
    \begin{tabularx}{0.8\textwidth}{lc|CCCCCCCC|C}
      & & \multicolumn{9}{c}{PSNR$\uparrow$}
      \\
      \multicolumn{1}{c}{Model} & $N_p$
      & Fern & Flower & Fortress & Horns & Leaves & Orchids & Room & T-Rex & Avg.
      \\ \hline
      TensoRF-VM-48 & 1
      & \third{25.18} & \second{27.88} & \third{31.11} & \second{27.83} & \first{21.27} & \third{19.94} & \first{31.66} & 26.99 & \third{26.48}
      \\
      MIMO-TensoRF-VM-48-2 & 2
      & \first{25.23} & \first{27.95} & \second{31.14} & \first{27.88} & \third{21.24} & 19.93 & \second{31.59} & \thirdr{27.08} & \first{26.51}
      \\
      MIMO-TensoRF-VM-48-4 & 4
      & \second{25.21} & \third{27.85} & \first{31.19} & \second{27.83} & \second{21.26} & \second{19.97} & \third{31.51} & \firstr{27.19} & \second{26.50}
      \\
      MIMO-TensoRF-VM-48-8 & 8
      & 25.13 & 27.82 & 31.06 & 27.74 & 21.20 & \first{19.98} & 31.25 & \secondr{27.09} & 26.41
      \\ \hline
      TensoRF-VM-96 & 1
      & 25.00 & \second{28.29} & \second{31.47} & \second{28.35} & \second{21.09} & 19.81 & \first{32.22} & \firstr{27.63} & \first{26.73}
      \\
      MIMO-TensoRF-VM-96-2 & 2
      & \second{25.14} & \first{28.36} & \third{31.43} & \first{28.38} & 21.00 & \third{19.86} & \third{32.17} & 27.40 & \second{26.72}
      \\
      MIMO-TensoRF-VM-96-4 & 4
      & \first{25.16} & \third{28.21} & \first{31.48} & \third{28.29} & \first{21.10} & \second{19.89} & \second{32.18} & \thirdr{27.47} & \second{26.72}
      \\
      MIMO-TensoRF-VM-96-8 & 8
      & \third{25.12} & 28.08 & 31.27 & 28.22 & \third{21.08} & \first{19.98} & 31.87 & \secondr{27.54} & 26.64
      \\ \hline \hline
      TensoRF-VM-48~\cite{AChenECCV2022} & 1
      & 25.31 & 28.22 & 31.14 & 27.64 & 21.34 & 20.02 & 31.80 & 26.61 & 26.51
      \\
      TensoRF-VM-96~\cite{AChenECCV2022} & 1
      & 25.27 & 28.60 & 31.36 & 28.14 & 21.30 & 19.87 & 32.35 & 26.97 & 26.73
      \\
      \\
      & & \multicolumn{9}{c}{SSIM$\uparrow$}
      \\
      \multicolumn{1}{c}{Model} & $N_p$
      & Fern & Flower & Fortress & Horns & Leaves & Orchids & Room & T-Rex & Avg.
      \\ \hline
      TensoRF-VM-48 & 1
      & 0.806 & \second{0.854} & 0.889 & \third{0.865} & \first{0.745} & \first{0.651} & \first{0.946} & 0.898 & \second{0.832}
      \\
      MIMO-TensoRF-VM-48-2 & 2
      & \second{0.808} & \first{0.855} & \first{0.891} & \first{0.868} & \third{0.744} & \first{0.651} & \first{0.946} & \thirdr{0.899} & \first{0.833}
      \\
      MIMO-TensoRF-VM-48-4 & 4
      & \first{0.809} & \third{0.852} & \first{0.891} & \third{0.865} & \first{0.745} & \third{0.650} & \third{0.945} & \firstr{0.901} & \second{0.832}
      \\
      MIMO-TensoRF-VM-48-8 & 8
      & \third{0.807} & 0.850 & \first{0.891} & \second{0.866} & 0.739 & 0.649 & 0.939 & \thirdr{0.899} & 0.830
      \\ \hline
      TensoRF-VM-96 & 1
      & 0.800 & \second{0.861} & \third{0.899} & \second{0.883} & \first{0.744} & 0.643 & \first{0.952} & \firstr{0.910} & \first{0.837}
      \\
      MIMO-TensoRF-VM-96-2 & 2
      & \third{0.803} & \first{0.865} & \second{0.900} & \first{0.884} & \third{0.739} & \second{0.644} & \first{0.952} & 0.907 & \first{0.837}
      \\
      MIMO-TensoRF-VM-96-4 & 4
      & \first{0.806} & \third{0.857} & \first{0.901} & \second{0.883} & \third{0.739} & \second{0.644} & \third{0.950} & \thirdr{0.909} & \third{0.836}
      \\
      MIMO-TensoRF-VM-96-8 & 8
      & \second{0.804} & 0.855 & 0.897 & 0.882 & \second{0.740} & \first{0.648} & 0.945 & \firstr{0.910} & 0.835
      \\ \hline
      TensoRF-VM-48~\cite{AChenECCV2022} & 1
      & 0.816 & 0.859 & 0.889 & 0.859 & 0.746 & 0.655 & 0.946 & 0.890 & 0.832
      \\
      TensoRF-VM-96~\cite{AChenECCV2022} & 1
      & 0.814 & 0.871 & 0.897 & 0.877 & 0.752 & 0.649 & 0.952 & 0.900 & 0.839
      \\
      \\
      & & \multicolumn{9}{c}{LPIPS$_{\text{VGG}}$$\downarrow$}
      \\
      \multicolumn{1}{c}{Model} & $N_p$
      & Fern & Flower & Fortress & Horns & Leaves & Orchids & Room & T-Rex & Avg.
      \\ \hline
      TensoRF-VM-48 & 1
      & 0.244 & \second{0.186} & 0.157 & \third{0.207} & \first{0.226} & \first{0.282} & \first{0.179} & 0.219 & \third{0.213}
      \\
      MIMO-TensoRF-VM-48-2 & 2
      & \third{0.243} & \first{0.184} & \third{0.155} & \first{0.203} & \second{0.227} & \second{0.283} & \first{0.179} & \thirdr{0.216} & \first{0.211}
      \\
      MIMO-TensoRF-VM-48-4 & 4
      & \first{0.240} & \third{0.187} & \second{0.154} & 0.208 & \second{0.227} & \third{0.284} & \first{0.179} & \firstr{0.212} & \first{0.211}
      \\
      MIMO-TensoRF-VM-48-8 & 8
      & \second{0.241} & 0.188 & \first{0.153} & \second{0.205} & 0.233 & 0.285 & 0.196 & \secondr{0.215} & 0.215
      \\ \hline
      TensoRF-VM-96 & 1
      & 0.249 & \second{0.172} & 0.142 & \second{0.180} & \first{0.220} & \first{0.281} & \second{0.162} & \secondr{0.201} & \first{0.201}
      \\
      MIMO-TensoRF-VM-96-2 & 2
      & \third{0.245} & \first{0.168} & \second{0.141} & \first{0.179} & \third{0.227} & \third{0.283} & \first{0.161} & 0.205 & \first{0.201}
      \\
      MIMO-TensoRF-VM-96-4 & 4
      & \second{0.241} & \third{0.176} & \first{0.139} & \third{0.181} & \second{0.226} & 0.284 & \third{0.167} & \firstr{0.199} & \third{0.202}
      \\
      MIMO-TensoRF-VM-96-8 & 8
      & \first{0.240} & 0.179 & \second{0.141} & 0.182 & 0.228 & \second{0.282} & 0.179 & \secondr{0.201} & 0.204
      \\ \hline \hline
      TensoRF-VM-48~\cite{AChenECCV2022} & 1
      & 0.237 & 0.187 & 0.159 & 0.221 & 0.230 & 0.283 & 0.181 & 0.236 & 0.217
      \\
      TensoRF-VM-96~\cite{AChenECCV2022} & 1
      & 0.237 & 0.169 & 0.148 & 0.196 & 0.217 & 0.278 & 0.167 & 0.221 & 0.204
      \\
      \\
      & & \multicolumn{9}{c}{LPIPS$_{\text{Alex}}$$\downarrow$}
      \\
      \multicolumn{1}{c}{Model} & $N_p$
      & Fern & Flower & Fortress & Horns & Leaves & Orchids & Room & T-Rex & Avg.
      \\ \hline
      TensoRF-VM-48 & 1
      & 0.156 & \second{0.113} & 0.078 & \third{0.125} & \first{0.155} & \first{0.195} & \first{0.088} & 0.091 & \third{0.125}
      \\
      MIMO-TensoRF-VM-48-2 & 2
      & \third{0.155} & \first{0.112} & \first{0.075} & \second{0.120} & \second{0.156} & 0.198 & \second{0.089} & \thirdr{0.088} & \first{0.124}
      \\
      MIMO-TensoRF-VM-48-4 & 4
      & \first{0.152} & 0.114 & \first{0.075} & 0.126 & \second{0.156} & \second{0.197} & \second{0.089} & \firstr{0.086} & \first{0.124}
      \\
      MIMO-TensoRF-VM-48-8 & 8
      & \second{0.153} & \second{0.113} & \first{0.075} & \first{0.119} & 0.160 & \second{0.197} & 0.102 & \firstr{0.086} & 0.126
      \\ \hline
      TensoRF-VM-96 & 1
      & 0.156 & \second{0.101} & \third{0.066} & \third{0.103} & \first{0.143} & \second{0.193} & \second{0.076} & \thirdr{0.079} & \first{0.115}
      \\
      MIMO-TensoRF-VM-96-2 & 2
      & \third{0.154} & \first{0.098} & \first{0.065} & \second{0.102} & \third{0.148} & \third{0.194} & \first{0.075} & 0.082 & \first{0.115}
      \\
      MIMO-TensoRF-VM-96-4 & 4
      & \second{0.151} & \second{0.101} & \first{0.065} & \third{0.103} & \second{0.147} & 0.196 & \third{0.082} & \firstr{0.077} & \first{0.115}
      \\
      MIMO-TensoRF-VM-96-8 & 8
      & \first{0.148} & 0.103 & 0.067 & \first{0.101} & 0.152 & \first{0.192} & 0.088 & \firstr{0.077} & 0.116
      \\ \hline \hline
      TensoRF-VM-48~\cite{AChenECCV2022} & 1
      & 0.161 & 0.121 & 0.084 & 0.146 & 0.167 & 0.204 & 0.093 & 0.108 & 0.135
      \\
      TensoRF-VM-96~\cite{AChenECCV2022} & 1
      & 0.155 & 0.106 & 0.075 & 0.123 & 0.153 & 0.201 & 0.082 & 0.099 & 0.124
      \\
    \end{tabularx}
  \end{center}
  \caption{Comparison of PSNR, SSIM, LPIPS$_{\text{VGG}}$, and LPIPS$_{\text{Alex}}$ for each scene on the LLFF dataset between TensoRFs and MIMO-TensoRFs.
    The scores for the model with citation~\cite{AChenECCV2022} are taken from another report~\cite{AChenECCV2022}.
    We provide them as references.
    The other scores were calculated in our environment.
    We implemented all the models based on the official TensoRF source code.
    See Appendix~\ref{subsec:implementation_details_tensorf} for the implementation details.
    The scores for the other metrics are summarized in Table~\ref{tab:scores_tensorf_ex}.}
  \label{tab:scores_tensorf_each_llff}
\end{table*}

\clearpage
\clearpage
\section{Implementation details}
\label{sec:implementation_details}

The following implementation details are provided in this appendix:
\begin{itemize}
\item Appendix~\ref{subsec:implementation_details_nerf}:
  Implementation details of NeRF (Sections~\ref{subsec:investigation_performance}--\ref{subsec:ablation_studies} and Appendices~\ref{subsec:effect_grouping}--\ref{subsec:comparison_with_autoint})
\item Appendix~\ref{subsec:implementation_details_donerf}:
  Implementation details of DONeRF (Section~\ref{subsec:application_donerf} and Appendix~\ref{subsec:detailed_analysis_donerf})
\item Appendix~\ref{subsec:implementation_details_tensorf}:
  Implementation details of TensoRF (Section~\ref{subsec:application_tensorf} and Appendix~\ref{subsec:detailed_analysis_tensorf})
\end{itemize}

\subsection{Implementation details of NeRF}
\label{subsec:implementation_details_nerf}

\subsubsection{Datasets}
\label{subsubsec:dataset_nerf}

In the experiments discussed in Sections~\ref{subsec:investigation_performance}--\ref{subsec:ablation_studies}, we used two datasets commonly employed in previous studies on NeRFs.
The detailed information is as follows:

\smallskip\noindent\textbf{Blender dataset~\cite{BMildenhallECCV2020}}.
The dataset included eight scenes: \textit{Chair}, \textit{Drums}, \textit{Ficus}, \textit{Hotdog}, \textit{Lego}, \textit{Materials}, \textit{Mic}, and \textit{Ship}.
Each scene contained $360^\circ$ views of complex objects at a resolution of $800 \times 800$ pixels.
They were rendered using a Blender Cycles path tracer and exhibited complicated geometries and non-Lambertian materials.
For the training and testing, 100 and 200 views were used, respectively.
The data were downloaded from the NeRF authors' website~\cite{BMildenhallECCV2020}.\footnote{\label{foot:nerf-data}\url{https://drive.google.com/drive/folders/128yBriW1IG_3NJ5Rp7APSTZsJqdJdfc1}}
The license information is provided on the website.

\smallskip\noindent\textbf{Local Light Field Fusion (LLFF) dataset~\cite{BMildenhallTOG2019}.}
Specifically, we used the dataset with addition obtained from~\cite{BMildenhallECCV2020}.
The dataset consists of eight complex real-world scenes: \textit{Fern}, \textit{Flower}, \textit{Fortress}, \textit{Horns}, \textit{Leaves}, \textit{Orchids}, \textit{Room}, and \textit{T-Rex}.
Each of these included $20$--$62$ forward-facing views at a resolution of $1008 \times 756$ pixels.
They were captured using a forward-facing handheld cell phone.
One-eighth of the images were used for testing, and the rest were used for training.
The data were downloaded from the NeRF authors' website~\cite{BMildenhallECCV2020}.\footnoteref{foot:nerf-data}
The license information is provided on the website.

\smallskip\noindent
As mentioned in Section~\ref{subsec:ablation_studies}, we primarily used half-sized images following the default settings of an open-source NeRF code\footnoteref{foot:nerf-pytorch} to better investigate the various configurations.
We also used full-sized images for representative cases to confirm whether the effectiveness of MIMO-NeRF was independent of the image size.
We discuss these cases in Appendix~\ref{subsec:full_size}.

\subsubsection{Model configurations}
\label{subsubsec:model_nerf}

\noindent\textbf{NeRF.}
We implemented the baseline NeRF using the open-source code of NeRF.\footnoteref{foot:nerf-pytorch}
The model configuration of the baseline NeRF followed the default settings provided in the code.
Specifically, the input position $\mathbf{x} \in \mathbb{R}^3$ and view direction $\mathbf{d} \in \mathbb{S}^2$ were encoded to a $63$-dimensional vector $\gamma(\mathbf{x})$ and $27$-dimensional vector $\gamma(\mathbf{d})$, respectively, using positional encoding~\cite{BMildenhallECCV2020,MTancikNeurIPS2020}.
Subsequently, the encoded position $\gamma(\mathbf{x})$ was applied to an $8$-layer MLP with rectified unit (ReLU) activation~\cite{VNairICML2010}, each layer of which had $256$ hidden units.
The MLP included a skip connection that incorporated $\gamma(\mathbf{x})$ into the fifth layer.
The volume density $\sigma \in \mathbb{R}^+$ was calculated from the output of the MLP using a linear layer.
At a different branch, the output of the MLP was converted using a linear layer with $256$ hidden units, and the encoded direction $\gamma(\mathbf{d})$ was then concatenated into the converted result.
After the concatenated vector was converted to a $128$ vector using a $1$-layer MLP with ReLU activation, it was used to calculate the RGB color $\mathbf{c} \in \mathbb{R}^3$ using an additional linear layer.
We used the same network architecture for coarse and fine MLPs.
For the half-sized images, the numbers of coarse and fine samples (i.e., $N_c$ and $N_f$) were set to $64$ and $128$ for the Blender dataset and to $64$ and $64$ for the LLFF dataset, respectively.
For the full-sized images, $N_c$ and $N_f$ were set to $64$ and $128$, respectively, for both datasets.

\smallskip\noindent\textbf{MIMO-NeRF.}
MIMO-NeRF has the same network architecture as the baseline NeRF, except for the inputs and outputs.
Particularly, the above-mentioned network was modified to accept $N_p$ inputs, that is, $(\mathbf{x}_i, \dots, \mathbf{x}_j)$, with view direction $\mathbf{d}$, and produce $N_p$ outputs, that is, $(\mathbf{c}_i, \dots, \mathbf{c}_j)$ and $(\sigma_i, \dots, \sigma_j)$, where $N_p$ was the number of grouped samples and $j = i + N_p - 1$.
The other parameters, such as the dimensions of the hidden units, number of layers, type of activation function, $N_c$, and $N_f$, were the same as those in the baseline NeRF.

\begin{table}[t]
  \setlength{\tabcolsep}{1.5pt}
  \begin{center}
    \scriptsize
    \begin{tabularx}{\columnwidth}{l|CCCCc|CCCCc}
      & \multicolumn{5}{c}{Blender} & \multicolumn{5}{c}{LLFF}
      \\
      \multicolumn{1}{c|}{Model} & $N_c$ & $N_f$ & $N_p$ & $F$ & FLOPs & $N_c$ & $N_f$ & $N_p$ & $F$ & FLOPs
      \\
      & & & & & (M) & & & & & (M)
      \\ \hline
      NeRF
      & 64 & 128 & 1 & 256 & 303.82
      & 64 & 64 & 1 & 256 & 227.87
      \\ \hline
      MIMO-NeRF
      & 64 & 128 & 2 & 256 & 160.33
      & 64 & 64 & 2 & 256 & 120.25
      \\
      NeRF-few
      & 34 & 68 & 1 & 256 & 161.41
      & 34 & 34 & 1 & 256 & 121.06
      \\
      NeRF-small
      & 64 & 128 & 1 & 184 & 160.72
      & 64 & 64 & 1 & 184 & 120.54
      \\ \hline
      MIMO-NeRF
      & 64 & 128 & 4 & 256 & 88.59
      & 64 & 64 & 4 & 256 & 66.44
      \\
      NeRF-few
      & 19 & 38 & 1 & 256 & 90.20
      & 19 & 19 & 1 & 256 & 67.65
      \\
      NeRF-small
      & 64 & 128 & 1 & 135 & 89.09
      & 64 & 64 & 1 & 135 & 66.82
      \\ \hline
      MIMO-NeRF
      & 64 & 128 & 8 & 256 & 52.72
      & 64 & 64 & 8 & 256 & 39.54
      \\
      NeRF-few
      & 11 & 22 & 1 & 256 & 52.22
      & 11 & 11 & 1 & 256 & 39.16
      \\
      NeRF-small
      & 64 & 128 & 1 & 103 & 53.62
      & 64 & 64 & 1 & 103 & 40.22
      \\
    \end{tabularx}
  \end{center}
  \caption{Comparison of the number of coarse samples ($N_c$), number of fine samples ($N_f$), number of grouped samples ($N_p$), number of features in a hidden layer ($F$), and FLOPs between NeRF, MIMO-NeRF, NeRF-few, and NeRF-small.
    The hyperparameters of NeRF-few and NeRF-small were adjusted such that their FLOPs became almost the same as that of MIMO-NeRF.}
  \label{tab:comparison_parameters}
\end{table}

\smallskip\noindent\textbf{NeRF-few.}
In NeRF-few, which was used in the experiment described in Section~\ref{subsec:investigation_tradeoff}, the number of samples ($N_c$ and $N_f$) was adjusted such that its FLOPs became almost the same as those of MIMO-NeRF.
Detailed values are listed in Table~\ref{tab:comparison_parameters}.

\smallskip\noindent\textbf{NeRF-small.}
In NeRF-small, which was used in the experiment discussed in Section~\ref{subsec:investigation_tradeoff}, the number of features in the hidden layers ($F$) was adjusted such that its FLOPs were almost the same as those of MIMO-NeRF.
Detailed values are listed in Table~\ref{tab:comparison_parameters}.

\subsubsection{Training settings}
\label{subsubsec:training_nerf}

\noindent\textbf{Half-sized images.}
For a fair comparison, we trained all the models using the same training settings except that in MIMO-NeRF, the NeRF loss function, i.e., $\mathcal{L}_{\text{pixel}}$ (Equation~\ref{eq:pixel_loss}), was replaced with $\mathcal{L}_{\text{MIMO}} = \mathcal{L}_{\text{pixel}}^{\text{MIMO}} + \lambda \mathcal{L}_{\text{3D}}$ (Equation~\ref{eq:mimo_objective}).
Specifically, when we trained the models using half-sized images, we referred to the default settings provided in the open-source code of NeRF.\footnoteref{foot:nerf-pytorch}
More precisely, the models were trained for $200k$ iterations using the Adam optimizer~\cite{DPKingmaICLR2015} with an initial learning rate of $5 \times 10^{-4}$ and momentum terms $\beta_1$ and $\beta_2$ of $0.9$ and $0.999$, respectively.
The batch size was set to $1024$ rays.
For MIMO-NeRF, we set $\lambda = 1$ for the Blender dataset and $\lambda = 0.4$ for the LLFF dataset.

\smallskip\noindent\textbf{Full-size images.}
For full-size images, we trained the models according to the configurations provided in the official NeRF source code~\cite{BMildenhallECCV2020}.\footnote{\label{foot:nerf}\url{https://github.com/bmild/nerf}}
For the Blender dataset, the models were trained for $500k$ iterations using the Adam optimizer~\cite{DPKingmaICLR2015} with an initial learning rate of $5 \times 10^{-4}$ and momentum terms $\beta_1$ and $\beta_2$ of $0.9$ and $0.999$, respectively.
The batch size was set to $1024$ rays.
For the LLFF dataset, the models were trained for $200k$ iterations using the Adam optimizer~\cite{DPKingmaICLR2015} with an initial learning rate of $5 \times 10^{-4}$, $\beta_1$ of $0.9$, and $\beta_2$ of $0.999$.
The batch size was set to $4096$ rays.
For MIMO-NeRF, $\lambda$ was set to $1$ for the Blender dataset and $0.4$ for the LLFF dataset.

\subsubsection{Evaluation metrics}
\label{subsubsec:evaluation_nerf}

We used seven evaluation metrics to measure the performance of NeRF and MIMO-NeRF quantitatively: peak signal-to-noise ratio (\textit{PSNR}), structural similarity index (\textit{SSIM})~\cite{ZWangTIP2004}, learned perceptual image patch quality (\textit{LPIPS})~\cite{RZhangCVPR2018}, number of MLPs running (\textit{\#~Run}), inference time (\textit{I-time}), training time (\textit{T-time}), and number of parameters (\textit{\#~Params}).
The PSNR, SSIM, and LPIPS were used as image quality metrics, following the original NeRF study~\cite{BMildenhallECCV2020}.
I-time and T-time were used to measure the inference and training speeds, respectively.
\#~Run and the \#~Params were provided as supplement information.
The details of these metrics are as follows:

\smallskip\noindent\textbf{PSNR.}
PSNR is a metric that is widely used for assessing the signal quality and is calculated as $\text{PSNR} = -10 \log_{10} \| \mathbf{\hat{I}} - \mathbf{I} \|_2^2$, where $\mathbf{\hat{I}}$ and $\mathbf{I}$ denote the synthesized and ground-truth images, respectively, assuming that images are in $[0, 1]$.
It measures the ratio between the maximum possible power of a signal and the power of the noise, which affects the signal quality.
The larger the PSNR, the better the image quality.

\smallskip\noindent\textbf{SSIM.}
SSIM measures the structural similarity between two images and is commonly used to evaluate image quality.
The larger the SSIM, the better the image quality.

\smallskip\noindent\textbf{LPIPS.}
LPIPS measures the distance between two images using the features of a pretrained DNN.
The LPIPS has been demonstrated to have a better correlation with human perceptual judgment than the PSNR or SSIM~\cite{RZhangCVPR2018}.
We used the VGG network~\cite{KSimonyanICLR2015} as the pretrained DNN, following the NeRF study~\cite{BMildenhallECCV2020}.
The smaller the LPIPS, the better the image quality.

\smallskip\noindent\textbf{\#~Run.}
\#~Run indicates the number of MLPs running required for rendering a single pixel.
In NeRF and MIMO-NeRF, it is calculated as $\frac{N_c}{N_p} + \frac{N_c + N_f}{N_p}$, where $\frac{N_c}{N_p}$ is the \#~Run for the MLP in the coarse strategy, and $\frac{N_c + N_f}{N_p}$ is the \#~Run for the MLP in the fine strategy.
In the baseline NeRF, $N_p = 1$.
The smaller the value of \#~Run, the faster the rendering speed when the speed for each run is the same.

\smallskip\noindent\textbf{I-time.}
The inference time was measured using a single NVIDIA GeForce RTX 3080 Ti Laptop GPU.
The smaller the I-time, the faster the inference.
For simplicity and a fair comparison, we measured the inference time using a standard PyTorch implementation.\footnoteref{foot:nerf-pytorch}
Optimizing the implementation for faster inference (e.g., using custom CUDA kernels) would be interesting for future research.

\smallskip\noindent\textbf{T-time.}
The training time was measured using a single NVIDIA A100-SXM4-80GB GPU.
The smaller the T-time, the faster the training.
Similar to I-time, for simplicity and a fair comparison, we measured the training time using a standard PyTorch implementation.\footnoteref{foot:nerf-pytorch}
Optimizing the implementation for faster training (e.g., using custom CUDA kernels) would be interesting for future research.

\smallskip\noindent\textbf{\#~Params.}
\#~Params indicates the number of parameters of the MLPs, including one in the coarse strategy and the other in the fine strategy.
As mentioned in Section~\ref{subsec:ablation_studies}, \#~Params increases in MIMO-NeRF mainly because the total dimension of the encoded position $\gamma(\mathbf{x})$ increased by $N_p$ times according to the increase in the inputs, as described in Appendix~\ref{subsubsec:model_nerf}.
It should be noted that MIMO-NeRF has the same network as the baseline NeRF except for the inputs and outputs; therefore, the \#~Params does not increase $N_p$ times.
For example, in the experiments discussed in Section~\ref{subsec:ablation_studies}, the \#~Params increased by $1.06$, $1.17$, and $1.39$ times when $N_p$ was $2$, $4$, and $8$, respectively.

\subsection{Implementation details of DONeRF}
\label{subsec:implementation_details_donerf}

\subsubsection{Dataset}
\label{subsubsec:dataset_donerf}

In the experiments discussed in Section~\ref{subsec:application_donerf}, the models were evaluated using the \textit{DONeRF dataset} introduced by DONeRF~\cite{TNeffCVF2021}.
The detailed information is as follows:

\smallskip\noindent\textbf{DONeRF dataset~\cite{TNeffCVF2021}.}
The dataset included six synthetic indoor and outdoor scenes: \textit{Barbershop}, \textit{Bulldozer}, \textit{Classroom}, \textit{Forest}, \textit{Pavillon}, and \textit{San Miguel}.
They exhibit fine and high-frequency details and a wide depth range.
Each scene included $300$ forward-facing views with $800 \times 800$ pixels each.
They were rendered using the Blender Cycles path tracer.
The poses were randomly sampled within the view cell, where the rotation was limited to $30^\circ$ in pitch and $20^\circ$ in yaw relative to the initial camera direction.
For training, validation, and testing, $70\%$, $10\%$, and $20\%$ of images were used, respectively.
Following the original DONeRF study~\cite{TNeffCVF2021}, the images were downsampled to $400 \times 400$ pixels to accelerate the training.
We downloaded the data from the DONeRF authors' website~\cite{TNeffCVF2021}.\footnote{\label{foot:donerf-data}\url{https://repository.tugraz.at/records/jjs3x-4f133}}
The license information is provided on the website.

\subsubsection{Model configurations}
\label{subsubsec:model_donerf}

\noindent\textbf{DONeRF.}
We implemented DONeRF using the source code provided by the authors~\cite{TNeffCVF2021}.\footnoteref{foot:donerf}
In particular, DONeRF was composed of two networks: a depth oracle network and a shading network.

\smallskip\noindent\textit{Depth oracle network.}
The depth oracle network predicted the depth from the position $\mathbf{x} \in \mathbb{R}^3$ and view direction $\mathbf{d} \in \mathbb{S}^2$.
In this network, positional encoding was not adopted for the inputs because it has been demonstrated that it does not improve performance~\cite{TNeffCVF2021}.
After $\mathbf{x}$ and $\mathbf{d}$ were concatenated, they were converted to depth using an $8$-layer MLP, where each layer had $256$ hidden units and ReLU activation~\cite{VNairICML2010} except for the last output layer.

\smallskip\noindent\textit{Shading network.}
The shading network predicted the RGB color $\mathbf{c} \in \mathbb{R}^3$ and the volume density $\sigma \in \mathbb{R}^+$ from $\mathbf{x}$ and $\mathbf{d}$ for the samples selected by the depth oracle network.
It had the same network architecture as that of the depth oracle network except for the following two points: (1) positional encoding~\cite{BMildenhallECCV2020,MTancikNeurIPS2020} was applied to $\mathbf{x}$ and $\mathbf{d}$ to obtain a $63$-dimensional vector $\gamma (\mathbf{x})$ and $27$-dimensional vector $\gamma (\mathbf{d})$, respectively, and (2) only $\gamma (\mathbf{x})$ was used at the first layer and $\gamma (\mathbf{d})$ was concatenated to the feature vector before the last layer.

\smallskip\noindent
In \textit{DONeRF-$N_s$} (e.g., DONeRF-16), the number of selected samples in the shading network was set to $N_s$ (e.g., 16).

\smallskip\noindent\textbf{MIMO-DONeRF.}
We incorporated the MIMO-NeRF concept into the shading network because the depth oracle network is already a fast network that runs only once for each ray.
The difference between the shading network of DONeRF and that of MIMO-DONeRF is limited to the difference in the inputs and outputs.
Specifically, the shading network of DONeRF was modified to accept $N_p$ inputs, that is, $(\mathbf{x}_i, \dots, \mathbf{x}_j)$, with view direction $\mathbf{d}$, and generate $N_p$ outputs, that is, $(\mathbf{c}_i, \dots, \mathbf{c}_j)$ and $(\sigma_i, \dots, \sigma_j)$, where $N_p$ was the number of grouped samples and $j = i + N_p - 1$.
The other parameters, such as the dimensions of the hidden units, number of layers, and type of activation function, were the same as those in DONeRF.
In \textit{MIMO-DONeRF-$N_s$/$N_p$} (e.g., MIMO-DONeRF-16/4), the number of samples selected by the depth oracle network was set to $N_s$ (e.g., $16$), and the number of grouped samples was set to $N_p$ (e.g., $4$).

\subsubsection{Training settings}
\label{subsubsec:training_donerf}

For a fair comparison, we trained DONeRF and MIMO-DONeRF using the same configurations, except that in MIMO-DONeRF, $\mathcal{L}_{\text{MIMO}}$ (Equation~\ref{eq:mimo_objective}) was used as an alternative to the NeRF loss function, that is, $\mathcal{L}_{\text{pixel}}$ (Equation~\ref{eq:pixel_loss}).
Specifically, we trained them using the default settings provided in the official DONeRF source code.\footnoteref{foot:donerf}
The depth oracle and shading networks were separately trained for $300k$ iterations using the Adam optimizer~\cite{DPKingmaICLR2015} with a learning rate of $5 \times 10^{-4}$ and momentum terms $\beta_1$ and $\beta_2$ of $0.9$ and $0.999$, respectively.
The batch size was set to $4096$ rays.
The hyperparameter for MIMO-DONeRF was set to $\lambda = 0.001$.

\subsubsection{Evaluation metrics}
\label{subsubsec:evaluation_donerf}

We used six evaluation metrics to quantitatively investigate the performance of DONeRF and MIMO-DONeRF: \textit{PSNR}, \textit{FLIP}~\cite{PAnderssonCGIT2020}, \textit{\#~Run}, \textit{I-time}, \textit{T-time}, and \textit{\#~Params}.
The PSNR and FLIP were used as image quality metrics, following the original DONeRF study~\cite{TNeffCVF2021}.
I-time and T-time were used to assess inference and training speeds, respectively.
\#~Run and \#~Params were provided as supplement information.
The definitions of PSNR, I-time, T-time, and \#~Params are the same as those in Appendix~\ref{subsubsec:evaluation_nerf}.
Detailed information on the other two metrics (FLIP and \#~Run) is as follows:

\smallskip\noindent\textbf{FLIP.}
FLIP is a metric that evaluates the differences between rendered images and the corresponding ground-truth images.
The effectiveness of the FLIP was demonstrated through a user study~\cite{PAnderssonCGIT2020}.
The smaller the FLIP, the better the image quality.

\smallskip\noindent\textbf{\#~Run.}
In DONeRF and MIMO-DONeRF, \#~Run is calculated as $1 + \frac{N_s}{N_p}$, where $N_s$ indicates the number of samples selected by the depth oracle network and used as inputs in the shading networks, and $N_p$ indicates the number of grouped samples.
In DONeRF, $N_p$ is $1$.
In the above equation, the first term, $1$, represents \#~Run for the depth oracle network, and the second term, $\frac{N_s}{N_p}$, represents \#~Run for the shading network.
The smaller the value of \#~Run, the faster the rendering speed when the speed for each run is the same.

\subsection{Implementation details of TensoRF}
\label{subsec:implementation_details_tensorf}

\subsubsection{Datasets}
\label{subsubsec:dataset_tensorf}

In the experiments described in Section~\ref{subsec:application_tensorf}, we evaluated performance using the Blender and LLFF datasets.
In particular, full-sized images were used.
The details of these two datasets are presented in Appendix~\ref{subsubsec:dataset_nerf}.

\subsubsection{Model configurations}
\label{subsubsec:model_tensorf}

\noindent\textbf{TensoRF.}
TensoRF was implemented using the official source code provided by the authors ~\cite{AChenECCV2022}.\footnoteref{foot:tensorf}
Particularly, in the experiments described in Section~\ref{subsec:application_tensorf}, we used two variants of TensoRF to achieve the best image quality:
For the Blender dataset, we used \textit{TensoRF-VM-192-30k}, which had $192$ components with $R_{\sigma} = 16$ and $R_c = 48$.
For the LLFF dataset, we used \textit{TensoRF-VM-96}, which had $96$ components with $R_{\sigma, 1} = R_{\sigma, 2} = 4$, $R_{\sigma, 3} = 16$, $R_{c, 1} = R_{c, 2} = 12$, and $R_{c, 3} = 48$.
In the experiments described in Appendix~\ref{subsec:detailed_analysis_tensorf}, we additionally used three variants of TensoRF:
For the Blender dataset, we used \textit{TensoRF-VM-48}, which had $48$ components with $R_{\sigma} = R_c = 8$, and \textit{TensoRF-VM-192-15k}, which had the same network architecture as that of TensoRF-VM-192-30k but the number of training iterations was halved.
For the LLFF dataset, we use \textit{TensoRF-VM-48}, which had $48$ components with $R_{\sigma, 1} = R_{\sigma, 2} = 4$, $R_{\sigma, 3} = 16$, $R_{c, 1} = R_{c, 2} = 4$, and $R_{c, 3} = 16$.
In all models, a two-layer MLP with $128$-dimensional hidden layers and ReLU activation~\cite{VNairICML2010} was used as the RGB color decoding function.
The MLP receives an embedding of the viewing direction and features extracted from the tensor factors.
Embedding was performed using positional encoding~\cite{BMildenhallECCV2020,MTancikNeurIPS2020} with frequencies of two.

\smallskip\noindent\textbf{MIMO-TensoRF.}
We applied the MIMO-NeRF concept to the RGB color decoding function, that is, the two-layer MLP, and changed this MLP from a SISO MLP to a MIMO MLP.
More precisely, we modified the MLP to receive $N_p$ features, that is, $(\mathbf{f_i}, \dots, \mathbf{f_j})$, with view direction $\mathbf{d}$, and produced $N_p$ RGB colors, that is, $(\mathbf{c}_i, \dots, \mathbf{c}_j)$, where $N_p$ was the number of grouped samples, and $j = i + N_p - 1$.
Other parameters, such as the dimensions of the hidden units, number of layers, and type of activation function, were the same as those used in TensoRF.
In the experiments, we varied $N_p \in \{ 2, 4, 8 \}$ and denoted MIMO-TensoRF with $N_p$ as \textit{MIMO-TensoRF-$N_p$}.

\subsubsection{Training settings}
\label{subsubsec:training_tensorf}

For a fair comparison, we trained TensoRF and MIMO-TensoRF using the same training settings.
As discussed in Section~\ref{subsec:application_tensorf}, in MIMO-TensoRF, the correspondence ambiguity is relatively small because the volume density, $\sigma$, is calculated using an unambiguous explicit representation.
Hence, we trained MIMO-TensoRF using standard TensoRF loss functions and did not use $\mathcal{L}_{\text{MIMO}}$ while prioritizing the training speed.
Specifically, we trained TensoRF and MIMO-TensoRF using the default settings provided in the official TensoRF source code.\footnoteref{foot:tensorf}
The models were trained for $T$ iterations using the Adam optimizer~\cite{DPKingmaICLR2015} with initial learning rates of $0.02$ for tensor factors and $0.001$ for the MLP decoder and momentum terms $\beta_1$ and $\beta_2$ of $0.9$ and $0.99$, respectively.
For the Blender dataset, $T$ was set to $30k$ except for TensoRF-VM-192-15k and MIMO-TensoRF-VM-192-15k, where $T$ was set to $15k$.
For the LLFF dataset, $T$ was set to $25k$.
The batch size was set to $4096$ rays.

\subsubsection{Evaluation metrics}
\label{subsubsec:evaluation_tensorf}

We used eight evaluation metrics to assess the performance of TensoRF and MIMO-TensoRF quantitatively:
\textit{PSNR}, \textit{SSIM}, \textit{LPIPS$_{\text{VGG}}$}, \textit{LPIPS$_{\text{Alex}}$}, \textit{\#~Run}, \textit{I-time}, \textit{T-time}, and \textit{\#~Params}.
PSNR, SSIM, LPIPS$_{\text{VGG}}$, and LPIPS$_{\text{Alex}}$ were used as image quality metrics following the original TensoRF study~\cite{AChenECCV2022}.
I-time and T-time were used as the inference and training speed metrics, respectively.
\#~Run and \#~Params were provided as supplement information.
The definitions of PSNR, SSIM, I-time, T-time, and \#~Params are the same as those in Appendix~\ref{subsubsec:evaluation_nerf}.
Detailed information on the other three metrics (LPIPS$_{\text{VGG}}$, LPIPS$_{\text{Alex}}$, and \#~Run) is as follows:

\smallskip\noindent\textbf{LPIPS$_{\text{VGG}}$.}
This metric is identical to the LPIPS metric described in Appendix~\ref{subsubsec:evaluation_nerf}.
LPIPS$_{\text{VGG}}$ measures the distance between two images by using the feature space of the VGG network~\cite{KSimonyanICLR2015}.
The smaller the LPIPS$_{\text{VGG}}$, the better the image quality.

\smallskip\noindent\textbf{LPIPS$_{\text{Alex}}$.}
LPIPS$_{\text{Alex}}$ measures the distance between two images using the feature space of the Alex network~\cite{AKrizhevskyNIPS2012}.
The smaller the LPIPS$_{\text{Alex}}$, the better the image quality.

\smallskip\noindent\textbf{\#~Run.}
In TensoRF and MIMO-TensoRF, only samples with weights (i.e., $T_i \alpha_i$ in Equation~\ref{eq:volume_rendering}) greater than the threshold were provided to the RGB decoding function.
Therefore, the number of samples, i.e., $N$, was adaptively determined per scene and per pixel.
Consequently, $\text{\#~Run} = \frac{N}{N_p}$ was also adaptively determined for each scene and pixel.
In our experiments, we report the values averaged over the scenes and pixels.

\end{document}